\documentclass[11pt, a4paper, logo, twocolumn, internal, copyright, nonumbering]{googledeepmind}

\usepackage[authoryear, sort&compress, round]{natbib}
\usepackage{multirow}
\usepackage{listings}
\usepackage{caption}
\usepackage{cuted}
\usepackage{subcaption}
\usepackage{amsmath}
\usepackage{booktabs}
\usepackage{xcolor}
\newcommand{\beginsupplement}{%
        \setcounter{tocdepth}{3}
        \setcounter{secnumdepth}{3}
        \renewcommand{\thesection}{S\arabic{section}}%
        \setcounter{table}{0}
        \renewcommand{\thetable}{S\arabic{table}}%
        \setcounter{figure}{0}
        \renewcommand{\thefigure}{S\arabic{figure}}%
     }

\title{AlphaEarth Foundations: An embedding field model for accurate and efficient global mapping from sparse label data}

\correspondingauthor{cfb@deepmind.com}

\author[*,1]{Christopher F. Brown}
\author[*,1]{Michal R. Kazmierski}
\author[*,2]{Valerie J. Pasquarella}
\author[2]{William J. Rucklidge}
\author[1]{Masha Samsikova}
\author[1]{Chenhui Zhang}
\author[1]{Evan Shelhamer}
\author[2]{Estefania Lahera}
\author[1]{Olivia Wiles}
\author[2]{Simon Ilyushchenko}
\author[2]{Noel Gorelick}
\author[1]{Lihui Lydia Zhang}
\author[1]{Sophia Alj}
\author[2]{Emily Schechter}
\author[2]{Sean Askay}
\author[2]{Oliver Guinan}
\author[2]{Rebecca Moore}
\author[1]{Alexis Boukouvalas}
\author[1]{Pushmeet Kohli}

\affil[*]{Equal contributions}
\affil[1]{Google DeepMind}
\affil[2]{Google}

\begin{abstract}
Unprecedented volumes of Earth observation data are continually collected around the world, but high-quality labels remain scarce given the effort required to make physical measurements and observations. This has led to considerable investment in bespoke modeling efforts translating sparse labels into maps. Here we introduce AlphaEarth Foundations, an embedding field model yielding a highly general, geospatial representation that assimilates spatial, temporal, and measurement contexts across multiple sources, enabling accurate and efficient production of maps and monitoring systems from local to global scales. The embeddings generated by AlphaEarth Foundations are the only to consistently outperform a suite of other well-known/widely accepted featurization approaches tested on a diverse set of mapping evaluations without re-training. We have released a dataset of global, annual, analysis-ready embedding field layers from 2017 through 2024.
\end{abstract}

\begin{document}

\maketitle

\section{Introduction}

Management of global food supplies, public health, and disaster response all start from maps that geographically anchor questions like "which forests pose an unacceptable wildfire risk?" or "where are soybeans grown?". The launch of the first Landsat satellite in 1972 marked the dawn of an era where spaceborne monitoring could serve the interests of global environmental policy-making and provide critical insights into our changing planet \citep{cohen2004landsat}. Over the following decades Earth observation (EO) data became widely available, and streams from both historic and modern EO instruments are now routinely used to create maps that answer questions about the past, present, and future of Earth’s ecosystems and climate \citep{wulder2022fifty}. Nonetheless, advancements in deriving planetary-scale insights from petabytes of satellite imagery and other environmental datasets remain hamstrung by the relative scarcity of ground-based measurements and annotations, and a new problem: the overwhelming volume of geospatial data \citep{tuia2024artificial}. In this work, we introduce a foundational geospatial embedding model that solves fundamental challenges in the institution of mapping through the generation of a universal feature space. The features produced by our model consistently achieve top performance in all application domains tested when compared to other general and even domain specific approaches (Figure \ref{fig:headline}A). This marks a shift from the previous state-of-the-art for which no single approach was dominant.

\section{From sparse labels to maps}
High-quality maps depend on high-quality labeled data, yet when working at global scales, a balance must be struck between measurement precision and spatial coverage. Many global mapping efforts focus on individual ecosystems like forests \citep{hansen2013high}, water \citep{pekel2016high}, tidal wetlands \citep{murray2022high} or other broad legends, e.g., \citep{zanaga2022worldcover, brown2022dynamic}. This simplifies the label collection process, allowing trained interpreters to collect larger volumes at scale at the expense of descriptive power for certain use cases. In the cases where high-quality annotations and/or field

\newpage
\begin{strip}
    \begin{minipage}{\textwidth}
        \includegraphics[width=\textwidth]{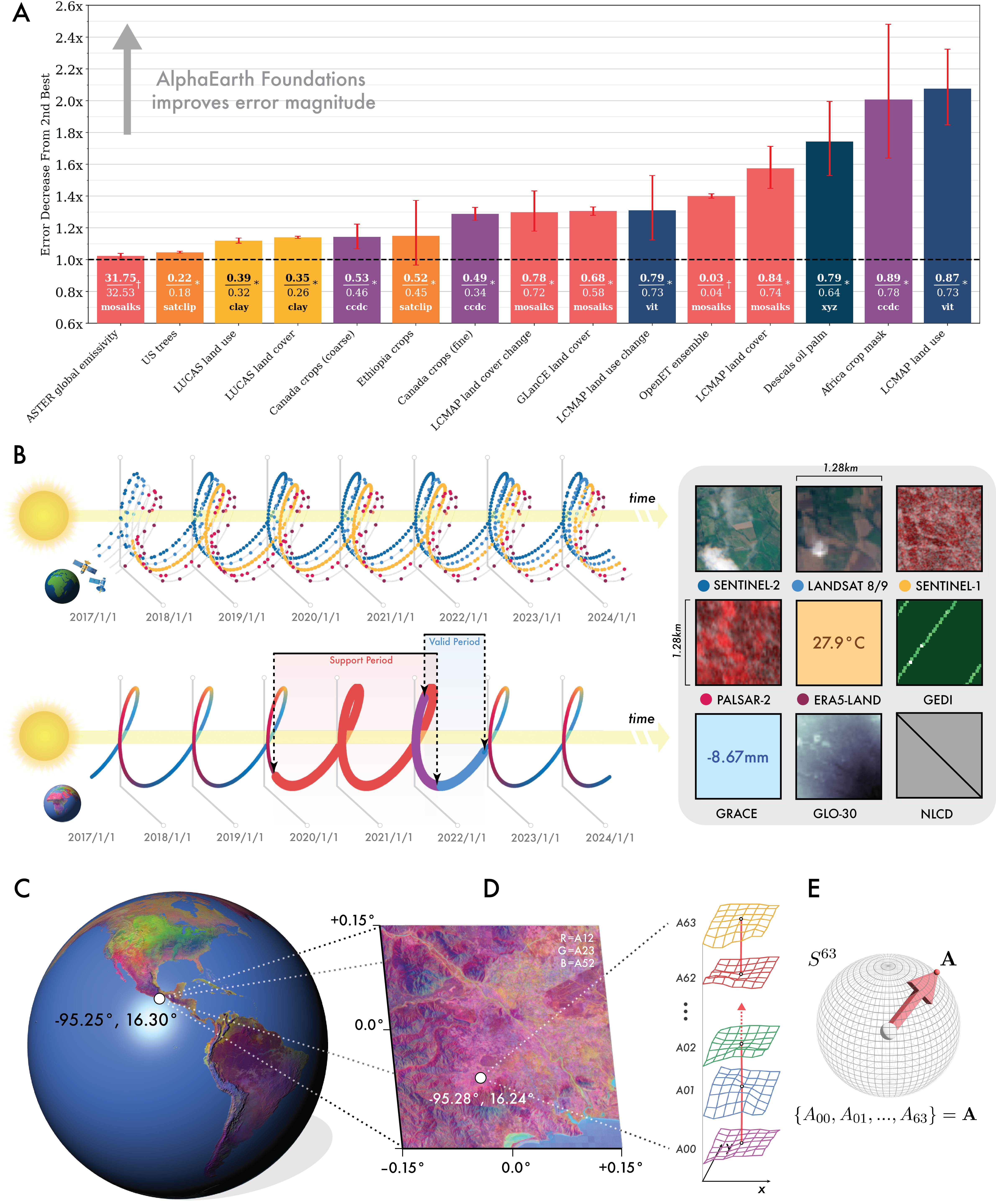}
        \captionof{figure}{\textbf{Embedding fields paradigm.} (A) Error ratios across evaluations from the next-best model/dataset to AlphaEarth Foundations (AEF). Classification errors (bars marked with *) are measured in Balanced Error Rate kappa (BER$\kappa$), and regression errors are measured in MAE$^{-1}$ (bars marked with †). The pair of numbers on each bar indicate balanced accuracy (BA) for classification tasks and MAE for regression tasks, with AEF on top and the next-best model/dataset below. Best-case performance was selected independently for both the next-best model/dataset and for AEF by selecting the most performant method of transfer (kNN k=1, kNN k=3, linear) for each evaluation. For each}
        \label{fig:headline}
    \end{minipage}

     evaluation shown, all available training data or the "max-trial" was used. Error bars indicate the $1\sigma$ best and worst case ratio or $\sim$90\% confidence interval by bootstrapping and k-folds when possible. Evaluations are sorted by increasing mean error ratio. The dashed line represents a ratio of 1 with higher values indicating AEF outperforming the next best solution. (B) AEF reconciles multiple sparse, non-uniformly sampled observation records into a continuous record, regardless of fluctuations in availability. AEF "embedding fields" are a result of static temporal summaries drawn over a conditional "valid period" that need not fully intersect the "support period", where the latter defines the temporal range of the input data. Multiple raster and scalar measurement sources are modeled as sources or targets by AEF. These may be any combination of temporally, geographically, and spatially sparse. In the example shown here, NLCD is not present, and GEDI is available in only a sparse fraction of the spatial context. (C) A view of our global embedding field for the year 2023, note apparent climatic gradients at large scales. (D) AEF produces highly resolved features at 10m$^2$, shown here plotting arbitrary axes in Oaxaca, Mexico. (E) A stack of 64 rasterized AEF layers forms an embedding field, and each individual vector maps to a coordinate on the unit sphere $S^{63}$.
\end{strip}

\noindent
measurements are available, systematic coverage is usually much more localized, e.g., \citep{nagy2021harnessing, lister2020use, dAndrimont2020harmonised}. Accurate and efficient scaling of such highly-detailed yet spatially and temporally sparse data remains an open challenge, e.g., \citep{sun2021research}.

A natural approach to better leveraging sparse observations is to isolate the relevant information content of the feature space used to generate maps. Designed EO features like vegetation indices \citep{zeng2022optical}, best-available-pixel composites \citep{white2014pixel}, 1D harmonics \citep{zhu2014continuous, wilson2018harmonic}, and kernel-based filters \citep{haralick1973textural, lee2017review} power many of the map data products used for policy making, e.g., \citep{zanaga2022worldcover,brown2020lessons, wulder2024development}. When heuristics are carefully chosen, they can offer an efficient mechanism for geographically extrapolating labels and measurements. However, designed features are often noisy, sensor-dependent, and highly region- and application-specific, compounding the challenges inherent to working with satellite imagery and other planetary-scale datasets.

Machine learning has revolutionized fields from biochemistry to natural language understanding. Unsurprisingly, combining disparate EO sources through the use of machine learning has become an active area of research \citep{zhu2017deep, rolf2024mission}. A new generation of geospatial foundation model approaches can be roughly characterized as derivatives of SatMAE \citep{cong2022satmae} or implicit models such as SatCLIP \citep{klemmer2025satclip}. While these approaches represent progress in the application of ML to EO data, none satisfy all of the following key properties: (1) multi-source or multi-modality,  (2) inclusion of time into the modeling framework, or (3) spatial resolution at a precision useful for serving operational mapping use cases. Critically, as we will show, existing learned featurization approaches don’t always outperform designed featurization methods in scarce data regimes.

\section{AlphaEarth Foundations}

AlphaEarth Foundations (AEF) is the only learned EO featurization approach to outperform a representative sample of featurization methods tested across a broad set of sparse data domains (Figure \ref{fig:headline}A), reducing error magnitudes by $\sim$23.9\% ($\sim$1.4x error magnitude reduction) on average while maintaining a best-in-class 10-meter spatial resolution that requires 16x less information per-representation (64 bytes) compared to the next-most compact learned method. We achieve this leap in performance across challenging mapping applications through a number of innovations; namely, we employ an adaptive decoding scheme that considers time and sensor parameters as continuous variables in an implicit decoder with associated losses, a spatially dense information time-bottleneck, time-conditional summarization, and spatially-precise alignment with geotagged text from Wikipedia articles joined with locations from Global Biodiversity Information Facility (GBIF) species occurrence records. To the best of our knowledge, AEF is the first EO featurization approach to support continuous time (Figure \ref{fig:headline}B; see supplemental materials \ref{S2.2.1}). Additionally we introduce a challenging evaluation suite composed of high-quality reference data that attempts to faithfully replicate realistic mapping scenarios. We make our annualized planet-scale feature maps or "embedding field" layers (Figure \ref{fig:headline}C-D) and evaluation suite available under an open license to encourage further exploration and use.

\begin{figure*}
    \centering
    \includegraphics[width=\textwidth]{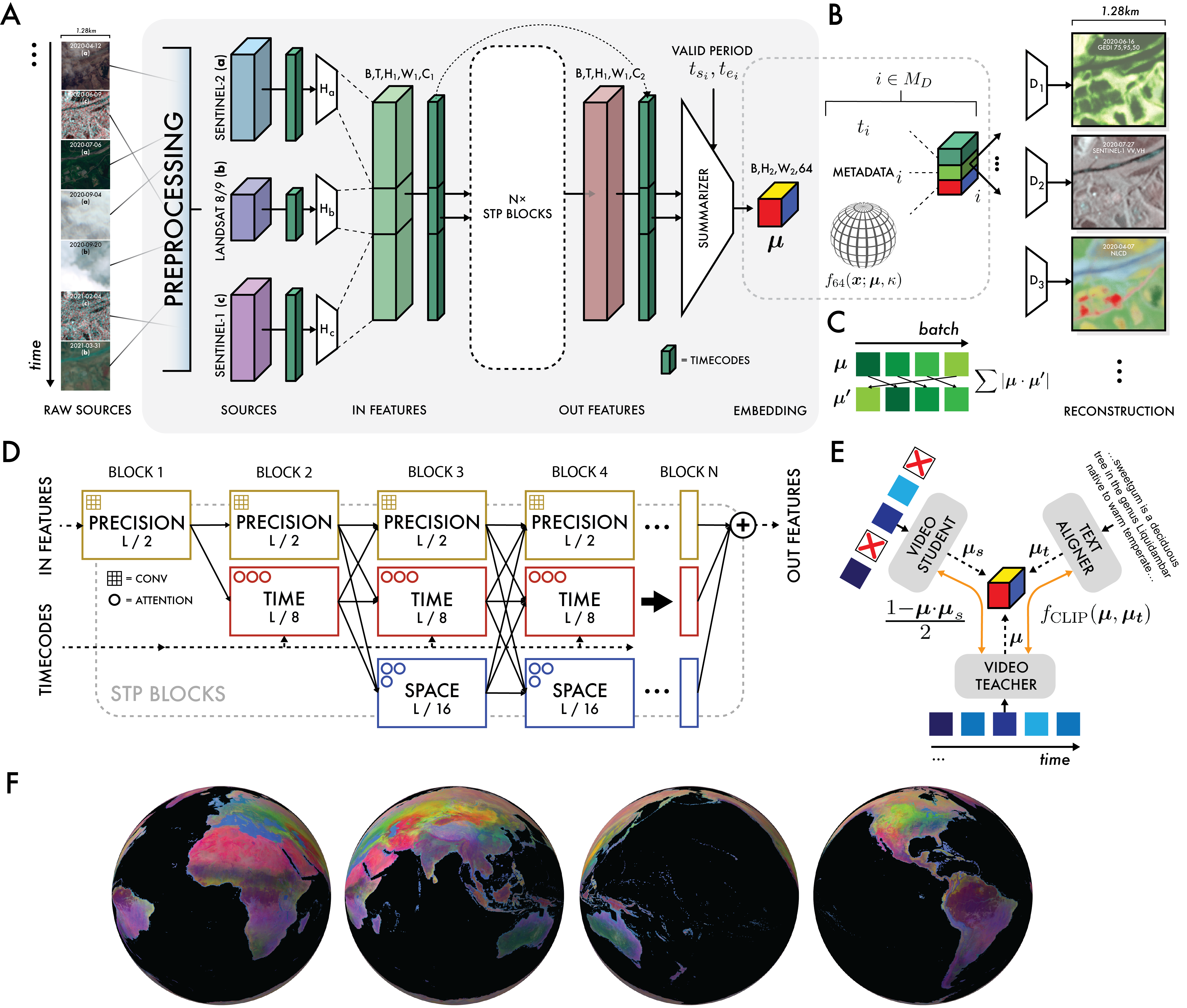}
    \caption{\textbf{AlphaEarth Foundations.} (A) Block diagram of the overall network architecture used for video analysis. Preprocessing converts raw observation data via normalization using global statistics, and acquisition timestamps are converted to sinusoidal timecodes. Individual source encoders transform inputs to the same latent space before entering the bulk of the model. Outputs are summarized using conditional timecodes or "summary periods", unique to each decoded source and contrastive learning task. $\mu$ refers to the embedding outputs of the model. The grey shaded region constitutes the model. (B) Model outputs are treated as the mean direction of a von Mises-Fisher distribution, and decoding proceeds by sampling this distribution, and concatenating it with sensor geometry metadata and a timecode indicating the relative position in the valid period to decode. Decoding proceeds for all sources, with losses dependent on the characteristics of each source (see supplemental materials \ref{S1}). (C) To prevent collapse and improve performance, embeddings are compared to equivalent batch-rotated embeddings using a dot product. The absolute value of this quantity is minimized as a necessary condition for an empirically uniform distribution in $S^{63}$. (D) Block diagram of the model bulk, consisting of simultaneous pathways at different resolutions to maintain efficiency and spatial precision. (E) Contrastive learning between the video teacher and student model, and text encoder. (F) Complete 360° view of 2023 annual embedding field covering Earth's land surface including minor islands over approximately $\pm$ 82°.}
    \label{fig:results_model}
\end{figure*}

AEF is designed to accept $N_i$ frames for $i \in M_E$ input (encoded) data sources with $C_i$ channels resampled to the same spatial resolution, and a millisecond epoch timestamp $t_j, 1 \leq j \leq {\Sigma}N_i$ (Figure \ref{fig:results_model}A). The range of the input timestamps we refer to as the ``support period''. For the purposes of learning or at inference time, we support a pair of conditioning timestamps or ``valid period'' $t_s, t_e$ where $t_s < t_e$  provides a temporal summary of the Earth's surface and climatic activity over $[t_s, t_e)$, even when there is no $t_j$ for which $t_s\leq t_j < t_e$ (interpolation), or when $t_e \leq t_j,  \forall j \in \{1, 2, ..., {\Sigma}N_i\}$ or $t_s>t_j, \forall j \in \{1, 2, ..., {\Sigma} N_i\}$ (extrapolation). These summaries or ``embeddings'' are 64 bytes in size, and each embedding contains information that reproduces the temporal trajectory of variables listed in Table \ref{table:datasources} over the summary period (Figure \ref{fig:results_model}B) using conditional metadata from each source (see supplemental materials \ref{S2.2.1}). By explicitly separating the input intervals from those used for the temporal summary, we can apply AEF to time dependent problems requiring a precise date range without fine-tuning. Embeddings are further constrained to distribute uniformly in $S^{63}$ using a so-called ``batch uniformity'' objective (Figure \ref{fig:results_model}C; see supplemental materials \ref{S2.2.4}).

Our video summarization architecture must simultaneously maintain highly localized representations as well as model long distance relationships through time and space in a computationally efficient way; for this we've designed an encoder termed Space Time Precision or ``STP'' that consists of repeated blocks of three simultaneous operators interleaved with spatial pyramid "exchanges" (Figure \ref{fig:results_model}D) inspired by \cite{wang2020deep} but more efficiently utilizing learned resampling stages. Given a square input of $L$ pixels a side, each block consists of a $\frac{1}{16}L$ "space" operator following ViT-like spatial self-attention \citep{dosovitskiy2020image}, a $\frac{1}{8}L$ ``time'' operator utilizing time-axial self-attention, and a $\frac{1}{2}L$ "precision" operator utilizing 3x3 convolutions. Each sequence element in the ``time'' operator is conditioned on the associated input timestamp ($t_j$) after conversion to a sinusoidal timecode. STP blocks terminate with learned Laplacian pyramid rescaling, such that each operator can pass its state to each of the operators in the subsequent block. STP itself terminates with a final learned spatial resampling to the resolution of the "precision" operator. Thus for ${\sum}N_i$ inputs, STP produces ${\sum}N_i$ output feature maps at $\frac{1}{2}L$ pixels resolution. Initial down-scaling to $\frac{1}{2}L$ is performed in the input projectors marked with an "H" in Figure \ref{fig:results_model}A. 

We train a trio of neural network models that work in tandem: a teacher video embedding model with implicit decoders, a student video embedding model sharing the same parameters and architecture as the teacher, and a text alignment model (Figure \ref{fig:results_model}E). We trained $\sim$1B and $\sim$480M parameter variants of AEF, and ultimately proceeded with the smaller variant for improved inference efficiency. We discuss the training set, model training, and architecture in greater detail in supplemental materials \ref{S2}. The results of running inference at scale are ``embedding fields'' tiling Earth's terrestrial surface in approximate 10m$^2$ grids (Figure \ref{fig:results_model}E).

\section{Evaluation in realistic data-scarce scenarios}
To establish the performance of AEF relative to other learned and designed representations, we required an evaluation dataset that included archetypal examples of realistic mapping applications. These include thematic mapping, biophysical variable estimation, and change detection at annual and sub-annual cadences. We found most if not all datasets in existing geospatial benchmark suites provide annotations at an object- or image-level rather than pixel-level, rely on labels sampled from existing (machine-generated) datasets, require running the benchmark analysis using provided source imagery, have limited geographic coverage, and/or do not provide sufficient spatial precision or temporal information, e.g., \citep{lacoste2023geo, bountos2023fomo}, limiting their value in assessing practical use.

To address the need for high-quality labeled datasets that can be used to simulate low-shot (i.e., tens to hundreds of samples) performance in data-scarce regimes requiring precise map outputs, we developed a set of 15 evaluations sourced from 11 openly available datasets. These datasets were selected to represent archetypal classification, regression, and change detection use cases, with all datasets directly linked to real-world products and applications, including land use/land cover mapping and change detection, crop type mapping at different hierarchies, tree genera and plantation classification, and estimating evapotranspiration and emissivity  at national to global scales (Table \ref{table:evals}). For each evaluation dataset, we selected a balanced number of training samples from each class or equally-spaced partition, with the number of samples determined based on the minimum class size (Table \ref{table:evals}, Max Trial Size) and the remainder of the dataset reserved for testing. We then assess performance across a suite of trials designed to test very-low shot training sizes with $n$ samples per class ($n=1$, $10$, max) via transferal methods with minimal parameters: k-nearest neighbors, and linear layers fit to the features (see supplemental materials \ref{S4}). The ``max-trial'' scenario is meant to represent more realistic sparse dataset sizes (hundreds as opposed to thousands or millions of points), whereas the ten and one shot trials are meant to evaluate performance given extreme data sparsity.

\begin{table*}
\centering
\footnotesize
\begin{tabular}{p{0.19\linewidth} p{0.17\linewidth} p{0.12\linewidth} p{0.10\linewidth} p{0.10\linewidth} p{0.07\linewidth} p{0.07\linewidth}}
\toprule
{\bf Dataset Name} & {\bf Domain} & {\bf Evaluation type} & {\bf Geographic Extent} & {\bf Temporal Cadence} & {\bf Max Trial Size  (n)} & {\bf Total Sample Size  (n)}\\
\midrule
\multirow{4}{\linewidth}{LCMAP \citep{brown2020lessons, pengra2023lcmap}} & Land cover & classification (6 classes) & CONUS  & Annual & 300 & 26,510 \\
\cmidrule{2-7}
 & Land use & classification (6 classes) & CONUS  & Annual & 300 & 26,513 \\
\cmidrule{2-7}
 & Land use change & change detection (binary) & CONUS  & Annual & 150 & 991 \\
\cmidrule{2-7}
 & Land cover change & change detection (binary) & CONUS  & Annual & 300 & 2,320 \\
\midrule
 \multirow{2}{\linewidth}{LUCAS \citep{dAndrimont2020harmonised, toth2013lucas}}  & Land cover & classification (15 classes) & Europe & Single-date & 300 & 203,569 \\
\cmidrule{2-7}
 & Land use & classification (40 classes) & Europe & Annual & 300 & 226,858 \\
 \midrule
 GLaNCE \citep{stanimirova2023global} & Land cover & classification (11 classes) & Global & Annual & 300 & 34,885 \\
\midrule
 Africa crop mask  \citep{kerner2024accurate, kerner2024dataset} & Crop type & classification (4 classes) & Sub-Saharan Africa & Annual & 200 & 2,556 \\
\midrule
\multirow{4}{\linewidth}{Canada crops \citep{AAFC2024dataset}}  & Fine crop type & classification (24 classes) & Canada & Single-date & 75 & 14,566 \\
\cmidrule{2-7}
 & Coarse crop type & classification (9 classes) & Canada &  & 68 & 16,079 \\
\midrule
 Ethiopia crops \citep{blasch2024ethiopian} & Crop type & classification (4 classes) & Ethiopia & Annual & 49 & 2,530 \\
\midrule
 US trees \citep{gbif2024occurrence} & Tree genera & classification (39 classes) & United States & Single-date & 300 & 45,382 \\
\midrule
 Descals oil palm  \citep{descals2024dataset, descals2021high} & Palm plantations & classification (3 classes) & Global & Annual & 200 & 17,477 \\
\midrule
 OpenET ensemble  \citep{melton2022openet} & Evapotranspiration & regression (continuous) & Western US & Monthly & 300 & 35,683 \\
\midrule
 ASTER GED \citep{hulley2015aster, nasa2014aster} & Surface emissivity & regression (continuous) & Global & Annual & 200 & 17,636 \\
\bottomrule
\end{tabular}
\caption{\textbf{Overview of evaluation datasets.} For each dataset, we indicate which mapping domain it represents, its geographical coverage, and its temporal cadence. All datasets are permissively licensed and have been modified to ensure a minimum spacing of 1.28km between sample points and guarantee balanced sample sizes across classes or bins. Maximum trial size (n) for each evaluation is noted below. Evaluation results are reported in Balanced Accuracy and R$^2$ for classification / change detection and regression respectively.}
\label{table:evals}
\end{table*}

We used this set of evaluations to compare AEF with a representative set of domain-specific baselines specifically designed for Earth observation applications, including three designed featurization approaches: CCDC \citep{zhu2014continuous, gorelick2023global}, MOSAIKS \citep{rolf2021generalizable}, and composites \citep{qiu2023evaluation}, and three learned featurization approaches: SatCLIP \citep{klemmer2025satclip}, Prithvi \citep{jakubik2023foundation}, and Clay \citep{Clay2024model}. We also include three controls: spatial coordinates (XY), coordinates and elevation (XYZ), and a ViT (Vision Transformer) pre-trained on ImageNet \citep{dosovitskiy2020image, deng2009imagenet}. Where applicable, baselines were provided with identical inputs to AEF and baseline hyperparameters were tuned to maximize performance on our evaluation suite (see supplemental materials \ref{S5}).

Our evaluations showed AEF consistently outperforms both designed and learned featurization methods in all trial settings. AEF reduced error magnitudes overall by $\sim$23.9\% on average when compared to the next-best approach and method of transfer in the max-trial setting (Figure \ref{fig:headline}A). For ten-shot trials, AEF reduced error magnitudes by $\sim$10.4\% on average compared to the next-best approach, and for one-shot trials AEF reduced error magnitudes by $\sim$4.18\%. We show quantitative and qualitative results for select evaluations in Figure \ref{fig:qual_quant} (and see supplemental materials \ref{S6} for full quantitative results). The next-best approach varies across evaluation dataset and method, indicating both non-uniform progress and that AEF unlocks progress in historically challenging mapping scenarios. We discuss these results and the effect of scaling training data in further detail below.

\section{Thematic mapping}
Thematic mapping or "semantic segmentation" refers to spatially-dense discrete classification over an area. We group 11 classification evaluations into thematic mapping applications including land use, land cover, crop detection, crop type, and species distribution mapping. These classification datasets vary in their number of classes, complexity of semantics they represent, and their summary periods, i.e., instantaneous observations versus persistence over a reference period (Table \ref{table:evals}). Assessing max-trial performance in Figure \ref{fig:headline}A, we find that AEF achieves the greatest error reductions for evaluations over annual periods: e.g., LCMAP land cover, Descals oil palm, Africa crop mask, and LCMAP land use. Other

\newpage
\begin{strip}
    \begin{minipage}{\textwidth}
        \begin{center}
            \includegraphics[width=0.93\textwidth]{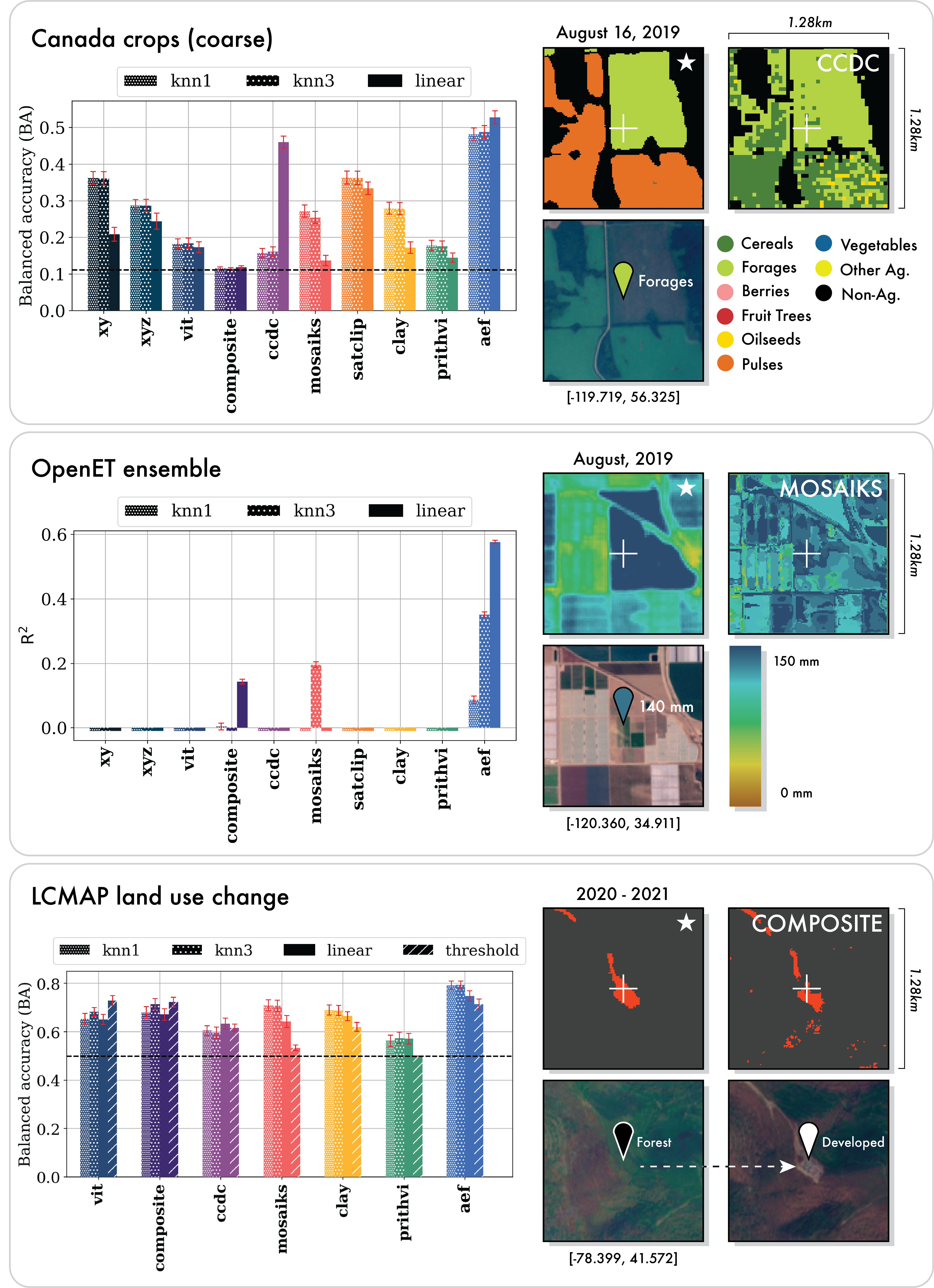}
        \end{center}
        \captionof{figure}{\textbf{Detailed quantitative and qualitative results from select evaluations.} The black dotted-line indicates random chance for classification evaluations. Error bars indicate $1\sigma$ accuracy / R$^2$ or $\sim$68.27\% confidence interval by bootstrapping and k-folds when possible. Most baselines are completely unable to explain evapotranspiration from our testing, achieving a negative $R^2$, and so the}
        \label{fig:qual_quant}
    \end{minipage}

   OpenET evaluation omits results for the majority of transfer/baseline combinations. To the right of each chart we show a qualitative comparison of AEF (starred, top left) to the next-best model or dataset (top right) on their respective most performant method of transfer on a test example at 10m$^2$ resolution, and cloud-free Sentinel-2 L1C RGB image/composites (bottom row) of the location that is not necessarily coincident with timing of the example but is at least using imagery from the same year. We note that AEF demonstrates improved spatial coherence without loss of spatial precision.
\end{strip}

\noindent
 than Ethiopia crops, all thematic mapping evaluations had > 1.0x reductions in error within the  $\sim$90\% confidence interval. AEF's consistent performance across these diverse evaluations suggests a degree of generality that was previously not possible even with higher-dimensional learned embeddings.

\section{Estimating biophysical variables}
The estimation of biophysical variables goes beyond problems of semantics and perception. Effectively extrapolating sparse measurements of properties not easily observed in satellite or other overhead imagery stands to benefit applications from greenhouse gas emissions to the heating/cooling impact of crops. We consider two biophysical variables: emissivity, which is a unitless measurement of surface radiation, and evapotranspiration, which characterizes loss of water to the atmosphere from Earth’s land surface. In the max-trial setting, we find that all baselines were able to explain emissivity for some method of transfer with $R^2$ > 0.5 except for xy, xyz, and CCDC. AEF had the highest $R^2$ ($0.72\pm0.00$), followed by MOSAIKS ($0.69\pm0.00$). In characterizing evapotranspiration, AEF demonstrates a significant departure from the other baselines tested being the only method with $R^2 > 0.2$, achieving $R^2 = 0.58\pm0.01$ (Figure \ref{fig:qual_quant}). We note that the two baselines with explanatory power in this evaluation, composites and MOSAIKS, are simple transformations from raw satellite data, indicating a gap in applicability of both learned and designed featurization approaches prior to AEF.

\section{Change detection}
Responses to natural and man-made disasters, illegal logging, and other emergent phenomena rely on effective and timely regional monitoring. We consider two approaches to embedding-based change detection: \textit{direct classification of change}, which treats change between two summary periods as a binary label and trains the same supervised models used above, and \textit{unsupervised change detection}, which characterizes a continuous magnitude of deviation from an expected value and thresholds this to generate a change mask (see supplemental materials \ref{S4.1}). Our change evaluations are a variant on the LCMAP labels used for thematic mapping that combines labels from different years to produce a binary label indicating whether or not a change in use or cover has occurred. For comparisons we omit the XY and XYZ controls and SatCLIP baseline as these are time-invariant.

In the max-trial setting with direct supervision of change, we find that AEF's performance exceeds performance of other models and datasets achieving $78.4\%\pm1.11$ balanced accuracy (BA) (linear) and $79.3\%\pm1.67$ BA (kNN, k=3) on the land cover and land use evaluations respectively. The next-best baseline achieved $72.0\%\pm1.28$ BA (MOSAIKS, kNN, k=3) and $71.5\%\pm2.33$ BA (composite, kNN, k=3) respectively. In the max-trial setting with unsupervised thresholding, AEF exceeds performance of all other baselines when detecting land cover change (Figure \ref{fig:qual_quant}), but not land use change, achieving $71.3\%\pm1.14$ BA and $71.4\%\pm2.08$ BA compared to $67.0\%\pm1.28$ BA (ViT) and $72.9\%\pm1.97$ BA (ViT) respectively, suggesting the value of supervision for this use case.

\begin{figure*}
    \centering
    \includegraphics[width=\textwidth]{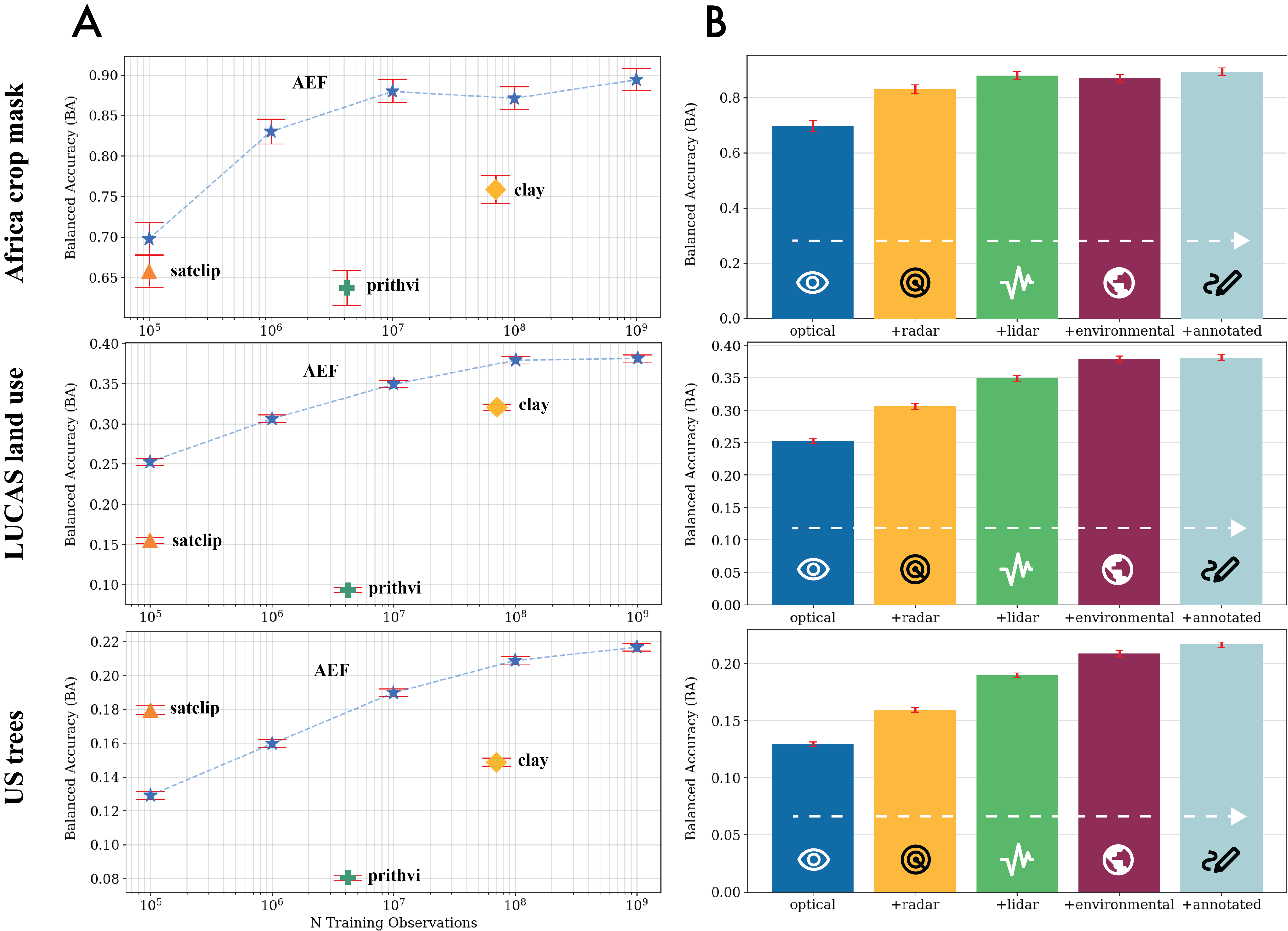}
    \caption{\textbf{Effects of scaling.} Error bars indicate $1\sigma$ accuracy / $R^2$ or $\sim$68.27\% confidence interval by bootstrapping and k-folds when possible. (A) BA as a function of training examples in AEF for select evaluations compared to other learned featurization approaches. AEF generally outperforms other approaches when trained on the same number of unique observations or fewer. From published documentation, SatCLIP uses 100k observations, Prithvi 4.2M observations, and Clay 70M. (B) The effect of compounding training targets on BA for select evaluations. All BA differences for each additional source group are significant for $\alpha=5\%$, though saturation effects are apparent following the LiDAR or Environmental source group for some evaluations.}
    \label{fig:ablate}
\end{figure*}

\section{Scaling source data quantity and type}
The AEF training dataset includes over 3 billion observations across nine different gridded data sources and one unstructured text source, and represents approximately 1.1\% of Earth's land surface area (see supplemental materials section \ref{S2.1}). We find that increasing the number of unique observations used to train AEF leads to more performant embeddings (Figure \ref{fig:ablate}A). For some evaluations (LUCAS land use, Africa crop mask), performance saturated between 100 million and 1 billion observations, whereas for others the saturation point was not obviously reached (US trees). AEF performance generally exceeds that of approaches trained with an equivalent number of observations across all evaluations, and always outperforms other evaluations with the full training set. A noteworthy outlier is US trees for which AEF requires an additional $\sim$100x observations compared to SatCLIP, which we speculate is related to AEF receiving no coordinate information, therefore requiring more examples to learn climate gradients.

We hypothesized that the number of distinct data sources and observation modalities used in training would positively correlate with model performance. To test this, we categorized the sources into the following groups: Optical (Sentinel-2, Landsat 8/9), Radar (Sentinel-1, PALSAR2), LiDAR (GEDI), Environmental (GLO-30, ERA5-Land, GRACE), and Annotated (NLCD, Wikipedia) and iteratively added additional groups to training. We find AEF the most performant when trained on the full set of source groups, though with diminishing returns as additional groups are added (Figure \ref{fig:ablate}B).

\section{Global embeddings dataset}
To facilitate usage of AEF by EO practitioners, we have produced a collection of annual embedding summaries generated by AEF and hosted it as an image dataset on Google Earth Engine \citep{google2025embeddings}. For many use-cases we expect these annual embedding fields to revolutionize mapping workflows that typically require large training datasets, compute intensive models, and custom inference systems to apply those models. To further minimize compute and storage overhead, we quantize the 32-bit floating point embeddings generated by AEF to 8 bits, resulting in an 4x reduction in storage with negligible impact on performance (see supplemental materials \ref{S8} for additional details on inference and quantization). 

\section{Conclusions}
AlphaEarth Foundations (AEF) combines a multitude of diverse geospatial observation records into a time-continuous embedding space by precisely modeling temporal dynamics and relationships across sources. By separating information pertinent only to the act of measurement from the mutual information across all sources, we are able to compactly describe Earth's surface properties while maintaining robustness to the noise and sparsity inherent to Earth imaging missions. 

Our findings indicate that AEF consistently outperforms designed and learned featurization approaches in relatively sparse data regimes and that AEF embeddings are broadly applicable to a diverse range of fields such as biodiversity, ecology and agriculture. For these fields, obtaining maps to model both spatial and temporal changes efficiently is of key importance even when large annotation corpora are not available. As new measurement platforms are launched, others decommissioned, and the accelerating pace of observational data collection pushes forward, we believe it is critical to support the community of applied scientists and practitioners deriving the insights about our planet that inform decision-making and policy action. With AlphaEarth Foundations, we introduce a solution to accurately and generally extrapolate annotations and field measurements to the growing archives of Earth observation now, and into the future.

\section{Acknowledgements}
We would like to thank Olaf Ronnenberger for his manuscript review and feedback, Carlos Guerra for his contributions to our dataset production infrastructure and manuscript review, Katelyn Tarrio for her assistance with the GLaNCE dataset, Maxim Neumann for his assistance preparing an earlier evaluation dataset, Sai Cheemalapati for his assistance with serving our data, Jonathan Thompson and Xiaojie Gao at Harvard Forest for testing an earlier version of our embedding fields, Eric Smit for additional proofreading, and the memory of The Moose for her cat support over the lifetime of this project, as best she could offer it.

\section{Author contributions}
Conceptualization: C.F.B., M.K., V.J.P., E.Sh., O.G., A.B., Methodology: C.F.B., M.K., V.J.P., W.J.R., M.S., E.Sh., O.W., Software: C.F.B., M.K., V.J.P., W.J.R., M.S., C.Z., E.Sh., E.L., N.G., S.A., A.B., Validation: C.F.B, M.K., V.J.P, W.J.R., M.S., C.Z., E.L., O.W., E.Sc., S.A., O.G., A.B., Formal analysis: C.F.B., M.K., V.J.P., W.J.R., M.S., C.Z., E.Sh., Investigation: C.F.B., M.K., V.J.P., W.J.R., C.Z., E.Sh., A.B., Resources: C.F.B., V.J.P., W.J.R., S.A., E.Sc., S.A., O.G., R.M., A.B., Data Curation: C.F.B., M.K., V.J.P., W.J.R., C.Z., S.I., N.G., A.B., Writing - Original Draft: C.F.B., M.K., V.J.P., M.S., E.Sh., O.W., Writing - Review \& Editing: C.F.B., M.K., V.J.P., W.J.R., E.Sh., E.L., O.W., L.L.Z, O.G., A.B., P.K., Visualization: C.F.B., V.J.P., M.S., A.B., Supervision: C.F.B., V.J.P., E.Sc., S.A., O.G., R.M., A.B., P.K., Project administration: C.F.B., V.J.P., L.L.Z., S.A., E.Sc., S.A., O.G., A.B.

\section{Data and Materials Availability}
We release annualized embedding field layers from 2017-2024, our suite of evaluation datasets, and the locations of our training sample sites under an open license for further exploration and applied use. 

AlphaEarth Foundations was trained using publicly available data from the Copernicus Program, the United States Geological Survey (USGS), the National Aeronautics and Space Administration (NASA), the Japan Aerospace Exploration Agency (JAXA), and the Copernicus Climate Change Service (C3S) of the European Commission and the European Centre for Medium-Range Weather Forecasts (ECMWF).

Our evaluation datasets were derived from publicly available data including: LCMAP CONUS Reference Data Product 1984-2021 land cover, land use and change process attributes, from the United States Geological Survey, which is in the public domain; LUCAS Harmonized (Theoretical Location, 2006-2018) V1, from the Joint Research Centre of the European Commission, whose use is governed by the Creative Commons Attribution 4.0 International License (CC-BY); GLanCE: A Global Land Cover Training Dataset from 1984 to 2020, from Boston University Global Land Cover Estimation (GLanCE), whose use is governed by the Creative Commons Attribution 4.0 International License (CC-BY); Comparison of Cropland Maps Derived from Land Cover Maps in Sub-Saharan Africa, whose use is governed by under the Creative Commons Attribution 4.0 International License (CC-BY); Canadian AAFC Annual Crop Inventory from the Canadian AAFC (Agriculture and Agri-Food Canada) whose use is governed under the Open Government Licence Canada; Ethiopian Crop Type 2020, whose use is governed by licensed under the Creative Commons Attribution 4.0 International License (CC-BY); iNaturalist whose use is governed by a Creative Commons Attribution Non-Commercial 4.0 License (CC-BY-NC); Global mapping of oil palm planting year from 1990 to 2021, whose use is governed by the Creative Commons Attribution 4.0 International License (CC-BY); OpenET Ensemble Monthly Evapotranspiration v2.0 from OpenET, Inc. whose use is governed by the Creative Commons Attribution 4.0 International License (CC-BY); and AG100: ASTER Global Emissivity Dataset 100-meter V003, which is available at no charge and with no restrictions on reuse, sale or redistribution.

Our training site selection was informed by the RESOLVE Ecoregions 2017 dataset, whose use is governed by the Creative Commons Attribution 4.0 International License (CC-BY); the Allen Coral Atlas (ACA) - Geomorphic Zonation and Benthic Habitat - v2.0, whose use is governed by the Creative Commons Attribution 4.0 International License (CC-BY); the Murray Global Intertidal Change Classification, whose use is governed by the Creative Commons Attribution 4.0 International License (CC-BY).

\section{Supplementary Material}
\beginsupplement

\section{Data sources and preprocessing}\label{S1}
AlphaEarth Foundations (AEF) was trained on both image and text data sources representing a diversity of imaging modes and measurement spaces (Table \ref{table:datasources}). All raster data sources were sampled from the Earth Engine Data Catalog \citep{gorelick2017google}, and we prioritized publicly available, moderate-resolution datasets covering the period 2017 to present. As text served more as an auxiliary task, only English Wikipedia was used for sourcing text data. Source dataset characteristics and sensor-specific preprocessing steps are described in the following sections.

We reproject all raster data to Universal Transverse Mercator (UTM) coordinates followed by spatial resampling to 10 m resolution using bilinear interpolation. We rescale pixel values to zero mean and unit variance using per-band statistics computed on the pretraining dataset, clipping values with magnitude larger than 6 standard deviations post-scaling. We do not perform any masking at the input stage, instead using the valid data masks to exclude masked pixels when computing the loss. Unless stated otherwise, at bare minimum, millisecond acquisition timestamps were saved as metadata used during reconstruction.

\begin{table*}
\footnotesize
\centering
\begin{tabular}{p{0.07\linewidth} p{0.13\linewidth} p{0.16\linewidth} p{0.31\linewidth} p{0.13\linewidth} p{0.05\linewidth}}
\toprule
{\bf Type} & {\bf Dataset} & {\bf Product} & {\bf Bands} & {\bf Resolution (m)} & {\bf Usage} \\
\midrule
Optical & Sentinel-2 & L1C & B2 (Blue), B3 (Green), B4 (Red), B8 (NIR), B11 (SWIR) & 10, 20, 60 & input, target \\
\midrule
Optical, Thermal & Landsat-8, Landsat-9 & L1C & B2 (Blue), B3 (Green), B4 (Red), B5 (NIR), B6 (SWIR), B8 (Panchromatic), B10 (Thermal) & 15, 30, 100 & input, target \\
\midrule
C-band SAR & Sentinel-1A, Sentinel-1B & GRD & VV, VH, HH, HV, angle & 10 & input, target \\
\midrule
L-band SAR & ALOS PALSAR ScanSAR & Level 2.2 & HH, HV, lin & 25 & target \\
\midrule
Elevation & Copernicus DEM & GLO-30 & DEM (elevation) & 30 & target \\
\midrule 
LiDAR & GEDI & L2A & Relative height metrics (rh*) & 25 & target \\
\midrule
Climate & ERA5-Land & Monthly aggregates & total precipitation (sum, min, max), air temperature 2m (and min, max), dewpoint temperature 2m (and min, max), surface pressure (and min, max) & 11132 & target \\
\midrule
Gravity fields & GRACE & Monthly mass grids & equivalent liquid water thickness & 11132 & target (@50\%) \\
\midrule
Land cover & National Land Cover Database & NLCD 2019, 2021 & landcover & 30 & target (@50\%) \\
\midrule
Text & Wikipedia & geocoded articles & text embeddings & N/A & target \\
\midrule
Text & GBIF & Research-grade obs & text embeddings (class, genus, and species) & N/A & target \\
\bottomrule
\end{tabular}
\caption{%
AlphaEarth training data sources. Data sources were selected to represent a diversity of measurement spaces, resolutions, and temporal refresh rates. All data sources were used as targets, only input data sources are required at inference-time to generate embedding fields.
}
\label{table:datasources}
\end{table*}

\subsection{Sentinel-2 (optical)}\label{S1.1}
Sentinel-2 is an optical remote sensing mission from the Copernicus Program that collects moderate spatial resolution (10 m to 60 m) multi-spectral imagery over land \citep{drusch2012sentinel}. Sentinel-2A was launched in June of 2015, with Sentinel-2B to follow in March of 2017. With a two-satellite constellation, Sentinel-2 is able to image the Earth once every 5 days at the equator.

We sample imagery from the Sentinel-2 Level-1C (L1C) collection (COPERNICUS/S2\_HARMONIZED), which has been processed to Top-Of-Atmosphere (TOA) reflectance \citep{gascon2017copernicus}. All images are processed in their source UTM projection and datatake identifiers are used to remove duplicate observations at the boundaries of Sentinel-2 tiles. Given additive storage requirements for each new band selected, we select a subset of bands to use for analysis, specifically the blue, green, red, near-infrared, and shortwave-infrared bands: "B2", "B3", "B4", "B8", "B11". To ensure a more even distribution of reflectance, we transform Sentinel-2 and Landsat 8/9 pixel intensities using the following formula (prior to standard scaling):

\begin{equation} \label{eq:1}
    s(x) = \frac{\log(x + 1)}{10}
\end{equation}

where x is the source pixel intensity.

We include the Cloud Score+ \citep{pasquarella2023comprehensive} "cloud score" (cs) band as a mask with each Sentinel-2 image, binarized to 0 / 1 by thresholding at 0.5. Mask information is only used during training, no input compositing or masking is performed, and mask information is not provided to the model at inference time. 

\subsection{Landsat 8 \& 9 (optical, thermal)}\label{S1.2}
The Landsat Program, a joint initiative of the United States Geological Survey (USGS) and National Aeronautics and Space Administration (NASA), has provided detailed, synoptic depictions of the Earth's surface for over fifty years \citep{wulder2022fifty}. Landsat 8 was launched in February 2013 \citep{loveland2016landsat} with Landsat 9 to follow in September 2021 \citep{masek2020landsat}. These satellites both carry separate optical and thermal instruments, and resulting images include a 15-meter panchromatic band, eight 30-meter optical bands, and two 100-meter thermal bands. Individual Landsat satellites have a revisit time of 16 days, and an 8-day revisit is achieved when two satellites are operating concurrently. Landsat provides another high-quality multi-spectral optical record that is complementary to those acquired by the Sentinel-2 mission, providing additional image frames, as well as the addition of thermal information.

We exclusively use Collection 2 Tier 1 TOA imagery (i.e., "LANDSAT/LC08/C02/T1\_TOA", "LANDSAT/LC09/C02/T1\_TOA"), which have the highest available data quality (terrain corrected, well-calibrated radiometry, intercalibrated across sensors), and we remove ascending (nighttime) imagery by filtering based on sun angle metadata. Acquisitions were deduplicated based on time proximity, i.e., at row overlaps, preference is given first to the image with the closest UTM central longitude, then the most recent. As with Sentinel-2, we select a subset of Landsat bands, specifically the Blue, Green, Red, NIR, SWIR, RGB Panchromatic, Thermal IR bands: "B2", "B3", "B4", "B5", "B6", "B8", "B10". We also include the FMASK pixel quality bitmask ("fmask"). All optical bands are log-transformed. The "fmask" value is set to 1 when the pixel is not "dilated cloud" or "cloud shadow", and 0 otherwise based on the value of the QA\_PIXEL band. Mask information is only used for training.

\subsection{Sentinel-1 (C-band SAR)}\label{S1.3}
Sentinel-1 is the Copernicus Program’s C-band Synthetic Aperture Radar (SAR) mission \citep{torres2012gmes}. Sentinel-1 instruments are designed to collect dual-polarized observations with several different imaging modes. Notably SAR instruments are water vapor (cloud) penetrating, and offer consistent ground measurements in the tropics or other persistently cloud areas. The Sentinel-1 constellation consists of two satellites, Sentinel-1A and Sentinel-1B, which were launched in April 2014 and April 2016, respectively \citep{potin2016sentinel}. The Sentinel-1B mission ended in December 2021 due to a power system failure, resulting in incomplete coverage (and illustrating challenges of continuity when working with Earth observation records).

We use Ground Range Detected (GRD) images ("COPERNICUS/S1\_GRD"), which have been processed using the Sentinel-1 Toolbox to generate a calibrated, ortho-corrected product. We select images acquired in the Interferometric Wide Swath (IW) instrument mode and include both ascending and descending orbits. We include all available bands: "VV", "VH", "HH", "HV", "angle", though noting that each scene contains only 1 or 2 out of 4 possible polarization bands depending on the instrument's polarization settings (i.e., VV, HH, VV+VH, HH+HV). Processed power values are log-scaled to convert to decibels (dB). The angle band is included on all images and is converted from degrees to radians. 

We additionally mask pixels with values less than -30.0 dB or greater than 10.0 dB. We retain metadata on platform heading and orbital inclination for use as reconstruction metadata. During training, we also introduce random gap artifacts as a form of data augmentation to simulate gaps that sometimes occur between Sentinel-1 scenes by sampling an angle uniformly at random and masking out a line of intensities with random width between 0.5 and 2 pixels at that angle in all Sentinel-1 images in an input sequence. These gaps are in-painted during reconstruction.

\subsection{PALSAR-2 (L-band SAR)}\label{S1.4}
The Advanced Land Observing Satellite-2 (ALOS-2) is the radar satellite operated by the Japan Aerospace Exploration Agency (JAXA) which carries PALSAR-2 (Phased Array type L-band SAR-2), an L-band Synthetic Aperture Radar (SAR) instrument \citep{kankaku2013alos}. L-band is a longer wavelength radar signal with greater ability to penetrate through dense vegetation, compared with the C-band frequency measured by Sentinel-1, which tends to be more sensitive to sparse and low biomass vegetation, e.g., \citep{koyama2022alos}.

We use PALSAR-2 ScanSAR ("JAXA/ALOS/PALSAR-2/Level2\_2/ScanSAR") imagery, which is ortho-rectified and radiometrically terrain-corrected. We include all available bands: "HH", "HV", "LIN". Most images have horizontal polarization ("HH"), vertical polarization ("HV"), local incidence angle ("LIN"), and QA mask ("MSK") bands are always present, though a small subset (<8\%) has only HH. We convert the intensity values from digital numbers (DN) to decibels using:

\begin{equation} \label{eq:2}
    \gamma_0 = 10 * \log_{10}(\mathrm{DN}^2) - 83.
\end{equation}

We rescale local incidence angle (lin) to radians. Observations are deduplicated based on path, and we preserve metadata on Pass Direction and Antenna Pointing for use in reconstruction.

\subsection{ERA5-Land (climate)}\label{S1.5}
As part of the Copernicus Climate Change Service (C3S) of the European Commission, the European Centre for Medium-Range Weather Forecasts (ECMWF) has produced an enhanced global dataset for the land component of the fifth generation of European ReAnalysis (ERA5), referred to as ERA5-Land \citep{munoz2021era5}.The ERA5-Land dataset is intended to provide a consistent view of the water and energy cycles at surface level.

We sample the ERA5-Land Monthly aggregation ("ECMWF/ERA5\_LAND/MONTHLY\_AGGR"), which is a post-processed subset of the full ERA5-Land dataset consisting of monthly statistics \citep{copernicus2019era5}. Specifically, we select total precipitation (sum, min, max), temperature at 2-meters (mean, min, max), dewpoint temperature at 2 meters (mean, min, max), and surface pressure (mean, min max) variables to represent general climatic conditions.

\subsection{GEDI (LiDAR)}\label{S1.6}
The Global Ecosystem Dynamics Investigation (GEDI) is a Light Detection and Ranging (LiDAR) mission launched by NASA to the International Space Station (ISS) in 2018 \citep{dubayah2022gedi}. LiDAR sensors like GEDI use laser pulses to estimate vegetation profiles, which can in turn be used to map canopy height and vegetation biomass, e.g., \citep{potapov2021mapping, campbell2021scaled}.

We use the GEDI L2A Raster Canopy Top Height (Version 2) dataset ("LARSE/GEDI/GEDI02\_A\_002\_MONTHLY") \citep{Dubayah2021l2a} This dataset is a rasterized version of the original Geolocated Elevation and Height Metrics Product (GEDI02\_A) product, which is primarily composed of 100 Relative Height (RH) metrics that describe the heights at which a given energy quantile was received by the GEDI instrument. We sample all relative height bands (RH[0-100]), and we mask out pixels where the ISS was busy or the waveform was bad based on the "degrade\_flag" and "quality\_flag" metadata. GEDI is extremely sparse in space/time compared to our other raster sources, nonetheless we find AEF's reconstruction of the full set of GEDI relative height metrics to have mean absolute error $\sim$3.85m during training, including sampling from the noisy bottleneck.

\subsection{GRACE (gravity fields)}\label{S1.7}
The Gravity Recovery and Climate Experiment (GRACE; launched 2002, decommissioned 2017) and its follow-on mission (GRACE-FO; launched 2018) both consist of a pair of satellites working in tandem to take detailed measurements of Earth's gravity field anomalies \citep{tapley2008gravity, kornfeld2019grace}. These measurements can be used to detect changes in the distribution of water across the planet and estimate terrestrial water storage \citep{landerer2012accuracy}. We considered the inclusion of GRACE an extreme test of the flexibility of our method, and were pleased to note that there was no significant negative impact on the loss or reconstruction quality of other sources.

We use GRACE Monthly Mass Grids Release 6.1 Version 3 - Global Mascons ("NASA/GRACE/MASS\_GRIDS\_V03/MASCON\\
\_CRI"). This dataset was derived from GRACE and GRACE-FO and processed at JPL using the Mascon approach (RL06.1Mv03) and an additional Coastal Resolution Improvement (CRI) filter to reduce errors across coastlines. We specifically sample the  equivalent liquid water thickness ("lwe\_thickness") band, which represents the total terrestrial water storage anomalies from soil moisture, snow, and surface water (including rivers, lakes, reservoirs, etc.), as well as groundwater and aquifers in units of centimeters. Given the very coarse resolution of the GRACE data (0.5° or about \textasciitilde 55 km at the equator) relative to other moderate-resolution sources, we apply an additional upsampling and downsampling step with bilinear resampling to smooth pixel borders.

\subsection{GLO-30 (topography)}\label{S1.8}
The Copernicus 30-meter global Digital Elevation Model (DEM), referred to as GLO-30, is a Digital Surface Model (DSM) that characterizes the surface of the Earth including buildings, infrastructure and vegetation. The GLO-30 dataset is primarily derived from an existing TanDEM-X DSM dataset (WorldDEM™) infilled on a local basis with other widely used DEMs including SRTM, ALOS, ASTER and TerraSAR-X. It is generally considered the most accurate, up-to-date radar-derived global DEM, e.g., \citep{guth2021lidar, simard2024global}.

We sample the DEM (elevation) band from the GLO-30 dataset ("COPERNICUS/DEM/GLO30"). Slope and aspect are calculated from the DEM after it is reprojected into the local coordinate system (UTM), and decomposed into sine and cosine. The GLO-30 DEM is assumed to be valid over the entire period of our training set, though we note with natural and man-made phenomena this is not universally true.

\subsection{NLCD (land cover)}\label{S1.9}
The National Land Cover Database (NLCD) is a suite of products developed for operational land cover monitoring in the United States. These machine-generated thematic maps are derived from 30 meter multi-season Landsat imagery and rely on a carefully curated training labels, hand-engineered spatial, spectral and temporal features, and classic machine learning (i.e., decision trees), as well as rigorous post-processing and accuracy assessment \citep{wickham2014multi, wickham2023thematic}. We include NLCD in our list of source datasets as a means of testing the value of existing maps as a form of weak supervision. We note that there was no significant negative impact on the loss or reconstruction quality of other sources, and the effect on evaluations was generally positive despite the temporally-static nature of NLCD (see supplement section S7.2 for ablation results).

We use the 2021 release for the year 2021 ("USGS/NLCD\_RELEASES/2021\_REL/NLCD"), and the 2019 release for all other years ("USGS/NLCD\_RELEASES/2019\_REL/NLCD"), and all samples are associated with the nearest mapped year (if sample date falls in between NLCD release years). We select the "landcover" band, which labels 16 land cover classes using a nested hierarchy. As noted in Figure \ref{fig:headline}, NLCD data is not available for all locations globally.

\subsection{Text sources}\label{S1.10}
Towards truly multi-modal (as opposed to strictly multi-source) embeddings, we assume that geocoded text can provide additional context that will help enrich our learned representations. We use two sources for obtaining locations and associated text: Wikipedia and the Global Biodiversity Information Facility (GBIF) species occurrence records. 

\subsection{Wikipedia}\label{S1.11}
We use geolocated articles from Wikipedia to provide text-based information for things like landmarks and other geographic features. We extract all articles with coordinates (using the P625 - coordinate location Wikidata property) from the 2024-04-21 snapshot of Wikipedia, with additional filters on the "globe" property to remove articles with "extraterrestrial" coordinates. We also drop articles with fewer than 100 words (including title and headers) or where more than 25\% of the total words in the article are contained in lists. References, Further Reading, and External Links sections were omitted, and any non-plain-text content was omitted.

\subsection{GBIF}\label{S1.12}
We obtain species occurrence records through GBIF occurrence dataset available through BigQuery \citep{gbif2024occurrence}. We specifically select records in the Plantae, Animalia and Fungi kingdoms for the period 2017-2023. Observations must be available by CC-BY 4.0 or CC0 1.0 license, be labeled as human or machine observations, have a maximum spatial uncertainty of 240 meters, and meet a number of other criteria to remove invalid or otherwise suspicious coordinates, bad date information, and uncertain taxon matches. Post-filtering, we limit our sampling to a maximum of 1000 observations per unique {family, genus, species} observation tags. We export the observation coordinates and timestamp together with the GBIF taxonomy ID for species, genus, family, and potentially also higher taxonomic levels, as well as the coordinate uncertainty to use for sampling corresponding video embeddings. Finally, we match observations with a subset of Wikipedia articles using the GBIF taxon ID property (Wikidata P846) after normalizing the observation’s GBIF ID to the accepted ID for its taxon.

\section{Modeling}\label{S2}

\subsection{Training Dataset}\label{S2.1}
AEF was trained over 8,412,511 video sequences containing interleaved, time-stamped frames from the sources and metadata listed in supplemental materials \ref{S1}. Each frame covered a 1.28 km x 1.28 km (128 x 128 pixel) area projected into the UTM zone of the area's centroid and were not limited in length: all available data was used totalling 3,047,520,515 frames. Video sequences were sourced from 5,145,244 sites, and each site was split into two non-overlapping approximately year-length periods from which two video sequences were drawn. Sequences were omitted for a variety of factors, including missing data in e.g., polar regions and insufficient frames from a particular source e.g., at swath edges. We make these training locations available in \citep{deepmind2025sites}.

\subsubsection{Training site selection}\label{S2.1.1}
Our global AEF training dataset was developed to provide a representative sample of Earth’s terrestrial land surface and near-shore ecosystems, while optimizing for coverage across space, time, and availability of data sources.

\paragraph{Gridded text}
Our sampling strategy prioritizes coverage of locations where we have geocoded text information by first taking a gridded sample that covers these locations. Our final geocoded text dataset includes point locations for GBIF species observations and other geotagged features from Wikipedia (as described in supplemental materials \ref{S1.10}) plus a key to join these locations with associated text embeddings. The locations represented in this dataset inevitably inherit the same sampling biases as the source GBIF and Wikipedia data, i.e., locations tend to be clustered in areas with denser human populations / more urbanized areas. To account for these spatial biases, we do not sample each Wikigeo location individually; instead, we establish a sampling grid such that multiple Wikigeo locations are associated with the same (non-overlapping) image samples. We use 1.28 km x 1.28 km as our base grid cell size and create grids in the native UTM projection of the zone intersected by the sampled points. Our final gridded text dataset includes centroid coordinates for 1,200,099 grid cells, representing 8,777,536 text points.

\paragraph{RESOLVE ecoregions}
The gridded sample does not fully represent the global land surface, so we use the 2017 RESOLVE Ecoregions dataset ("RESOLVE/ECOREGIONS/2017") \citep{dinerstein2017ecoregion} to draw an additional random stratified sample by ecoregion ID. This helps ensure we are sampling across distinct biogeographic assemblages and ecological habitats with uniform preference regardless of total extent. We use the ECO\_ID (n=846) and target 10,000 samples per ecoregion, then cull based on standard 1.28 km minimum distance requirement (i.e., remove sampled points that are too close together and/or too close to gridded text samples). This generated a total of 3,940,224 unique ($x$, $y$) locations based on ecoregion stratification.

\paragraph{Near-shore ecosystems}
We supplement our initial RESOLVE sample, which largely targets terrestrial ecosystems, with additional stratified samples from the Allen Coral Atlas and Global Intertidal Zones datasets to improve representation of near-shore ecosystems. The Allen Coral Atlas ("ACA/reef\_habitat/v2\_0") maps the geomorphic zonation and benthic habitat for the world's shallow coral reefs at 5 m pixel resolution, as well as a global reef extent product that maps additional reef areas unable to be explicitly included in the geomorphic and benthic mapping  \citep{lyons2024new, lyons2020dataset}. We resample the Atlas from 5 meter to 30 meter resolution, then draw a random sample of 5,000 points. After deduplicating, the final supplemental coral sample consists of 4,141 locations. We also add samples from the Murray Global Intertidal dataset \citep{murray2019global, murray2022high, murray2022losses}. The binary layers in this image collection depict tidal flat ecosystems around the global coastline. As with corals, we initially target 5,000 samples, which reduces to a final count of 2,968 intertidal samples after applying our 1.28 km minimum distance criteria.

\begin{figure}
    \includegraphics[width=\columnwidth]{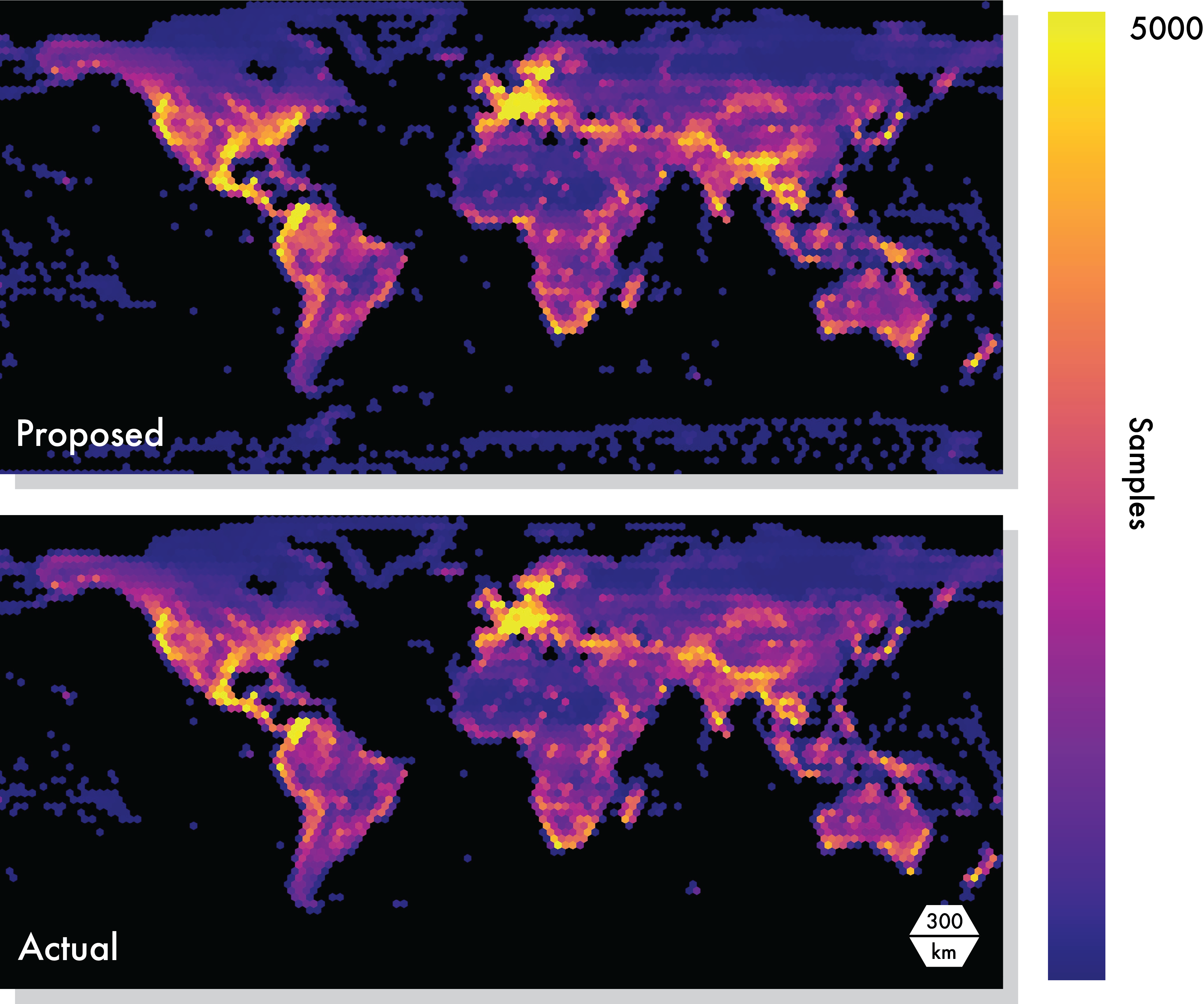}
    \caption{Training site selection. (Top) 5,145,244 unique ($x$, $y$) locations prioritized as potential training sites, (Bottom) final set of training sample locations ($x$, $y$, $t_{start}$, $t_{end}$).}
    \label{fig:training-locations}
\end{figure}

\paragraph{Proposed ($x$, $y$) locations}
We merge the four aforementioned samples (gridded text, ecoregions, corals, and intertidal) into a single dataset. We then reduce the number of samples over the “rock and ice” to n=500, preferentially targeting removal of samples from Antarctica and Greenland (priority to keep examples in high-altitude over high-latitude). We also remove samples over open water, i.e., 128 x 128 pixel image chips that would not sample anything other than offshore water, as we don’t expect the model to learn much from these examples and it reduces redundancy among text points associated with offshore observation transects. After these final filtering steps, our sampled location dataset consists of a total of 5,145,244 unique ($x$, $y$) locations (Figure \ref{fig:training-locations}A).

\subsubsection{Adding time coordinates}\label{S2.1.2}
To ensure that our sample also represents temporal variability in surface properties, we sample two temporal (support) periods per site. This has the added benefit of also effectively doubling the size of our training sample. The general temporal processing strategy is as follows: if there are no Wikigeo points present, a site can select any two non-overlapping periods in the sampling years. If there are points, we pick two periods to maximally allocate point date ranges that intersect those periods such that a point is allocated to only one of the two periods. Upon selection of the periods, we create a final dataset bearing all observations with all period bounds (two periods for each site). From this, we create our final collection of ($x$, $y$, $t_{start}$, $t_{end}$) coordinates, which resulted in a total of 10,203,798 unique rows that moved on to data source collection.

\subsubsection{Training sample}\label{S2.1.3}
\begin{figure}
    \includegraphics[width=\columnwidth]{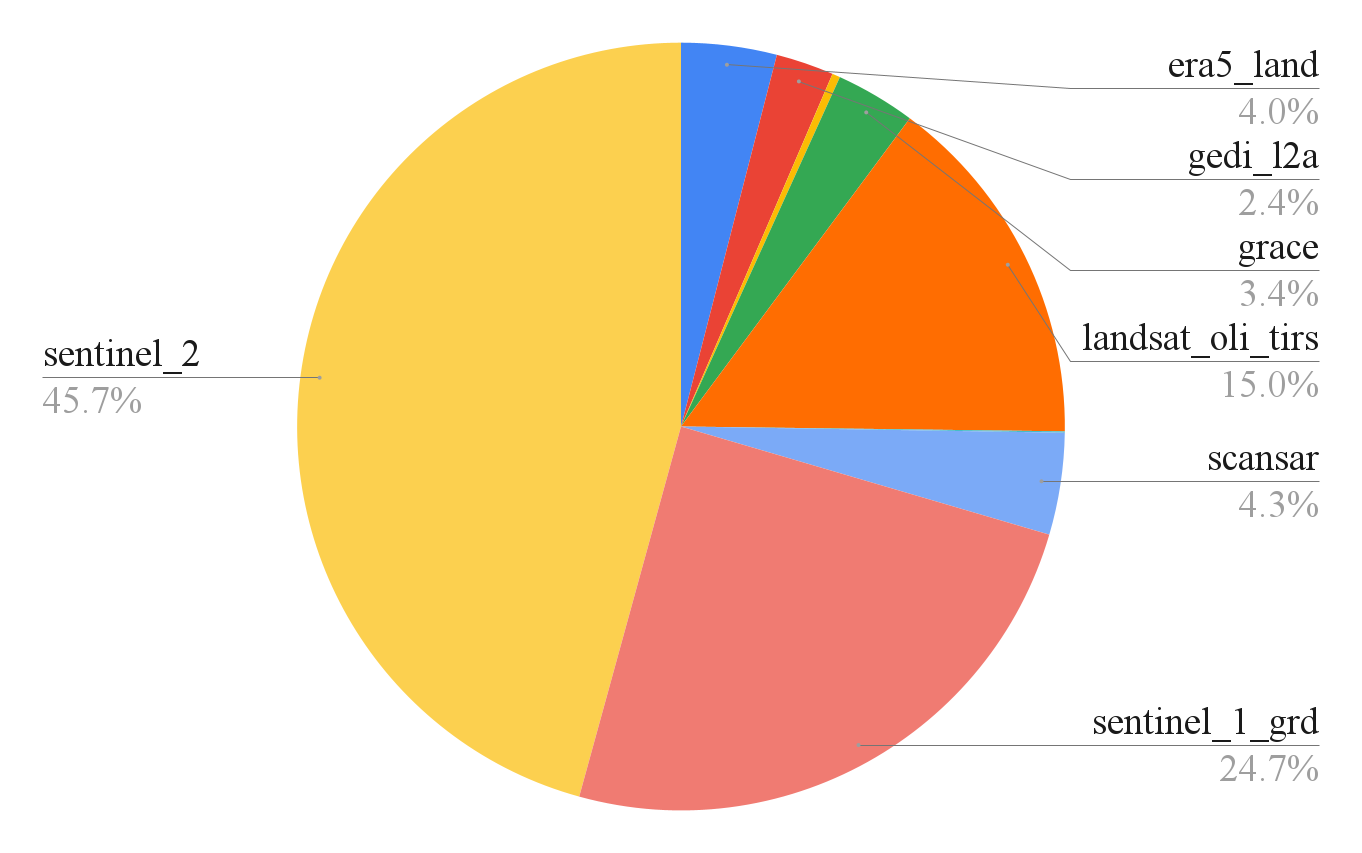}
    \caption{Breakdown of image samples by sensor.}
    \label{fig:supp-sensors}
\end{figure}

Our source data distillation system is designed to sample image sources for ($x$, $y$, $t_{start}$, $t_{end}$)  seed locations while accounting for sensor-specific minutia and geospatial attributes (i.e., projections). Some seeds (and associated imagery sequences) are ultimately dropped due to low (or no) image availability, i.e., some targeted locations may be outside coverage for one of our key inference sensors (Sentinel-1, Sentinel-2, Landsat). In total, we drop 1,791,287 sequences from our initial set of seeds, and ultimately exclude samples from Antarctica due to lack of sufficient Sentinel-1 imagery; while no single source is required for inference, training rows require all input sources to be present so they can be artificially dropped. This resulted in a final pretraining dataset representing 8,412,511 unique ($x$, $y$, $t_{start}$, $t_{end}$) coordinates (Figure \ref{fig:training-locations}B), and consisting of a total of 3,047,520,515 individual image frames (see Figure \ref{fig:supp-sensors} for breakdown by sensor). Training data rows were stored with all pixel data, mask and sensor metadata, and text and geometries intersecting the row physical area in a sharded format designed for rapid loading during training totaling $\sim$6PiB after replication.

\subsection{Training algorithm}\label{S2.2}
\subsubsection{Simulating a continuous observation record} \label{S2.2.1}
It is critical that embeddings should encapsulate temporal dynamics. In practice this means the resulting embedding can differentiate between similar surface conditions with different temporal ordering; for example differentiating between fields with the same crops where the planting happened at different times. 

For time conditional summarization following STP, we produce a temporal summary leveraging time-axial attention pooling based on a single learned query feature derived from the valid period $[t_s, t_e)$ following a conversion to sinusoidal timecodes. This summary is up-sampled to size L using a learned kernel. We then introduce a variational bottleneck with a key innovation: rather than collapse the output spatially, we estimate the mean direction across an $L x L$ grid of von Mises-Fisher (VMF) distributions in $S^{63}$. This bottleneck construction permits a high degree of spatial precision without the further fine-tuning that is typical of other unsupervised embedding models, and provides a mechanism to parameterize the "smoothness" of a given embedding manifold via the VMF concentration.

AEF is trained to conditionally decode embeddings from the bottleneck. For each of $i\in M_D$ decoded sources, a small decoder network accepts an embedding, and a set of conditional metadata specific to that source (Figure \ref{fig:results_model}B). Typically this is at least a sinusoidal timecode representing an instant in the valid period $[t_{s_i}, t_{e_i})$ normalized to $[0, 1)$, though may also contain orbital geometry and metadata that is only relevant to the act of measurement, not the measurement itself. The source decoder network is applied to every location in the output grid to produce an $L x L$ reconstruction $y'_i$, one for each $i\in M_D$ with $C_i$  channels. Interestingly, these decoders have the effect of generating spatially continuous predictions for an arbitrary timestamp (e.g., dense, superresolved LiDAR profiles from GEDI). We update the parameters in the entire network to minimize the error between $y'_i$ and $y_i$, a target source frame randomly selected to intersect the valid period and potentially held out of the inputs. The error metric varies depending on the source, and accounts for spatial misregistration of the instrument, missing data, and $y'_i$ vs. $y_i$ resolution mismatches as is the case when our nominal embedding resolution ($10m^2$) does not match the source's original resolution. We minimize cross-entropy loss for categorical sources, and L1 error for non-categorical sources.

Another unique requirement of working with EO data is that our model must be robust to the highly sparse nature of EO data sources. To reduce swath and tiling artifacts during learning, we utilize an additional forward pass or "student" model trained alongside the "teacher" as described above. The student's input frames are randomly dropped, and some input sources are removed entirely. A key insight is that simply augmenting the teacher's inputs in this way does not influence the objective function to reward yielding near-identical outputs in the same location and same time period regardless of input composition. We minimize 1 minus the dot product between the teacher and student embeddings, both conditioned on the same valid period, and the teacher and student share parameters.

\subsubsection{Learning algorithm}\label{S2.2.2}
Unlike other work pursuant of general geospatial modeling \citep{tseng2023lightweight}, we opted to minimize the number of sources used as model inputs to improve performance and avoid ill-posed reconstruction problems where e.g. climatic information must inform reconstruction of radar data. We found Sentinel-2 L1C, Sentinel-1 GRD, Landsat-8 C2 T1 TOA, and Landsat-9 C2 T1 TOA to be the minimal set providing satisfactory reconstructions across all sources. Inputs were normalized based on global image statistics and no further value modification or augmentation was performed.

Training proceeded using stochastic mini-batch gradient descent to minimize the following objective function with respect to model parameters:

\begin{multline} \label{eq:3}
    l = \frac{a}{M}\sum_{i\in M}{f_i(\textbf{y}_i, \textbf{y}_{i}')w_i} + 
    b\sum_{i=1}^{64}{|u_i \cdot u_{i}'|} \\
    + c\left(\frac{1 - \boldsymbol{u} \cdot \boldsymbol{u}_s}{2}\right) + df_{\text{CLIP}}(\boldsymbol{u}, \boldsymbol{u}_t)
\end{multline}

Indicated by weights {a, b, c, d} of the linear combination, the loss components are:
\begin{enumerate}[label=(\alph*)]
    \item Reconstruction objective with $f_i$ varying as a function of data source.
    \item Batch uniformity objective encouraging a uniform distribution over the training set of embeddings in $S^{63}$
    \item Contrastive consistency objective: encouraging model forward passes with missing inputs to yield embeddings identical to forward passes without missing inputs.
    \item Text contrastive objective: align embeddings derived from text descriptions with embeddings derived from the video sequence.
\end{enumerate}

Loss weights were normalized prior to training.

\begin{table*}
\footnotesize
\begin{tabular}{p{0.33\linewidth} p{0.12\linewidth} p{0.14\linewidth} p{0.15\linewidth} p{0.1\linewidth} }
\toprule
{\bf Data Source} & {\bf Shift invariant  loss distance (m) } & {\bf Re-gridding loss spacing (m)} & {\bf Error metric} & {\bf Loss weight ($w_i$)} \\
\midrule
Sentinel-2 L1C & 20 & \textendash & L1 & 1.0 \\
\midrule
Sentinel-1 GRD & 20 & \textendash & L1 & 1.0 \\
\midrule
Landsat Group & \textendash & 30 & L1 & 1.0 \\
\midrule
PALSAR-2 ScanSAR L2.2 & \textendash & 30 & L1 & 1.0 \\
\midrule
ERA5-Land Monthly Aggregated & \textendash & \textendash & L1 & 1.0 \\
\midrule
GEDI L2A & \textendash & 20 & L1 & 1.0 \\
\midrule
GRACE Monthly Mass Grids V4 & \textendash & 1280 & L1 & 0.5 \\
\midrule
Copernicus DEM GLO-30 & \textendash & 30 & L1 & 1.0 \\
\midrule
NLCD Group & \textendash & 30 & Cross Entropy & 0.5 \\
\bottomrule
\end{tabular}
\caption{Loss configurations for data sources}
\label{table:lossconfig}
\end{table*}

\subsubsection{Reconstruction objective}\label{S2.2.3}
To facilitate learning, in each row a frame was randomly selected from each source sequence. For sequences serving as model inputs these frames were removed from the input sequence directly. Randomly selected frames, and their corresponding metadata and timecode, are then used for computing reconstruction losses. The model's embedding output is concatenated with sensor metadata and the observation timecode, and a small decoder is applied at each pixel embedding to reconstruct the selected frame. Losses are computed against this reconstructed frame, with the nature of the loss $fi$ changing depending on the source $i \in M$ and a source-specific weight (Table \ref{table:lossconfig}).

Shift-invariant loss computes the minimum error metric across any planar shift in reconstruction up to the specified distance. Re-gridding loss re-grids the reconstruction and target using area-weighted averaging to the given nominal resolution before computing the error metric. All losses utilize per-frame per-pixel weights to account for swath edges and invalid pixels; decisions around derivation of weights and masks from source data are detailed in supplemental materials \ref{S1}.

We set the weight of the overall reconstruction objective $a = 1.0$.

Reconstructions each randomly selected summarization periods $[t'_{s_i}, t'_{e_i})$ for source $i \in M_D$ s.t.  $t'_{e_i}- t'_{s_i}>4$ days. For each reconstruction objective, a different embedding corresponding to the unique summarization period is generated, and the target timestamp is normalized to this period on $[0, 1)$. We use the embedding and normalized timecode to reconstruct source i, alongside any source specific metadata specific to the act of measurement detailed in supplemental materials \ref{S1}.

\subsubsection{Batch uniformity objective}\label{S2.2.4}
To increase the utilization of our embedding space, we introduced an objective to encourage the uniform distribution of a given embedding vector over $S^{63}$. Since, on average, random vectors on a sphere will be orthogonal \citep{cai2013distributions}, we can treat this as a necessary condition for our uniformity constraint. Across a batch of image-space embedding vectors \textbf{u}, we can rotate this vector through the batch dimension to get \textbf{u}$'$. Assuming a uniform sample from the training set, batch element pairs $u_i$ and $u'_i$ are effectively uniform random sample pairs from the training set, and we can compute an overall "orthogonality" across the batch:

\begin{equation} \label{eq:4}
    BatchUniformity=\sum_{i=1}^{64}{|u_i \cdot u_{i}'|}
\end{equation}

We minimize this batch uniformity term during training. We note that alone, there are perfectly valid non-uniform distributions for which this tends to zero e.g. clusters of points on opposite poles. In practice, setting the weight in the loss combination for this term > 0 prevented collapse scenarios where this term would tend to 1 otherwise. We ultimately settled on a weight of $b = 0.05$.

While tuning was not performed, as expected, evaluation scores improved when batch uniformity was present, or there was no difference. To better understand the effect of batch uniformity, we performed a sweep of $b\in [0, 0.001, 0.005, 0.01, 0.1]$. We found settings of $b = 0$ and $b = 0.1$ to be the least performant across evaluations, and $b = 0.005$ to be optimal. For some evals, the difference in performance was notable e.g. GLanCE land cover max-trial linear transfer BA scores were \textasciitilde 66.6\% for $b = 0.005$ and \textasciitilde 64.7\% for b = 0. For others, there was only a small improvement e.g. LUCAS land cover max-trial linear transfer BA scores were \textasciitilde 34.7\% for $b = 0.005$ and \textasciitilde 34.2\% for $b = 0$.

\subsubsection{Consistency objective}\label{S2.2.5}
Earth observation data is irregular in space and time. Acquisition campaigns are not always global, have acquisition periods unique from other instruments, vary as a function of solar angle, come on and offline for a number of reasons, and atmospheric conditions at the time an observation is made are unpredictable at local scales. As our model is intended to produce continuous embedding fields over arbitrary regions of Earth's surface, it is crucial that we reduce the effect of these space and time varying irregularities. Unlike purely supervised models we cannot rely on labels to help reduce noisy artifacts. Additionally, we need to ensure that our model provides consistent embeddings for a location regardless of the condition of the inputs as we want to model Earth's underlying landscape dynamics not the measurement process.

To achieve this, we run our forward pass twice. We utilize a teacher model that has access to all inputs, and a student model that has its inputs perturbed. Perturbation proceeds in two stages:
\begin{enumerate}
    \item Entirely drop a source from the inputs. The Landsat Group is randomly dropped 30\% of the time, and Sentinel-1 GRD is dropped 30\% of the time. Sentinel-2 L1C is never dropped.
    \item Select one of three perturbation strategies:
    \begin{enumerate}
        \item Randomly drop time-steps across all sources. 30\% of images from the Landsat Group are randomly dropped, 30\% of images from Sentinel-1 GRD are dropped, and 50\% of images from Sentinel-2 L1C are dropped.
        \item The latter six months of the input sequence across all sources is dropped (forecasting-like).
        \item The former six months of the input sequence across all sources is dropped (backcasting-like).
    \end{enumerate}
\end{enumerate}

If perturbation strategy (a) is used, we choose a unique, random summarization period $[t'_s, t'_e)$ s.t.  $t'_e- t'_s>4$ days intersecting the annual period of the non-perturbed inputs. If strategy (b) is used, we choose a summarization period across the latter six months of the input sequence. If strategy ($c$) is used, we choose a summarization period across the former six months of the input sequence. We note sequence start and end times are not aligned to any calendar unit.

The student and teacher model now embed their inputs based on the shared summary period. The teacher must produce an embedding that the student can mimic with limited inputs while the student must produce an embedding that agrees with the teacher. Given teacher embeddings  and student embeddings s, we minimize:

\begin{equation} \label{eq:5}
    ConsistencyLoss = \frac{1 - \boldsymbol{\mu} \cdot \boldsymbol{\mu}_{s}}{2}
\end{equation}

We set the weight of the overall contrastive objective $c = 0.02$ to balance reconstruction visual quality and maximize student / teacher agreement. We note that while considerably reduced, tile artifacts are still visible in our embedding fields layers resulting from irregular inputs, and these could be removed in future work with a more aggressive consistency objective term.

\subsubsection{Text-contrastive objective}\label{S2.2.6}
We co-train with a frozen language model \citep{team2024gemini} with the goal that embeddings characterizing points on Earth's surface with similar semantics will cluster together.

All points and corresponding text intersecting a training row's geometry and date range were stored with the row. During training, a random text point is selected if available, and we choose a unique random summary period $[t''_s, t''_e)$ s.t.  $t''_e- t''_s > 4$ days intersecting the annual period of the teacher model's inputs. We condition an MLP decoder on the language model's output with this summary period to produce an embedding aligned with the teacher model using standard CLIP loss \citep{radford2021learning}.

We set the weight of the overall text-contrastive objective $d = 0.001$.

\subsection{Model training}\label{S2.3}
AEF was trained for 56 hours on 512 TPU v4 devices over 100k steps in batches of 256 video sequences. Training was sharded by batch, and further by sequence s.t. two TPU v4 devices were allocated to each batch element. Input sequences were subsampled from the training row to 103 frames ($N_i$), comprising 65 Sentinel-2 L1C, 17 Sentinel-1 GRD, and 21 Landsat Group observations. Masks were substituted for unavailable or perturbed frames (see supplemental materials \ref{S2.2.3}).

Learning utilized the Adam optimization strategy \citep{kingma2014adam}, with a piecewise linear learning rate schedule from $0$ to $1e{-4}$ over $[0, 1\text{e}{3})$ steps, then $1\text{e}{-4}$ to $0$ over [$1\text{e}{3}, 1\text{e}{5}]$. Learning hyperparameters were selected to minimize training loss while maintaining satisfactory reconstruction visual quality, stability in the contrastive and batch uniformity objectives, and desired performance on a set of diagnostic evaluations designed to assess whether embeddings could distinguish the presence of specific input sources.

\subsection{Architectural details}\label{S2.4}
We used a model dimension of $DP = 128$ along the precision path, $DT = 512$ along the time path, and $DS = 1024$ along the space path. 15 STP blocks were used in total.
Implicit decoders were two-hidden-layer MLPs with a width of 512.
The VMF bottlenecks utilized a fixed concentration ($\kappa$) of $8\text{e}{3}$.

\section{Evaluation datasets}
We assess AEF performance using a set of evaluation datasets we derived from ten publicly-available reference datasets representing archetypal classification, regression, and change detection use cases (Table \ref{table:evals}).

\begin{table*}
\footnotesize
\centering
\begin{tabular}{p{0.24\linewidth} p{0.09\linewidth} p{0.54\linewidth}}
\toprule
{\bf Property} & {\bf Units} & {\bf Notes} \\
\midrule
x & decimal degrees & Longitude coordinate. Must have at least $10^{-4}$ precision to be considered valid at $\sim$10m resolution.  \\
\midrule
y & decimal degrees & Latitude coordinate. Must have at least $10^{-4}$ precision to be considered valid at $\sim$10m resolution.  \\
\midrule
label & numeric (int or float) & Column recording the label or measurement field used for evaluation. Either dense sequential remapping of ‘label\_name’ values (classification) or  measurement value (regression). \\
\midrule
label\_name & str & (optional) This field can be used to preserve values/codes from the original dataset for readability and visualization. Not required for regression evals.  \\
\midrule
valid\_time\_start\_ms 
valid\_time\_end\_ms & millis & Start and end times defining the range over which the label or measurement is valid (may be same for single-date measurements) and over-which the embedding summary is created. Should not extend more than 6 months before or after the support period.  \\
\midrule
support\_time\_start\_ms 
\mbox{support\_time\_end\_ms} & millis & Start and end times defining a support period for informing prediction. This is the period over which input data is fetched for each row. It must be no longer than 1 year in length.  \\
\midrule
split & str ('train' or 'test') & Each label/observation (row) should be assigned to a fixed train/test split. \\
\midrule
shard & numeric (int) & (optional) Assign a shard to each row for efficient ingestion. A shard should be associated with no more than 2000 rows.  \\
\midrule
\textit{label\_before} & numeric (int) & Integer label for “before” class. \\
\midrule
\textit{label\_before\_name} & str & This is used to preserve “before” values/codes from the original dataset.  \\
\midrule
\textit{label\_after} & numeric (int) & Integer label for “after” class  \\
\midrule
\textit{label\_after\_name} & str & This is used to preserve “after” values/codes from the original dataset.  \\
\midrule
\textit{valid\_time\_start\_before\_ms 
valid\_time\_end\_before\_ms 
valid\_time\_start\_after\_ms
valid\_time\_end\_after\_ms}
& millis & Start and end times defining the range over which the “before” and “after” labels/measurements are valid (may be same for single-date measurements) and over-which embedding summaries are created.   \\
\midrule
\textit{support\_time\_start\_before\_ms
support\_time\_end\_before\_ms
support\_time\_start\_after\_ms
support\_time\_end\_after\_ms} & millis & Start and end times defining the before and after change support periods.  \\
\bottomrule
\end{tabular}
\caption{Evaluation dataset properties. Fields in \textit{italics} are required for change detection datasets only.}
\label{table:evalproperties}
\end{table*}

\subsection{Selection criteria}\label{S3.1}
We selected publicly available datasets to represent a range of different real-world classification, regression and change detection applications. We did not generate any of our own annotations for these evaluations; rather, we identified existing datasets that represented high-quality observation/measurement information and could be used with minimal processing.
We prioritized reference datasets with human-assigned interpretations or physical measurements over model-generated predictions. In some cases (like OpenET), we do include “model proxy tasks” that sample model-generated predictions, but these were selected with strong justification, e.g., proxying a computationally intensive ensemble approach. We generally avoided harmonized datasets that combine multiple sets of annotations initially collected with differing protocols/criteria since this makes it more difficult to understand/interpret results/errors. Given we are evaluating a global model, we attempted to construct an overall suite with large area / global coverage.
We preferred point measurements or annotations over polygons, since reasoning about labels for a specific point rather than over a larger area is more straightforward, and this simplifies sampling of embeddings and other feature vectors for comparisons across approaches. We only selected datasets where point coordinate data (longitude, latitude) in decimal degrees was sufficiently precise relative to a nominal 10-meter resolution, i.e., at least four decimal points of precision (0.0001), which is about 11.1m at the equator. Given our focus on temporal precision, we also required that labels have a clearly defined “valid period” over which the label could be reasonably applied, i.e., a range or instant (annual, monthly, single-date), and this period must intersect 2017 onward, and we ensured that we had representation of different temporal aggregations across our final set of evaluations (Table \ref{table:evals}).
For all candidate datasets, we required that the georeference of a given annotation was not tied to a specific observation. In the spirit of typical computer vision benchmarks or evaluations, many recent general purpose geospatial evaluations provide source imagery (see \cite{schmitt2023there} for review), though we argue this is not appropriate for assessing general purpose geospatial analysis approaches that may leverage time and additional sources uniquely. Were we to require that all baseline approaches tested share the same sources, many would be artificially penalized or would not have been usable at all. As most observational data is tied to a ground truth, not a specific measurement, we argue that future geospatial benchmarking work moves towards evaluations with precise timing and without requisite inputs. We present our evaluation suite as an example of such.

\subsection{Processing}\label{S3.2}
All reference datasets were processed to the standard format and properties in Table \ref{table:evalproperties}. In some cases, e.g., LCMAP, LUCAS, and Canada crops, multiple evals were created using different hierarchies or combinations of source labels, resulting in a final total of 15 derivative datasets (Table \ref{table:evals}). Point observations were filtered to guarantee a minimum distance of 1.28 km between sampled points in order to reduce spatial autocorrelation between training and test sites (and this process will be hereafter referred to as “spatial proximity filtering”). Sample points were allocated to train and test splits such that the training datasets were balanced by class (or regularly spaced bins in the case of regression datasets), with no per-class sample exceeding 300 points and the remainder of the points allocated to an unbalanced test split. When possible, we used existing Google Earth Engine assets for publicly available datasets; otherwise reference datasets were downloaded from archived sources. Additional details on sources and processing for individual datasets are provided in the following sections, and we make our processed evaluation datasets available as a supplemental dataset \citep{deepmind2025evals}.

\subsection{Evaluation datasets}\label{S3.3}
\subsubsection{LCMAP}\label{S3.3.1}
LCMAP (Land Change Monitoring, Assessment, and Projection) is a USGS project aimed at generating annual land cover and land cover change maps for the United States \citep{brown2020lessons}. The LCMAP CONUS Reference Dataset is a collection of human-interpreted labels for 27,000 30m x 30m plots across CONUS, which includes an initial sample of 25,000 randomly distributed points and a supplemental sample of 2,000 stratified random "intensification" sites \citep{pengra2023lcmap}. Land use, land cover, and change process information for each plot are available for annual timesteps for the year 1984 to 2021.

\begin{figure}
    \includegraphics[width=\columnwidth]{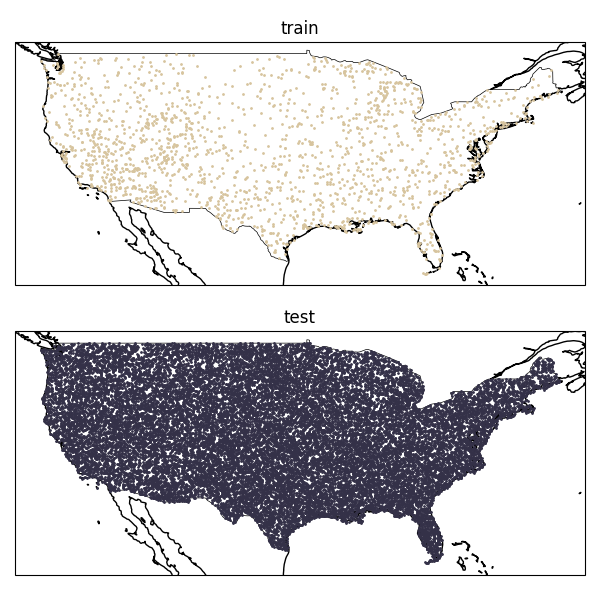}
    \caption{Distribution of LCMAP land use/cover sample locations.}
    \label{fig:lcmap}
\end{figure}

\begin{figure}
    \includegraphics[width=\columnwidth]{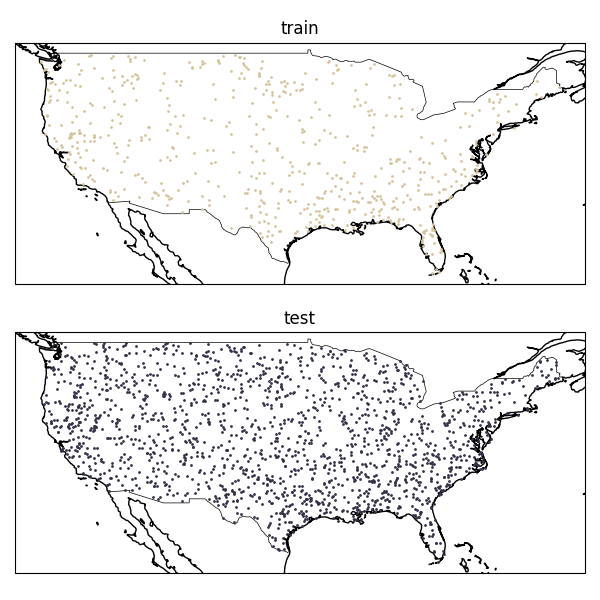}
    \caption{Distribution of LCMAP land cover change sample locations.}
    \label{fig:lcmap-change}
\end{figure}

We use the LCMAP reference datasets, which include multiple label properties and distinct legends, to create two classification datasets (LCMAP land cover and LCMAP land use) and two change detection evaluation datasets (LCMAP land cover change and LCMAP land use change). These datasets are representative of the contiguous United States (CONUS; Figures \ref{fig:lcmap} and \ref{fig:lcmap-change}), and we take performance on LCMAP subdatasets as indicative of performance for operational national-scale land use and land cover mapping and change detection.

\begin{table}
\footnotesize
\begin{tabular}{p{0.10\linewidth} p{0.40\linewidth} p{0.13\linewidth} p{0.13\linewidth}}
\toprule
{\bf label} & {\bf label name} & {\bf train} & {\bf  test} \\
\midrule
0 & Impervious & 300 & 192 \\
\midrule
1 & Grass/forb/herb & 300 & 13545 \\
\midrule
2 & Trees & 300 & 6815 \\
\midrule
3 & Water & 300 & 1326 \\
\midrule
4 & Barren & 300 & 970 \\
\midrule
5 & Shrubs & 300 & 1862 \\
\bottomrule
\end{tabular}
\caption{LCMAP land cover classes and sample counts by split.}
\label{table:lcmap-lc}
\end{table}

\begin{table}
\footnotesize
\begin{tabular}{p{0.10\linewidth} p{0.40\linewidth} p{0.13\linewidth} p{0.13\linewidth}}
\toprule
{\bf label} & {\bf label name} & {\bf train} & {\bf  test} \\
\midrule
0 & Developed & 300 & 1368 \\
\midrule
1 & Agriculture & 300 & 4116 \\
\midrule
2 & Forest & 300 & 7843 \\
\midrule
3 & Other & 300 & 1533 \\
\midrule
4 & Rangeland & 300 & 9510 \\
\midrule
5 & Non-forest Wetland & 300 & 343 \\
\midrule
\bottomrule
\end{tabular}
\caption{LCMAP land use classes and sample counts by split.}
\label{table:lcmap-lu}
\end{table}

\paragraph{LCMAP land cover \& LCMAP land use}
The LCMAP land cover and land use evaluation datasets are based on the "dominant\_landcover" and "dominant\_landuse" fields in the source LCMAP dataset. The LCMAP land cover legend includes six broad cover-type classes (Table \ref{table:lcmap-lc}), while the LCMAP land use legend includes an alternative set of six land use categories (Table \ref{table:lcmap-lu}). The full LCMAP dataset includes labels for all sample locations across all years; we subset the full set of interpretations to 2017-2021 to overlap with our period of interest, and randomly select one year from the time series of labels for each sample point (based on the 'image\_year' property). We sample a total of 300 points from each label class, with the remaining points allocated to the test split. For all points, the valid period (i.e., the period over which embeddings are generated/summarized) is assumed to be January 1 of the "image\_year" from the source dataset through January 1 of the following year. Our final LCMAP land cover evaluation has a total of 1800 training points and 24,710 test points, while our final LCMAP land use evaluation also has a total of 1800 training points and 24,713 test points after pre-processing and spatial proximity filtering.

\begin{table}
\footnotesize
\begin{tabular}{p{0.10\linewidth} p{0.38\linewidth} p{0.15\linewidth} p{0.15\linewidth}}
\toprule
{\bf label} & {\bf label name} & {\bf train} & {\bf  test} \\
\midrule
0 & no change & 150 & 549 \\
\midrule
1 & change & 150 & 142 \\
\bottomrule
\end{tabular}
\caption{LCMAP land use change classes and sample counts by split.}
\label{table:lcmap-luc}
\end{table}

\begin{table}
\footnotesize
\begin{tabular}{p{0.10\linewidth} p{0.38\linewidth} p{0.15\linewidth} p{0.15\linewidth}}
\toprule
{\bf label} & {\bf label name} & {\bf train} & {\bf  test} \\
\midrule
0 & no change & 300 & 1315 \\
\midrule
1 & change & 300 & 405 \\
\bottomrule
\end{tabular}
\caption{LCMAP land cover change classes and sample counts by split.}
\label{table:lcmap-lcc}
\end{table}

\paragraph{LCMAP land cover change \& LCMAP land use change}
Though the source LCMAP dataset includes change process labels, we generated new change labels directly from  "dominant\_landcover", "dominant\_landuse", and "image\_year" fields for parity with the LCMAP classification evaluations. We again subset the full reference dataset to 2017-2021 to overlap with our period of interest. We then label pairs of sequential years where the label is not consistent (i.e., yeart and yeart+1 have different labels) as “change” and years with consistent labels as “no change”. We select only one year-pair for each reference point, and we sample a total of 300 points per class for land cover change (Table \ref{table:lcmap-lcc}) and 150 points per class for land use change (Table \ref{table:lcmap-luc}). Because we want to compare embeddings for two annual labels, we set a valid time start and end for both the before and after periods, where both periods are one year in length from January 1 to January 1 of the following year and assigned based on the "image\_year" property in the source dataset. Our final LCMAP land cover change evaluation has a total of 600 training points and 1,720 test points, and our final LCMAP land use change evaluation has a total of 300 training points and 691 test points after pre-processing and spatial proximity filtering.

\subsubsection{LUCAS}\label{S3.3.2}
\begin{figure}
    \includegraphics[width=\columnwidth]{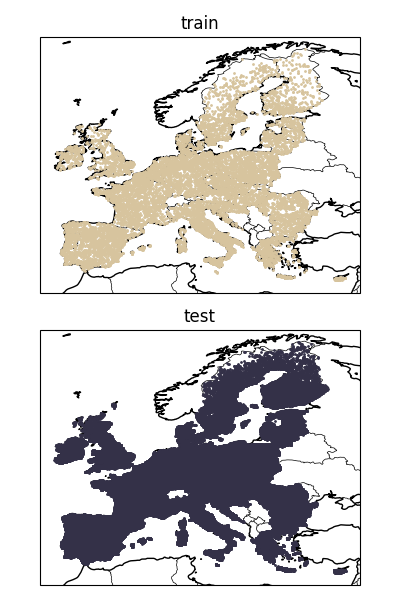}
    \caption{Distribution of LUCAS survey points.}
    \label{fig:lucas}
\end{figure}

The LUCAS (Land Use/Cover Area frame statistical Survey) was designed to gather information on land cover and land use updated via regular harmonised surveys across all European Member States in the survey years 2018 and 2022/2023 \citep{toth2013lucas}.The survey includes over 250,000 sample points throughout the EU (Figure \ref{fig:lucas}), and the survey is repeated every few years to identify changes to land use and cover. One of the primary purposes of the LUCAS dataset is to generate estimates of the area occupied by different land use or land cover types. Given the high level of detail in LUCAS, and the ground-based nature of its collection, we consider LUCAS a challenging assessment of how well a set of geospatial features distinguish detailed ground-level concepts.

We use the LUCAS Harmonized (Theoretical Location, 2006-2018) V1 dataset \citep{dAndrimont2020harmonised} sourced from the Earth Engine Data Catalog ("JRC/LUCAS\_HARMO/THLOC/V1"). We filter the full LUCAS dataset to keep only survey data intersecting our period of interest (year greater than or equal to 2017, and with a location precision of at least 10 meters ("gps\_prec" less than 10). We exclude points labeled as ‘ex\_ante’ given this indicates they were not visited in a given survey year. Finally, we keep only classes with at least 420 samples.

\begin{table*}
\footnotesize
\centering
\renewcommand{\arraystretch}{0.9}
\begin{tabular}{p{0.05\linewidth} p{0.5\linewidth} p{0.08\linewidth} p{0.08\linewidth}}
\toprule
{\bf label} & {\bf label name} & {\bf train} & {\bf  test} \\
\midrule
0 & other\_bare\_soil & 300 & 4538 \\
\midrule
1 & common\_wheat & 300 & 11149 \\
\midrule
2 & other\_leguminous\_and\_mixtures\_for\_fodder & 300 & 797 \\
\midrule
3 & shrubland\_without\_tree\_cover & 300 & 5883 \\
\midrule
4 & broadleaved\_woodland & 300 & 34253 \\
\midrule
5 & grassland\_without\_tree/shrub\_cover & 300 & 42460 \\
\midrule
6 & spruce\_dominated\_coniferous\_woodland & 300 & 4587 \\
\midrule
7 & oats & 300 & 1501 \\
\midrule
8 & lucerne & 300 & 1520 \\
\midrule
9 & inland\_marshes & 300 & 563 \\
\midrule
10 & non\_built-up\_area\_features & 300 & 3117 \\
\midrule
11 & pine\_dominated\_coniferous\_woodland & 300 & 9321 \\
\midrule
12 & pine\_dominated\_mixed\_woodland & 300 & 3883 \\
\midrule
13 & other\_mixed\_woodland & 300 & 3112 \\
\midrule
14 & maize & 300 & 7076 \\
\midrule
15 & grassland\_with\_sparse\_tree/shrub\_cover & 300 & 6984 \\
\midrule
16 & sunflower & 300 & 1236 \\
\midrule
17 & spontaneously\_vegetated\_surfaces & 300 & 8250 \\
\midrule
18 & shrubland\_with\_sparse\_tree\_cover & 300 & 4821 \\
\midrule
19 & barley & 300 & 6647 \\
\midrule
20 & dry\_pulses & 300 & 867 \\
\midrule
21 & sugar\_beet & 300 & 1003 \\
\midrule
22 & temporary\_grasslands & 300 & 3576 \\
\midrule
23 & spruce\_dominated\_mixed\_woodland & 300 & 3408 \\
\midrule
24 & potatoes & 300 & 593 \\
\midrule
25 & rape\_and\_turnip\_rape & 300 & 2793 \\
\midrule
26 & arable\_land\_(only\_pi) & 300 & 1523 \\
\midrule
27 & non\_built-up\_linear\_features & 300 & 5877 \\
\midrule
28 & clovers & 300 & 267 \\
\midrule
29 & buildings\_with\_1\_to\_3\_floors & 300 & 3629 \\
\midrule
30 & triticale & 300 & 587 \\
\midrule
31 & rye & 300 & 1148 \\
\midrule
32 & other\_coniferous\_woodland & 300 & 1066 \\
\midrule
33 & durum\_wheat & 300 & 1837 \\
\midrule
34 & olive\_groves & 300 & 487 \\
\midrule
35 & other\_fresh\_vegetables & 300 & 151 \\
\midrule
36 & mixed\_cereals\_for\_fodder & 300 & 462 \\
\midrule
37 & vineyards & 300 & 137 \\
\midrule
38 & other\_artificial\_areas & 300 & 338 \\
\midrule
39 & peatbogs & 300 & 122 \\
\bottomrule
\end{tabular}
\caption{LUCAS land cover classes and sample counts by split.}
\label{table:lucas-lc}
\end{table*}

\begin{table*}
\footnotesize
\centering
\begin{tabular}{p{0.05\linewidth} p{0.6\linewidth} p{0.08\linewidth} p{0.08\linewidth}}
\toprule
{\bf label} & {\bf label name} & {\bf train} & {\bf  test} \\
\midrule
0 & agriculture\_(excluding\_fallow\_land\_and\_kitchen\_gardens) & 300 & 121309 \\
\midrule
1 & semi-natural\_and\_natural\_areas\_not\_in\_use & 300 & 24198 \\
\midrule
2 & forestry & 300 & 48649 \\
\midrule
3 & road\_transport & 300 & 7037 \\
\midrule
4 & amenities\_museums\_leisure & 300 & 1273 \\
\midrule
5 & other\_abandoned\_areas & 300 & 1460 \\
\midrule
6 & residential & 300 & 9937 \\
\midrule
7 & kitchen\_garden & 300 & 824 \\
\midrule
8 & logistics\_and\_storage & 300 & 206 \\
\midrule
9 & fallow\_land & 300 & 5272 \\
\midrule
10 & community\_services & 300 & 902 \\
\midrule
11 & sport & 300 & 454 \\
\midrule
12 & electricity\_gas\_and\_thermal\_power\_distribution & 300 & 163 \\
\midrule
13 & commerce & 300 & 417 \\
\midrule
14 & mining\_and\_quarrying & 300 & 257 \\
\bottomrule
\end{tabular}
\caption{LUCAS land use classes and sample counts by split.}
\label{table:lucas-lu}
\end{table*}

\paragraph{LUCAS land cover}
Our LUCAS land cover evaluation uses the "lc1" label. After the initial filtering described above, we additionally checked that the ‘lc1’ label is not null, and that percent cover for the "lc1" label ("lc1\_perc") is classified as "50 - 75 \%" or "> 75 \%". This results in a label dataset with 40 land cover classes (Table \ref{table:lucas-lc}). We assume land cover represents an instantaneous observation of state, so we set the valid time to the time at which the land cover observation was made (i.e., time start = time end). We select 300 points per class for training and assign the remainder to the test split (Table \ref{table:lucas-lc}). Our final LUCAS land cover evaluation has a total of 12,000 training points and 191,569 test points after pre-processing and spatial proximity filtering.

\paragraph{LUCAS land use}
Our LUCAS land use evaluation uses the "lu1" label. We apply the same initial filtering as we do for land cover. We also check that the ‘lu1’ label is not null, and that percent cover for the lu1 label ("lu1\_perc") is classified as "50 - 75 \%", "75 - 90 \%", "> 90 \%". This results in a label dataset with 15 land use classes (Table \ref{table:lucas-lu}). We assume land use represents an integrated observation of state (e.g., "forestry" may include periods of tree cover, clearing, and regrowth), so we set the valid period to a one-year window centered on the time at which the land use observation was made, i.e., six months prior, six months after. As with land cover, we select 300 points per class for training and assign the remainder to the test split (Table \ref{table:lucas-lu}). Our final LUCAS land use evaluation has a total of 4,500 training points and 222,358 test points after pre-processing and spatial proximity filtering.

\subsubsection{GLaNCE land cover}\label{S3.3.3}
\begin{figure}
    \includegraphics[width=\columnwidth]{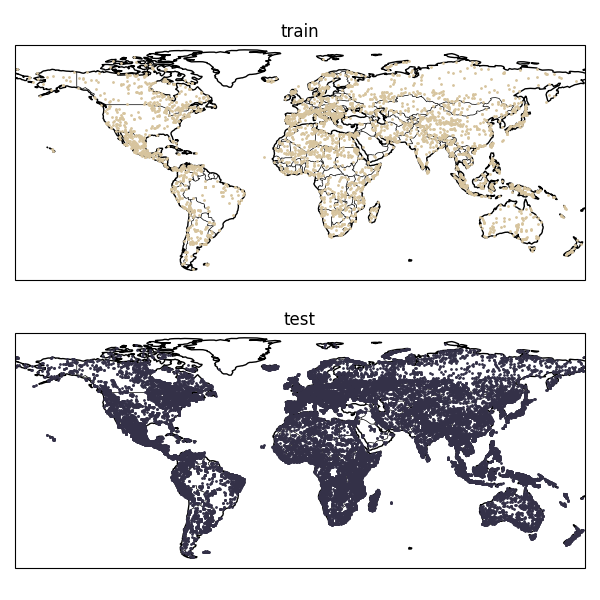}
    \caption{Distribution of GLaNCE sample locations.}
    \label{fig:glance}
\end{figure}

The NASA-funded Global Land Cover Estimation (GLanCE) project seeks to provide high-quality long-term records of land cover and land cover change at a 30m spatial resolution for the 21st century (2001 to present) \citep{friedl2022medium}. The GLanCE training dataset was designed for regional-to-global land cover and land cover change analyses \citep{stanimirova2023global}. Similar to LCMAP, the dataset legend is intended to support a broader community of end-users; however, the GLaNCE dataset is a global sample (Figure \ref{fig:glance}). Thus, we consider the GLaNCE dataset a good proxy for general-purpose global (as opposed to national) land cover mapping.

Our GLaNCE evaluation dataset is derived from the GLanCE training dataset in the GEE Community Catalog ("projects/sat-io/open-datasets/GLANCE/GLANCE\_TRAINING\_DATA\\
\_V1") \citep{Roy2025catalog}. Though published GLaNCE data products use Level 1 labels \citep{arevalo2022global}, we use the Level 2 of the labeling hierarchy ("Glance\_Class\_ID\_level2") as a test of maximizing thematic detail. Given that GLaNCE includes a number of other datasets, some of which overlap with other evaluation datasets, e.g., LCMAP, we select a subset of sources, specifically the MODIS STEP dataset (STEP), results of spectral-temporal clustering (CLUSTERING), a labeled dataset from the NASA Arctic-Boreal Vulnerability Experiment (ABoVE), and a set of annotations collection by the project team ("Dataset\_Code" = 1, 2, 4, or 704). GLaNCE labels are associated with time segments, i.e., labels have a start and end date similar to our use of a valid period. We select only labeled segments with an end year after 2017 ("End\_Year" greater than or equal to 2017). We remove null values as well as the  "ice\_and\_snow" and "moss" categories, which have fewer than 500 samples per class. This results in a final dataset with eleven classes, and we select 300 training points per class with the remainder allocated to the test split (Table \ref{table:glance}). Though we note that segments could be converted to a series of annual labels for each location, this approach would be subject to greater temporal autocorrelation across labels for the same location; instead, we sample a random year between segment start and end dates as an annual valid period to ensure more independent sampling of the time domain. Our final GLaNCE land cover evaluation has a total of 3,300 training points and 31,585 test points after pre-processing and spatial proximity filtering.

\begin{table}
\footnotesize
\begin{tabular}{p{0.10\linewidth} p{0.35\linewidth} p{0.15\linewidth} p{0.15\linewidth}}
\toprule
{\bf label} & {\bf label name} & {\bf train} & {\bf  test} \\
\midrule
0 & water & 300 & 1167 \\
\midrule
1 & developed & 300 & 878 \\
\midrule
2 & soil & 300 & 291 \\
\midrule
3 & rock & 300 & 921 \\
\midrule
4 & sand & 300 & 1195 \\
\midrule
5 & deciduous & 300 & 2700 \\
\midrule
6 & evergreen & 300 & 5959 \\
\midrule
7 & mixed & 300 & 2425 \\
\midrule
8 & shrub & 300 & 2118 \\
\midrule
9 & grassland & 300 & 6952 \\
\midrule
10 & agriculture & 300 & 6979 \\
\bottomrule
\end{tabular}
\caption{GLaNCE classes and sample counts by split.}
\label{table:glance}
\end{table}

\subsubsection{Africa crop mask}\label{S3.3.4}
\begin{figure}
    \includegraphics[width=\columnwidth]{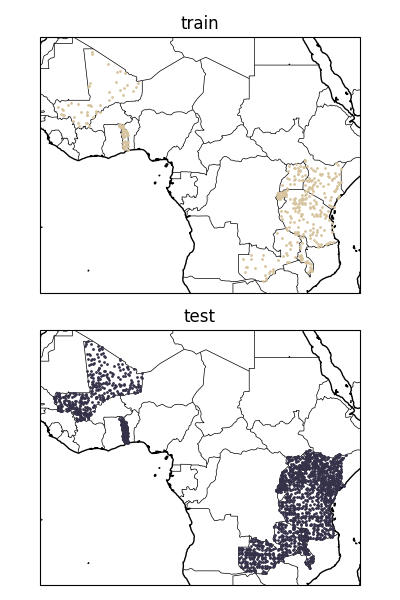}
    \caption{Distribution of Africa crop mask sample locations.}
    \label{fig:africa-crop-mask}
\end{figure}

Our Africa crop mask evaluation was derived from a manually-labeled reference dataset designed to validate the accuracy of cropland maps derived from land cover maps \citep{kerner2024accurate}. The dataset includes pointwise (binary) annotations for cropland versus non-cropland in eight Sub-Saharan African countries (Kenya, Rwanda, Uganda, Tanzania, Mali, Malawi, Togo, and Zambia; Figure \ref{fig:africa-crop-mask}). For all countries except Mali, where the percentage of cropland area is small, reference points were selected by drawing a random uniform sample of point locations within each country’s boundaries. For each sample, trained individuals inspected images from each month in the country’s growing season to determine whether the point contained active cropland, defined as "points where patterns of sowing, growing, and/or harvesting in an agricultural field could be observed during the relevant agricultural season within a 12-month period" \citep{kerner2024accurate}. At least two annotators labeled every point to maximize label confidence, and points that did not have unanimous agreement between annotators were discarded to ensure high-confidence labels in the final reference dataset. We consider the Africa crop mask reference dataset a good proxy for general agricultural land use in landscapes where there has been notable disagreement among existing mapping efforts.

\begin{table}
\footnotesize
\begin{tabular}{p{0.10\linewidth} p{0.38\linewidth} p{0.15\linewidth} p{0.15\linewidth}}
\toprule
{\bf label} & {\bf label name} & {\bf train} & {\bf  test} \\
\midrule
0 & not\_crop & 200 & 2038 \\
\midrule
1 & crop & 200 & 118 \\
\bottomrule
\end{tabular}
\caption{Africa crop mask classes and sample counts by split.}
\label{table:africa-crop}
\end{table}

We accessed the crop mask reference datasets for individual countries as Earth Engine assets provided by the authors ("projects/bsos-geog-harvest1/assets/harvest-reference-datasets/*"), though we note that these datasets are also available as archived shapefiles \citep{kerner2024dataset}. We merge separate datasets for Kenya, Rwanda, Uganda, Tanzania, Mali, Malawi, Togo, and Zambia into a single dataset and assign labels based on the “crop\_label” field. We use an annual valid period covering Jan 1 2019 to Jan 1 2020 for all countries except Malawi, where reference labels are for 2020, so we use a valid period of Jan 1 2020 to Jan 1 2021 instead. Our final Africa crop mask evaluation has a total of 400 training points and 2,156 test points after pre-processing and spatial proximity filtering (Table \ref{table:africa-crop})

\subsubsection{Canada crops}\label{S3.3.5}
\begin{figure}
    \includegraphics[width=\columnwidth]{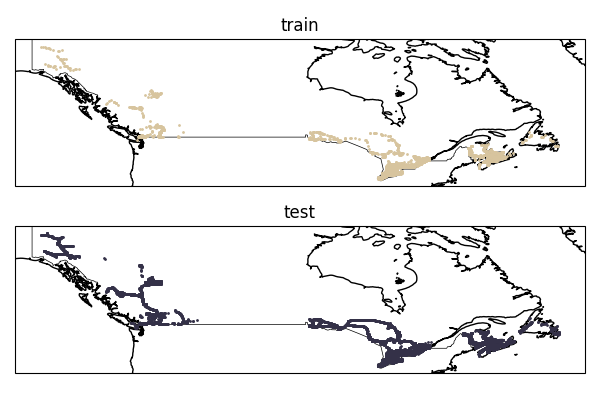}
    \caption{Distribution of Canada crops sample locations.}
    \label{fig:canada-crops}
\end{figure}

The Canadian AAFC (Agriculture and Agri-Food Canada) Annual Crop Inventory Ground Truth Data is an annual field-by-field inventory of Canadian crops \citep{AAFC2024dataset}. It does not cover the whole country; rather, a "windshield survey" is done annually for provinces where crop data is not provided to provincial crop insurance companies (Figure \ref{fig:canada-crops}). This data was originally collected as training and validation points for use in the AAFC Annual Crop Inventory (ACI, which looks at state and trends in national agriculture production. We create two evaluation datasets at two different levels of classification hierarchy, which we refer to as Canada crops coarse and Canada crops fine. We assume that these datasets are a good proxy for performance on multi-level crop classification at a national scale, and unlike reference datasets that rely on interpretation of imagery, the windshield-survey approach indicates performance scaling sparse ground-based observations from national inventory datasets.

We downloaded prepackaged shapefiles  for the years 2017, 2018, 2019, 2020, 2021, and 2022 (2023 data was also available but formatting was not consistent with other years, so we dropped from consideration). Processing for the two subdatasets (coarse and fine) are described in the following sections.

\paragraph{Canada crops (coarse)}
We use the Landuse Category Code (CATCODE) and corresponding English Landuse Category Name (CATNAME) to derive our Canada crops coarse evaluation. We impose a minimum overall per-class sample size of n=100, and cull any classes that do not meet this minimum count requirement. Given the class distribution is highly imbalanced and includes several classes with less than 200 samples per class, we set a proportional split rather than a fixed sample, allocating 60\% to training and reserving 40\% for testing and re-sampling as part of the evaluation process to re-balance the training set (Table \ref{table:canada-coarse}). We treat the observations as instantaneous and assign the Date Collected (DATE\_COLL) as the valid period start and end (single-date). Our final Canada crops coarse evaluation has a total of 2,831 training points and 13,248 test points after pre-processing and spatial proximity filtering.

\begin{table}
\footnotesize
\begin{tabular}{p{0.10\linewidth} p{0.38\linewidth} p{0.15\linewidth} p{0.15\linewidth}}
\toprule
{\bf label} & {\bf label name} & {\bf train} & {\bf  test} \\
\midrule
0 & Agr. Cereals & 500 & 3059 \\
\midrule
1 & Agr. Forages & 500 & 4914 \\
\midrule
2 & Agr. Fruits (Berry \& Annual) & 207 & 138 \\
\midrule
3 & Agr.  Fruits (Trees) & 75 & 51 \\
\midrule
4 & Agr.  Oilseeds & 500 & 1797 \\
\midrule
5 & Agr.  Pulses & 76 & 51 \\
\midrule
6 & Agr.  Vegetables & 277 & 186 \\
\midrule
7 & Agr.  Others & 196 & 132 \\
\midrule
8 & Non-Agr. & 500 & 2920 \\
\bottomrule
\end{tabular}
\caption{Canada crops (coarse) classes and sample counts by split.}
\label{table:canada-coarse}
\end{table}

\paragraph{Canada crops (fine) }
For the Canada crops fine evaluation, we use the Landuse Code (LANDCODE) and associated English Landuse Name (LANDNAME). These species-level labels present an excellent opportunity to characterize performance on highly detailed legends and viability for agricultural use cases that require this level of detail. However, many categories have few samples, and some categories may be too noisy to draw any sound conclusions about performance (particularly the “Undifferentiated” categories, which could include examples of the same crop type but grouping different phenologies already represented individually). 

\begin{table}
\footnotesize
\begin{tabular}{p{0.10\linewidth} p{0.38\linewidth} p{0.15\linewidth} p{0.15\linewidth}}
\toprule
{\bf label} & {\bf label name} & {\bf train} & {\bf  test} \\
\midrule
0 & Barley (Undiff) & 110 & 74 \\
\midrule
1 & Oats & 123 & 82 \\
\midrule
2 & Spring Wheat & 99 & 66 \\
\midrule
3 & Winter Wheat & 390 & 260 \\
\midrule
4 & Corn & 500 & 1546 \\
\midrule
5 & Pasture/Forage & 93 & 63 \\
\midrule
6 & Alfalfa & 257 & 172 \\
\midrule
7 & Mixed Forage & 500 & 2768 \\
\midrule
8 & Pasture & 500 & 628 \\
\midrule
9 & Unimproved Pasture & 250 & 167 \\
\midrule
10 & Blueberry (Undiff) & 112 & 75 \\
\midrule
11 & Canola/Rapeseed & 68 & 46 \\
\midrule
12 & Soybeans & 500 & 1673 \\
\midrule
13 & Potatoes & 145 & 97 \\
\midrule
14 & Native Grassland & 68 & 46 \\
\midrule
15 & Shrubland & 249 & 167 \\
\midrule
16 & Urban & 291 & 194 \\
\midrule
17 & Barren & 141 & 94 \\
\midrule
18 & Water & 106 & 72 \\
\midrule
19 & Coniferous & 153 & 102 \\
\midrule
20 & Mixedwood & 361 & 242 \\
\midrule
21 & Wetland & 222 & 149 \\
\midrule
22 & Abandoned (Overgrown) & 168 & 112 \\
\midrule
23 & Abandoned (Shrubs) & 159 & 106 \\
\bottomrule
\end{tabular}
\caption{Canada crops (fine) classes and sample counts by split.}
\label{table:canada-fine}
\end{table}

Specifically, we:
\begin{itemize}
    \item Remove "Cereals (Undiff)"
    \item Merge "Barley (Undiff)", "Winter Barley", "Spring Barley" 
    \item Remove "Wheat (Undiff)"
    \item Merge "Rye (Undiff)", "Winter Rye", "Spring Rye"
    \item Merge "Triticale (Undiff)", "Spring Triticale", "Winter Triticale”
    \item Merge "Blueberry (Undiff)", "Blueberry - High Bush", "Blueberry - Low Bush" 
    \item Merge "Beans (Undiff)", "Adzuki Beans", "Otebo Beans", "Black Beans","Cranbery Beans", "Fababeans", "Kidney Beans", "Lima Beans", "White Beans", "Edible (generic) Beans" 
    \item Merge "Peas (Undiff)" and "Field Peas".
    \item Remove "Vegetables (Undiff)"
\end{itemize}

After these merges and removals, we check to ensure remaining categories have a viable number of samples (greater than 100 per class). As with Canada crops coarse, there is a high degree of variability in per-class sample sizes, so we use a proportional rather than fixed per class sample size, allocating 60\% to training and 40\% to testing and re-balancing the training set using repeat sampling during training. We again treat window survey observation as instantaneous and assign the Date Collected (DATE\_COLL) as the valid period start and end (single-date). Our final Canada crops fine evaluation has a total of 5,565 training points and 9,001 test points after pre-processing and spatial proximity filtering (Table \ref{table:canada-fine}). 

\subsubsection{Ethiopia crops}\label{S3.3.6}
The Ethiopian Crop Type 2020 (EthCT2020) dataset is a benchmark for environmental and agricultural remote sensing applications in complex Ethiopian smallholder wheat-based farming systems \citep{blasch2024ethiopian}. The dataset consists of harmonized, quality-controlled, and georeferenced in-situ samples of annual crop types \citep{blasch2024dataset}. Like the Canada crops inventory, these in situ samples represent an important class of sparse-but-high-quality data, and we take additional steps to process this dataset for consistency with our evaluation dataset protocols and standards.

\begin{figure}
    \includegraphics[width=\columnwidth]{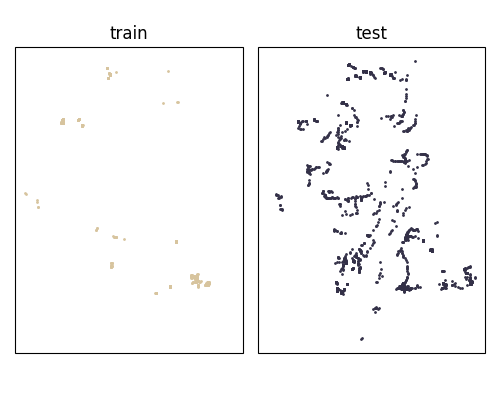}
    \caption{Distribution of Ethiopia crops sample locations.}
    \label{fig:ethiopia-crops}
\end{figure}

We downloaded the dataset from Mendeley \citep{blasch2024dataset}. Per the dataset description, this shapefile contains the delimitation of 2,793 circular plots (10 m radius) located in cultivated fields, and the crop information (crop group and crop class) of the 2020/21 main Meher season (June 2020 to February 2021) for each field plot. Given close proximity of many of the interpreted sites, we do not simply remove points to satisfy minimum distance criteria. Rather, we create connected components by joining points within 1.28 km radii, and assign points to the train or test split based on their component membership.This avoids the scenario where a train and test point are in the same spatial neighborhood. Component membership assignment is performed to minimize the number of points off from which all classes have allocated 20\% of their points to their train split, though the large size of some components lead to an imbalanced result. Given our evaluation protocol sub-samples to the minimum class size this was not problematic. We lastly remove all datapoints with crop classes that have a total of < 49 train points. The valid time is treated as instantaneous (single-date) and set to the data collection timestamp, ("sub\_dat") from the original dataset. Our final Ethiopia crops evaluation has a total of 873 training points and 1,657 test points after pre-processing and spatial proximity filtering (Figure \ref{fig:ethiopia-crops}, Table \ref{table:ethiopia-crops}). 

\begin{table}
\footnotesize
\begin{tabular}{p{0.10\linewidth} p{0.38\linewidth} p{0.15\linewidth} p{0.15\linewidth}}
\toprule
{\bf label} & {\bf label name} & {\bf train} & {\bf  test} \\
\midrule
0 & wheat & 377 & 1700 \\
\midrule
1 & barley & 82 & 20 \\
\midrule
2 & maize & 66 & 30 \\
\midrule
3 & teff & 49 & 206 \\
\bottomrule
\end{tabular}
\caption{Ethiopia crops classes and sample counts by split.}
\label{table:ethiopia-crops}
\end{table}

\subsubsection{US trees}\label{S3.3.7}
\begin{figure}
    \includegraphics[width=\columnwidth]{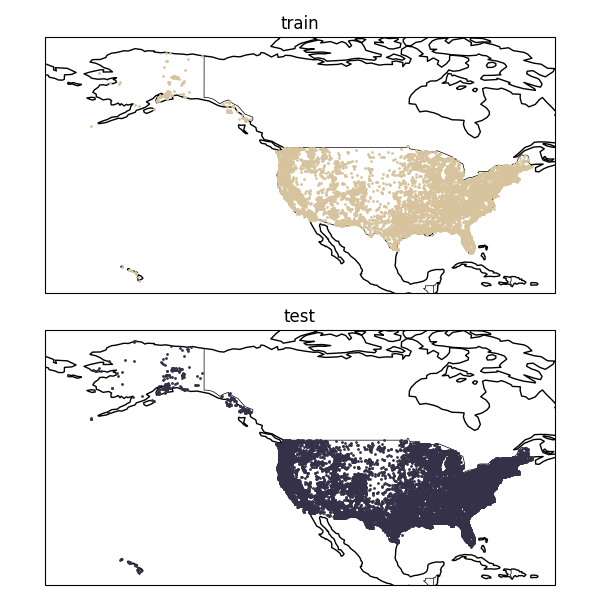}
    \caption{Distribution of US trees sample locations.}
    \label{fig:us-trees}
\end{figure}

To evaluate performance on biodiversity-related applications, we leveraged the Global Biodiversity Information Facility (GBIF) records, specifically research-grade observations from the iNaturalist citizen science repository \citep{gbif2024occurrence}. GBIF is a comprehensive collection of species occurrence records. Unlike our text pretraining dataset, here we focus on genus-level taxonomic labels (as opposed to alignment with text embeddings for the information associated with a given species). We chose to focus on tree genera in the United States given interest in forest species composition mapping across a diversity of forest types (Figure \ref{fig:us-trees}).

We select GBIF records for the period Jan 1 2017 to Jan 1 2023 and filter to just observations where the genus label is found in a list of tree genera sourced from the US Forest Service and country code is set to the US. Observations must be labeled as human or machine observations, and have a maximum spatial uncertainty of 10 meters. From this initial list of tree genera observations, we get the five most frequently observed species for each of the US states, including Alaska and Hawaii, then combine into a single deduplicated list of common US tree genera. We select these genera for further processing.

Of the remaining observations across common tree genera, we drop any that have less than 500 samples per class, resulting in a final set of 39 genera (Table \ref{table:us-trees}). We allocate 300 samples to the train split and the rest to the test split (Table \ref{table:us-trees}). We treat observations as instantaneous labels, using the date of the observation record (eventdate) as a single-date valid period. Our final US trees evaluation has a total of 11,700 training points and 33,682 test points after pre-processing and spatial proximity filtering.

\begin{table}
\footnotesize
\renewcommand{\arraystretch}{0.9}
\begin{tabular}{p{0.10\linewidth} p{0.38\linewidth} p{0.15\linewidth} p{0.15\linewidth}}
\toprule
{\bf label} & {\bf label name} & {\bf train} & {\bf  test} \\
\midrule
0 & abies & 300 & 827 \\
\midrule
1 & acer & 300 & 1296 \\
\midrule
2 & aesculus & 300 & 665 \\
\midrule
3 & ailanthus & 300 & 857 \\
\midrule
4 & alnus & 300 & 220 \\
\midrule
5 & amelanchier & 300 & 203 \\
\midrule
6 & asimina & 300 & 614 \\
\midrule
7 & betula & 300 & 917 \\
\midrule
8 & carya & 300 & 628 \\
\midrule
9 & cercis & 300 & 1296 \\
\midrule
10 & cornus & 300 & 1296 \\
\midrule
11 & diospyros & 300 & 934 \\
\midrule
12 & elaeagnus & 300 & 1296 \\
\midrule
13 & fagus & 300 & 1187 \\
\midrule
14 & gleditsia & 300 & 353 \\
\midrule
15 & ilex & 300 & 1296 \\
\midrule
16 & juglans & 300 & 620 \\
\midrule
17 & juniperus & 300 & 1296 \\
\midrule
18 & liquidambar & 300 & 1296 \\
\midrule
19 & liriodendron & 300 & 1173 \\
\midrule
20 & maclura & 300 & 317 \\
\midrule
21 & magnolia & 300 & 874 \\
\midrule
22 & morus & 300 & 387 \\
\midrule
23 & picea & 300 & 621 \\
\midrule
24 & pinus & 300 & 1296 \\
\midrule
25 & populus & 300 & 1296 \\
\midrule
26 & prosopis & 300 & 1057 \\
\midrule
27 & prunus & 300 & 1296 \\
\midrule
28 & pseudotsuga & 300 & 602 \\
\midrule
29 & quercus & 300 & 1296 \\
\midrule
30 & sabal & 300 & 272 \\
\midrule
31 & salix & 300 & 957 \\
\midrule
32 & sassafras & 300 & 1006 \\
\midrule
33 & taxodium & 300 & 271 \\
\midrule
34 & thuja & 300 & 474 \\
\midrule
35 & triadica & 300 & 300 \\
\midrule
36 & tsuga & 300 & 1190 \\
\midrule
37 & ulmus & 300 & 604 \\
\midrule
38 & yucca & 300 & 1296 \\
\bottomrule
\end{tabular}
\caption{US trees genera labels and sample counts by split.}
\label{table:us-trees}
\end{table}

\subsubsection{Descals oil palm}\label{S3.3.8}
\begin{figure}
    \includegraphics[width=\columnwidth]{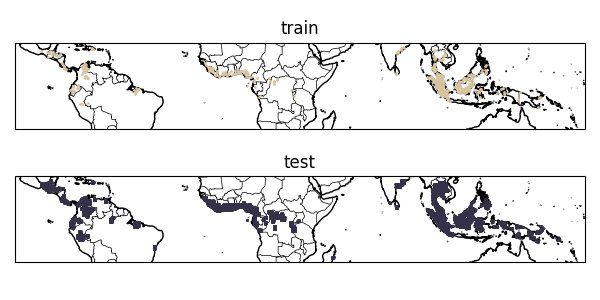}
    \caption{Distribution of Descals oil palm sample locations.}
    \label{fig:descals}
\end{figure}

The Global oil palm extent and planting year from 1990 to 2021 reference dataset is an updated version of a dataset used to validate a previously published global map of smallholder and industrial closed-canopy oil palm plantations \citep{descals2021high}. The updated dataset was refined to validate a 10-meter resolution global map of industrial and smallholder oil palm developed using Sentinel-1 data for the years 2016–2021 \citep{descals2024dataset}. The 2024 version of the dataset covers regions where oil palm is found worldwide (Figure \ref{fig:descals}) and makes several notable improvements over the previous version, including updating labels where coconut plantations were incorrectly labeled as oil palm and relabeling young plantations that were initially considered ‘other’ as oil palm. Reference sites from the initial study were selected using a simple random sample, making the sampling design reusable for creating statistically rigorous accuracy metrics. We interpret this dataset as reflecting land use as an oil-palm plantation: points are labeled as being part of a plantation if  they were a closed canopy plantation some time in 2016-2021. We consider the Descals oil palm reference dataset a useful evaluation of performance for subtle land use and tropical commodity mapping.

We download the validation points (Validation\_points\_GlobalOP2016-2021.zip) from the archived dataset \citep{descals2024dataset}. The dataset has three classes, where class 0 represents other land covers that are not closed-canopy oil palm; class 1 represents closed-canopy industrial oil palm; and class 2 represents closed-canopy smallholder oil palm (Table \ref{table:descals}). These classes were initially determined based on observations on imagery from 2019, but further informed by observations over the years 2016-2021. We assume that non-plantation to plantation transitions are more likely than the reverse, so while it's possible that observations in some years may not match the assigned label, it's less likely for later years. In order to make this into a dataset that is useful as a training dataset, we randomly assign one of the years in [2019, 2021] to each row and set the valid period to that entire year. Our final Descals oil palm evaluation has a total of 600 training points and 16,877 test points after pre-processing and spatial proximity filtering. 

\begin{table}
\footnotesize
\begin{tabular}{p{0.10\linewidth} p{0.42\linewidth} p{0.13\linewidth} p{0.13\linewidth}}
\toprule
{\bf label} & {\bf label name} & {\bf train} & {\bf  test} \\
\midrule
0 & Other & 200 & 16323 \\
\midrule
1 & Industrial oil palm & 200 & 461 \\
\midrule
2 & Smallholder oil palm & 200 & 93 \\
\bottomrule
\end{tabular}
\caption{Descals oil palm labels and sample counts by split.}
\label{table:descals}
\end{table}

\subsubsection{OpenET ensemble}\label{S3.3.9}
The OpenET project aims to make satellite-based estimates of the total amount of water that is transferred from the land surface to the atmosphere through the process of evapotranspiration (ET) available for improved water management \citep{melton2022openet, volk2024assessing}. OpenET datasets include ET estimates from six different satellite-driven ET models as well as an ensemble product, which is calculated as the mean of the ensemble after filtering and removing outliers using the median absolute deviation approach. All models currently use 30-meter Landsat data to produce ET estimates, and the monthly ET dataset provides data on total ET by month as an equivalent depth of water in millimeters.

\begin{figure}
    \includegraphics[width=\columnwidth]{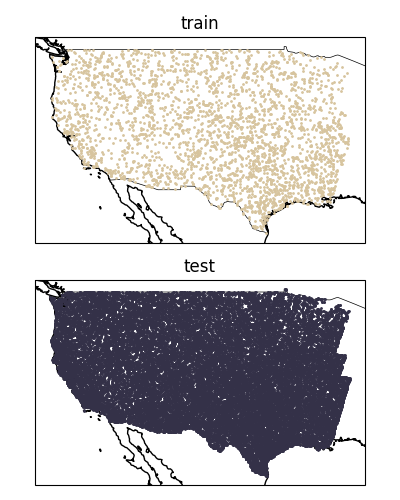}
    \caption{Distribution of OpenET ensemble sample locations.}
    \label{fig:openet}
\end{figure}

Our OpenET evaluation dataset is derived from the OpenET monthly total ensemble product in Earth Engine. This dataset is designed not to measure performance on ET estimation directly. Rather, it characterizes performance on proxying the OpenET model ensemble, given that ensemble approaches are inherently computationally intensive and challenging to scale and has historically limited OpenET ensemble coverage (i.e., Figure \ref{fig:openet}). Thus accurate proxy models could be a more viable means of scaling ensemble results over larger extents.

\begin{figure}
    \includegraphics[width=\columnwidth]{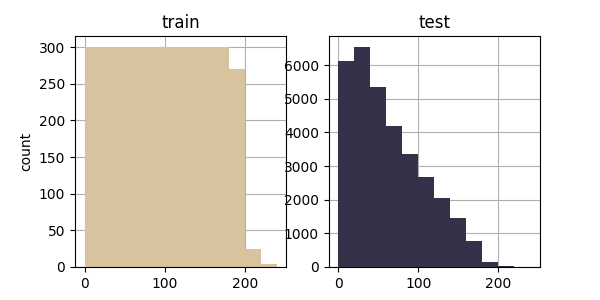}
    \caption{Distribution of OpenET ensemble values. Values are reported in total mm evapotranspiration (ET) per month.}
    \label{fig:openet-hist}
\end{figure}

We construct our OpenET evaluation by first tiling CONUS in 35km grid cells in the Albers conic projection with EPSG code 5070. For each grid cell, we select a random month from all possible months mapped in the source OpenET ensemble product, and sample 2 locations for each of 10 equally spaced 20mm bins between 0mm and 200mm. Locations with ET values > 200mm were assigned to the highest bin (Figure \ref{fig:openet-hist}). We ignore locations where less than 5 models in the ensemble ran, or the disagreement between the minimum and maximum model estimates exceeded 10mm. To each sample we assigned a valid period of the entire month from which it was drawn, and a support period ending with the end of the valid period, and extending 1 year prior: this was chosen to emulate the realistic scenario where evapotranspiration estimates are desired at the conclusion of a given calendar month. We selected 300 train points per bin, and allocated the remainder to test. Our final OpenET ensemble evaluation has a total of 3,000 training points and 32,683 test points after pre-processing and spatial proximity filtering.

\subsubsection{ASTER Global Emissivity Database (GED)}\label{S3.3.10}
Emissivity is an intrinsic property of materials that describes how efficiently a surface can emit radiation at a certain wavelength. The Advanced Spaceborne Thermal Emission and Reflection Radiometer (ASTER) is the most detailed emissivity map of the Earth. The dataset was created by processing millions of cloud free ASTER images acquired between 2000 to 2008 \citep{hulley2015aster}. ASTER-GED land surface temperature and emissivity are generated using a number of physical process models that aim to separate temperature and emissivity from overall reflectance signals. Like OpenET, we consider ASTER-GED a proxy task that we can use to evaluate performance in terms of reproducing the results of a more complex modelling workflow for estimating a material property.

\begin{figure}
    \includegraphics[width=\columnwidth]{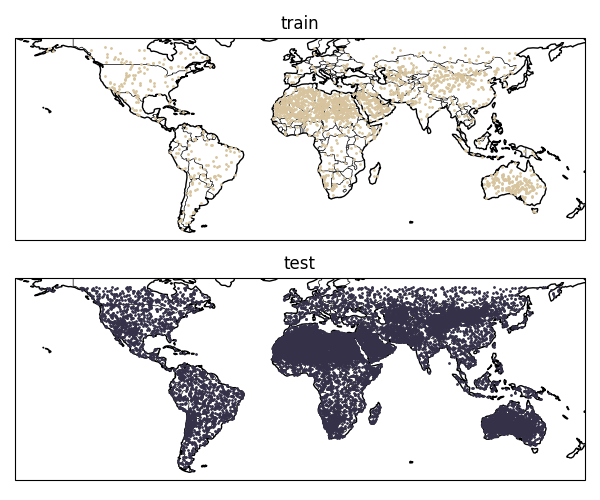}
    \caption{Distribution of ASTER GED sample locations.}
    \label{fig:asterged}
\end{figure}

\begin{figure}
    \includegraphics[width=\columnwidth]{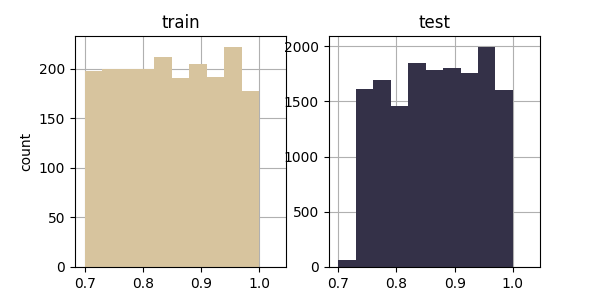}
    \caption{Distribution of ASTER GED values. Note that emissivity of natural Earth surfaces is a unitless quantity that typically ranges between 0.6 and 1.0. Surfaces with emissivities less than 0.85 are typically found over deserts and semiarid areas; vegetation, water, and ice have high emissivities above 0.95.}
    \label{fig:asterged-hist}
\end{figure}

We sample the 100-meter ASTER-GED product available in Earth Engine (AG100: ASTER Global Emissivity Dataset 100-meter V003; "NASA/ASTER\_GED/AG100\_003")  (Figure \ref{fig:asterged}). The original dataset was generated for the years 2000-2008. We assume surface properties are relatively stable at 100-meter resolution (though acknowledge this introduces added uncertainty in signal). We select 2017 for both the support and valid periods. We construct our sample by first arbitrarily selecting the 8.3 $\mu$m band ("emissivity\_band10") as our label, and subsetting to locations for which all wavelength emissivity estimates had standard deviations < 0.05. We further subsetted to locations exclusively over land, and within $\pm$ 60° latitude. We then oversampled by selecting 330k of the remaining locations over land, and culling them with spatial proximity filtering. We then sample 2000 locations from 10 equally spaced bins between emissivity values of 0.7 and 1.0. We selected 200 train points per 0.03 sized bin, and allocated the remainder to test (Figure \ref{fig:asterged-hist}). Our final ASTER GED evaluation has a total of 2,000 training points and 15,636 test points after pre-processing.

\section{Details on evaluation setup}\label{S4}
For each of our 15 evaluation datasets, we iterate through a suite of trials designed to test model performance with varying degrees of label sparsity and various transfer methods.

\subsection{Predictors}\label{S4.1}
Given a training set $(e^t_1, l^t_1),...,(e^t_N,l^t_N)$ of $N$ examples and a validation set $ l^v_1,...,l^v_M$ of $M$ embeddings with held out labels, we fit a predictor, or "transfer method", using the training set and then report results on the validation set.

The purpose of the predictor is to obtain a label $l_j$ for each embedding $e_j^v$ in the validation set. We consider two simple predictors: a linear predictor (or "linear probe") and kNN. We chose these predictors as they are applicable to low-shot data domains and require minimal parameterization which avoids unduly penalizing any given method due to non-optimal hyperparameters.

For linear classification, we follow the "RidgeClasifier" in scikit-learn \citep{pedregosa2011scikit} and use a one-vs-rest approach with a pure-linear model per class. For each class in the training set, we create labels $\{-1, 1\}$ for each item in the training set, where $-1$ denotes absence of the class and $1$ presence. We then use ordinary least squares to fit this model and obtain the predictor with $\lambda=0$. As the classes are mutually exclusive, a given example in the validation is then classified based on which of the classifiers gave the highest prediction.

For linear regression, we perform a simple least-squares fit between predictor and target variables with $\lambda=0$.

To run the kNN predictor, the nearest set of $k$ embeddings $e^t_{n1},...,e^t_{nk}$ in the training set is found under an l2 distance. For a classification evaluation, with a set $C$ of possible class labels, the majority class labels of those $k$ embeddings in the train set is chosen:

\begin{equation} \label{eq:6}
    l_j = \max_{c \in \mathcal{C}} \sum_i^k{l^t_{ni} = c} 
\end{equation}

For kNN regression, a simple average is used to obtain the final result:

\begin{equation} \label{eq:7}
    l_j = \frac{1}{k} \sum_i^k l^t_{ni}
\end{equation}

For direct classification of change, we simply concatenate each pair of embeddings characterizing Earth's local state before/after an event in both train and test, and follow through with the aforementioned predictors as stated.

For unsupervised anomaly detection, we discard the train set and l2 normalize each pair of validation embeddings providing a normalized embedding before the event: , and after the event . We now take the dot product between each pair, remapped s.t. $0$ = embeddings were the same, $1$ = embeddings were on opposite poles: 

\begin{equation} \label{eq:8}
    d_j = \frac{1 - \bar{e^v_j} \cdot \bar{p^v_j}}{2}
\end{equation}

We now choose a global threshold $s$ on $(0, 1)$ to binarize all $d_j$, and thus provide a predicted $l_j$ to compare to $l_jv$. $s$ is chosen from one of $[0.1, 0.2, 0.3, 0.4, 0.5, 0.6, 0.7, 0.8, 0.9]$ to maximize BA over the entire validation dataset, and then all other metrics are computed using this threshold.

\subsection{Kappa adjustment}\label{S4.2}
When stated, kappa adjusted metrics were rescaled by linearly transforming the metric range by the metric value a "random" predictor would achieve on average. For Balanced Error Rate (BER), 1 was remapped to the balanced error rate of a random predictor. Any scores that managed to achieve a higher error rate than a random predictor were clamped at 1.

\subsection{Max trial group folds}\label{S4.3}
For datasets in our max-trial setting, we did not always have an equivalent number of labels in each training class. In these instances we drew k-folds based on the least present class $c'$ using the formula:

\begin{equation} \label{eq:9}
    k = \left\lceil\frac{1000}{2^{\log_{10}c'}}\right\rceil
\end{equation}

E.g., Canada crops coarse for which $c' = 75$, we drew $k = 273$ folds (Table \ref{table:canada-coarse}). When all classes had an equivalent number of training labels per-class, $k = 1$.

\subsection{Uncertainty estimation}\label{S4.4}
We use two complementary approaches to compute the error bars for evaluation metrics. For "low-shot" evaluation trials that do not use the entire training split, we perform k-fold cross validation and randomly sample K class-balanced training sets by subsampling the full training split, fit an independent predictor to each set and compute a normal distribution over metrics.

\begin{align} \label{eq:10}
    m_i =& \mathcal{M}(\{f_i(e_j), l_j\}_{j=0}^{M-1})
\end{align}

\begin{align} \label{eq:11}
    \bar{m} =& \frac{1}{K}\sum_{i=0}^{K - 1} m_i
\end{align}

\begin{align} \label{eq:12}
    s =& \sqrt{\frac{1}{K - 1}\sum_{i=0}^{K - 1}(m_i - \bar{m})^2}
\end{align}

where $m_i$ is the value of the metric $\mathcal{M}$ computed on the validation set for the predictor $f_i$ fitted to the $i$th training fold.

For "full" evaluation trials, where we use the entire training split to fit the downstream predictor, we instead use bootstrap statistics instead, resampling the validation split with replacement $B = 100$ times and computing statistics across the samples:

\begin{align} \label{eq:13}
    m_i =& \mathcal{M}(\{f(e_j), l_j, w_j^{(i)}\}_{j=0}^{M-1}) 
\end{align}

\begin{align} \label{eq:14}
    \bar{m} =& \frac{1}{B}\sum_{i=0}^{B - 1} m_i 
\end{align}

\begin{align} \label{eq:15}
    s =& \sqrt{\frac{1}{B - 1}\sum_{i=0}^{B - 1}(m_i - \bar{m})^2}
\end{align}

where  is a weight indicating how many times the $j$th validation example is included in the $i$th bootstrap sample (satisfying $\sum_{j=0}^{M-1} w_j^{(i)} = M$).

For smaller datasets where it was not possible to create a balanced training set with at least 200 examples in each class (or 200 examples overall for regression tasks), we used a combination of multiple training sets (with number of examples per class equal to the size of the smallest class) and nested bootstrap resampling of the validation split to estimate the statistics.

\section{Baseline comparisons}\label{S5}
We considered both widely adopted feature engineering approaches developed by the EO research community, as well as new deep learning approaches adapted for use with geospatial image inputs. We also included a set of controls to establish the predictive power of purely geographic information as well as a generic vision model trained on camera imagery (see Table \ref{table:baselines} for summary).

\begin{table*}
\footnotesize
\begin{tabular}{p{0.11\linewidth} p{0.12\linewidth} p{0.35\linewidth} p{0.08\linewidth} p{0.15\linewidth}}
\toprule
{\bf Approach} & {\bf Category} & {\bf Description} & {\bf Dims} & {\bf Inputs (m)} \\
\midrule
XY & Control & Latitude and longitude only & 4 & XY\\
\midrule
XYZ & Control & Latitude, longitude and elevation & 5 & XY,  elevation\\
\midrule
ViT & Control & Standard vision-transformer pre-trained on ImageNet & 1024 & S2 (RGB-only)\\
\midrule
Composites & Designed & Basic mean/median compositing of normalized EO image inputs & 16 & S1, S2, Landsat 8/9\\
\midrule
CCDC & Designed & Harmonic spectral-temporal features & 54 & Landsat 8/9 (all bands)\\
\midrule
MOSAIKS & Designed & Engineered embedding space & 1024 & S1, S2, Landsat 8/9\\
\midrule
SatCLIP & Learned & Implicit model designed for EO data & 256 & XY\\
\midrule
Prithvi & Learned & EO foundation model & 768 \- 2304 & Harmonized Landsat-Sentinel (HLS) L30\\
\midrule
Clay & Learned & EO foundation model & 768 & S1, S2, Landsat 8/9
\\
\bottomrule
\end{tabular}
\caption{Summary of baseline approaches compared with AlphaEarth embeddings.}
\label{table:baselines}
\end{table*}

Due to differences in handling of spatial, temporal, and channel dimensions and hard-coded data source requirements, it is not always straightforward to apply them to our benchmark or in a comparable manner.  In the following sections, we describe the process by which we obtain embeddings (or embedding-like feature vectors), and how those embeddings/features are extracted and aggregated. This process has to be redone for each temporal window or spatial extent under study by obtaining the correct EO data and running their model and/or assessing resulting feature vectors using our standardized evaluation framework and datasets.

To extract features/embeddings from baselines with outputs at a coarser spatial resolution than AEF, we bi-linearly resample spatial dimensions to 10m, and extract the embedding at the precise location of the evaluation dataset sample. This location was centered as much as possible in the geographic tile used for inference, and as the (longitude, latitude) coordinates of labels were kept at full precision, sub-pixel aliasing was taken into account for extraction.

As we were interested in assessing the extrapolation power given only sparse in-situ observations, for linear probes we did not attempt to fit a full-patch linear decoder to the ViT-based approaches that produced spatially coarse tokens (ViT, Prithvi, Clay). We instead fit a per-pixel linear model after bi-linearly resampling the tokens to 10m. To ensure equivalence to full-patch decoding, all evaluations concerned the same pixel location (center) such that we were consistently fitting one of the many linear combinations we would have fit had we used a full-patch decoder.

\subsection{Controls (not EO-specific)}\label{S5.1}
\subsubsection{XY}\label{S5.1.1}
The XY coordinate baseline assumes that the geospatial coordinates (i.e., longitude and latitude) of a given set of sparsely distributed geocoded labels or observations can be used to simply interpolate values spatially. This control essentially tests the hypothesis that location "Is All You Need”. We decompose polar latitude and longitude coordinates in degrees into sine and cosine components. This puts coordinate values in a continuous range of values, removing the discontinuity that would otherwise result at the antimeridian. We concatenate the results, and treat the resulting vectors as four-dimensional positional XY embeddings.

We note that XY is not time-varying, so we were unable to assess change detection evaluations and we generally expect poor performance on evaluations where landscapes are dynamic (e.g. agriculture).

\subsubsection{XYZ}\label{S5.1.2}
The XYZ control extends the XY baseline to a three-dimensional location coordinate by including elevation (height) information. Elevation plays a role in both regulating abiotic conditions and may act as a proxy for other terrain-driven process interactions, e.g., \citep{hof2012usefulness}. Therefore, the XYZ control tests the hypothesis that adding height information to positional encodings will improve predictive power. As with the XY baseline, we decompose latitude and longitude into trigonometric components. We also retrieve elevation information from the Copernicus GLO-30 DEM (see supplemental materials \ref{S1.8} for more details on this dataset). We mosaic individual images in the GLO-30 collection and select the elevation band (‘DEM’). Images are re-projected to WGS84 with a 1 arcsecond resolution for sampling. Elevation values are normalized based on the mean and standard deviation of the training sample elevations. We concatenate XY coordinates with sampled elevation, resulting in five-dimensional positional XYZ embeddings.

We note that XYZ is not time-varying, so we were unable to assess change detection evaluations and we generally expect poor performance on evaluations where landscapes are dynamic (e.g. agriculture).

\subsubsection{Vision Transformer (ViT)}\label{S5.1.3}
The Vision Transformer (ViT) is a popular deep learning architecture for computer vision. We use a ViT trained on the standard ImageNet benchmark of images and classification annotations \citep{dosovitskiy2020image} as a control. This is nominally a general-purpose model for computer vision, so we include it as a control to assess performance in comparison to systems designed specifically for EO data.

We choose the ViT-L/16 model architecture and parameters because it is popular for benchmarking and transfer learning in machine learning and computer vision papers, e.g. \citep{chen2021empirical}. As a pre-trained vision model for ImageNet, the ViT is limited in its resolution, bands, and handling of multi-temporal imagery. Standard ViTs only accept a single RGB color image, so we select the RGB bands of Sentinel-2 (L1C) normalized to the training statistics as input. These are the bands B2 (blue), B3 (green), and B4 (red). Each input image is embedded independently across time, and the outputs are masked using the input mask. Time is then collapsed by output averaging (weighted by masks). The result is a 1024-dimensional multi-temporal ViT embedding.

To maximize the performance of the ViT control, we tuned its hyperparameters to the evaluation set. We tried additional versions including one using annual composites, random initialization, and stacking all features through time. We found the version described here the most performant, and surprisingly more performant than non-controls in a number of instances.

\subsection{Designed EO features}\label{S5.2}
\subsubsection{Composites}\label{S5.2.1}
Rather than rely on reflectance values from any individual image acquisition, composite approaches combine observations from multiple image acquisitions to generate spatially continuous mosaics that optimize for pixel "quality", acquisition timing, and increasing signal-to-noise. Compositing approaches are fairly ubiquitous in modern remote sensing applications, i.e., \citep{qiu2023evaluation, francini2023assessment}. Median composites are fairly standard for optical imagery and tend to be preferred over mean composites because (a) medians preserve observed data values, and (b) are less sensitive to outliers, particularly clouds and cloud shadows which spectrally lie at very extreme bright and dark values. For radar imagery, mean composites tend to be more common as the goal is less-so to avoid outliers and select real data values, and more-so to smooth across many noisy observations varying with slight deviations in acquisition geometry. Composites are inherently lower-dimensional and orderless (compared with a stack of images which would be higher-dimensional and preserve the time order of observations), but serve as an important baseline for pure spectral information from minimally transformed de-noised/cloud-free image inputs.

We composite inputs across time by taking the median for optical sources (Sentinel-2, Landsat-8/9) and the mean for radar sources (Sentinel-1). Composite inputs are pre-processed using the same methods described in supplemental materials \ref{S1} and standardized by the AEF pretraining dataset statistics. Individual images are padded and masked as necessary to get constant input data dimensions. We filter by taking the observations in the valid period when we have them, and by taking the closest observations to the valid period bounds when we do not. Image sources are combined by concatenation, so the dimension of the composite feature space varies with the choice and number of sources. The number of channels is equal to the sum of the number of bands across sources, i.e., compositing in this way takes the input shape $B x T x H x W x D$ and makes the output shape $B x H x W x D'$. The final composite feature vector or “embedding” has 16 dimensions across the same Sentinel-1, Sentinel-2, and Landsat 8/9 bands used for AEF inference.

To maximize the performance of the composite baseline, we tuned its hyperparameters to the evaluation set. We tried additional versions including one using a consistently annual date range, mean rather than median compositing, versions that omitted all combinations of optical / radar sources, and versions that omitted masking. We found the version described here the most performant.

\subsubsection{Continuous Change Detection and Classification (CCDC)}\label{S5.2.2}
Harmonic curve-fitting has become an increasingly common approach for generating features that characterize spectral-temporal trajectories, e.g., \citep{wilson2018harmonic, pasquarella2018improved}. The most basic way to generate these sorts of harmonics would be simply fitting linear models to time series of EO data, e.g., \citep{wilson2018harmonic}. However, there are several more sophisticated temporal segmentation approaches that generate such features as part of a larger change detection workflows, i.e., Breaks For Additive Season and Trend (BFAST) \citep{verbesselt2010detecting}, Exponentially Weighted Moving Average Change Detection (EWMACD) \citep{brooks2014fly}, and the Continuous Change Detection and Classification (CCDC) \citep{zhu2014continuous} approach. We chose to use the CCDC approach because it is (a) well-known and widely used for remote sensing applications (see \cite{pasquarella2022demystifying} for a review) and (b) has already been run globally and surfaced as a dataset, making it one of the few existing examples of a model-as-dataset currently available globally in a cloud-computing environment.

We use precomputed CCDC features/parameters from the global Landsat-based Earth Engine collection available for 1999-2024 (“projects/CCDC/measures/v4”), which is an updated version of the Google Global Landsat-based CCDC Segments (1999-2019) dataset described in \cite{gorelick2023global}. CCDC coefficients are stored as variable-length arrays and are accessible as an Earth Engine Image Collection. We select the eight harmonic coefficients in the *\_coefs bands [$offset$, $t$, $cos({\omega}_t)$, $sin({\omega}t)$, $cos(2{\omega}t)$, $sin(2{\omega}t)$, $cos(3{\omega}t)$, $sin(3{\omega}t)]$ plus the *\_rmse values for seven Landsat bands (Blue, Green, Red, NIR, SWIR1, SWIR2, thermal). To match CCDC coefficients with a valid period, we choose the segment that intersects the valid period date for single-date evaluations and the middle of the specified valid period for monthly and annual evaluations. The final CCDC feature vector or “embedding” has 56 dimensions (8 coefficients for 7 Landsat bands).

\subsubsection{MOSAIKS}\label{S5.2.3}
Multi-task Observation using Satellite Imagery \& Kitchen Sinks (MOSAIKS) is a designed nonlinear representation of satellite imagery intended as analysis-ready data for accessible use and efficient computation across downstream tasks (70). It can be seen as a randomly-initialized linear combination of source data in a small sliding window with an additional non-linearity. The MOSAIKS approach provides a spatially localized representation of input satellite imagery at a specific time, bearing the same lack-of temporal constraints as composites. In particular it was first designed for RGB composite inputs and then generalized to RGB Sentinel-2 inputs. The MOSAIKS output is high dimensional and configurable (with a default of 8192 in the original implementation and 1024 in our reference implementation).

The original model described in \citep{rolf2021generalizable} was developed for only single-date RGB imagery. We reference the Microsoft Planetary Computer implementation \citep{mosaiks2021tutorial}, which generates random filters rather than selecting input patches. Specifically, we sample random convolutional filter parameters, once, for all inputs, convolve each input image with these random filters, stack the filter responses and their negatives to double the channels, apply a ReLU nonlinearity so the representation is not simply linear, and pool out the spatial dimensions for a vector embedding of each input.

We also extend our implementation to support multi-source output averaging where embeddings are generated for each source then averaged across the sources. This approach is preferred over concatenating embeddings for multiple sources as the specified embedding dimensionality is preserved regardless of the number of sources. Similarly, we accommodate multi-temporal embeddings by averaging outputs across time. The resulting multi-source, multi-temporal MOSAIKS embeddings have 1024 dimensions.

To maximize the performance of the MOSAIKS baseline, we tuned its hyperparameters to the evaluation set. We tried additional versions including one using a consistently annual input date range, versions that utilize composited inputs like those described in supplemental materials \ref{S5.2.1}, versions that omitted all combinations of optical / radar sources, versions with $[64, 128, 256, 1024, 8192]$ output features, and versions stacking all features through time. We found the version described here the most performant.

\subsection{Learned EO features}\label{S5.3}

\subsubsection{SatCLIP}\label{S5.3.1}
SatCLIP is a deep learning-based approach that takes inspiration from CLIP approach \citep{radford2021learning}, training location and image encoders via contrastive learning and matching images to their corresponding locations \citep{klemmer2025satclip}. SatCLIP models are trained on the pre-extracted Sentinel-2-100k dataset, a collection of 100 000 Sentinel-2 images that includes all available bands (B01, B02, B03, B04, B06, B06, B07, B08, B08A, B09, B11, B12) resampled to 10m resolution. The SatCLIP authors provide six pretrained SatCLIP models, trained with different vision encoders and spatial resolution hyperparameters.

We evaluate only the SatCLIP Vit16-L40 model, which outperforms other versions according to the SatCLIP paper \citep{klemmer2025satclip}. During evaluation, we use only the location encoder, following the procedure suggested by the authors. No specific preprocessing of the locations is required beforehand, i.e., the location encoder takes longitude and latitude. Generated SatCLIP embeddings have 256 dimensions. We note that SatCLIP is not time-varying, so we were unable to assess change detection evaluations and we generally expect poor performance on evaluations where landscapes are dynamic (e.g. agriculture).

\subsubsection{Prithvi}\label{S5.3.2}
Prithvi is a temporal pre-trained ViT-based architecture trained in a fashion similar to \cite{cong2022satmae} on Harmonized Landsat Sentinel (HLS) L30 imagery with 30m nominal scale collected over the contiguous United States during 2017 \citep{jakubik2023foundation, Jakubik2023prithvi}. Input videos are limited to three in the model version provided by the authors. Prithvi is considered a geospatial foundation model, the encoder output from which can be used to solve various downstream tasks. While Prithvi authors only explicitly suggest its use with additional deep-learning decoders, it is common to transfer ViT-like architectures via linear decoding as we do here.

We use the Prithvi 1.0 model and adapt the approach described in the Prithvi-specific example HLS Multi-temporal Crop Classification Model \citep{Li2023classification}, where encoder outputs are used as embeddings in a crop classification task. The multi-temporal embedding dimension is equal to 768 multiplied by the number of frames available (1 to 3) yielding 768 to 2304 dimensions.

We retrieve input data from the HLSL30: HLS-2 Landsat Operational Land Imager Surface Reflectance and TOA Brightness Daily Global 30m collection \citep{Masek2021hls} on Earth Engine ("NASA/HLS/HLSL30/v002"). We sample patches of 224x224 at a 30-meter spatial resolution. We fetch all images under a specified cloud cover threshold (20\% or 50\%) for the support period and incrementally increase the frame (into the past) by the HLS maximum revisit period until we find an available image. We use a cloud coverage threshold of 20\% for all evaluation datasets with the exception of the Descals evaluation dataset, where we use a higher (50\%) cloud coverage threshold due to low imagery availability. In rare instances (< 1\% for any given dataset), the cloud coverage-based filtering yields no data, and in these cases we set embeddings for such entries as the expectation over the computed embeddings for a given dataset.

\subsubsection{Clay}\label{S5.3.3}
The Clay Foundation Model is more akin to a traditional ViT differentiated by the use of input metadata at inference time. This metadata includes nominal resolution, the geographic centroid of the input, a source encoding derived from a wavelength associated with the bandpass or transmission when applicable, and the observation timestamp \citep{Clay2024model}. Generating semantic embeddings for any location and time is a stated use case for the Clay model, and classification, regression, and detecting changes over time are suggested use-cases by the authors.

We sample 256x256 pixel images with a nominal scale of 10 meters for Sentinel-2 / Sentinel-1, and 30 meters for Landsat 8/9 per author-provided instructions. We filter out Landsat and Sentinel-2 images with cloud coverage exceeding 20\%, after that we limit the number of entries per time frame to fit them into NVIDIA V100 RAM (30 frames maximum). Input images are normalized using statistics unique to each evaluation's training split. To obtain the final embedding over the valid period, we average over per-source per-time spatial tokens for a final dimensionality of 768. When no imagery for a particular source is available within the valid period, we follow the procedure used for HLS detailed in supplemental materials \ref{S5.3.2}.

There are a number of suggested mechanisms for extracting embeddings from The Clay Foundation Model. For example, the authors suggest taking the class token as a patch embedding, and in others, the non-class tokens are averaged. To tune Clay hyperparameters to our evaluation set, we tried a number of Clay variants. These included using the class token for a patch representation and versions that omitted combinations of optical / radar sources. We found the version described here the most performant.

\section{Additional results \& discussion}\label{S6}
\subsection{Comparisons with 10 and 1 samples per-class}\label{S6.1}
In our extreme low-shot trials, overall performance of methods was close to random chance in many cases. For 500-fold 10-shot trials, AEF error reductions were only > 1.0x in the \textasciitilde 90\% confidence interval for 8/15 evaluations with an average range of variation $\pm0.38\text{x}$, and for 1000-fold 1-shot trials, AEF error reductions were only $> 1.0\text{x}$ in the \textasciitilde 90\% confidence interval in 5/15 evaluations with average variation $\pm0.49\text{x}$. While the mean gain was in AEF's favor for both settings, we consider the extreme degree of variability indicative that adequate general 1-shot or 10-shot performance remains an unsolved research frontier.

\subsection{Classification}\label{S6.2}
\begin{figure*}
    \includegraphics[width=\textwidth]{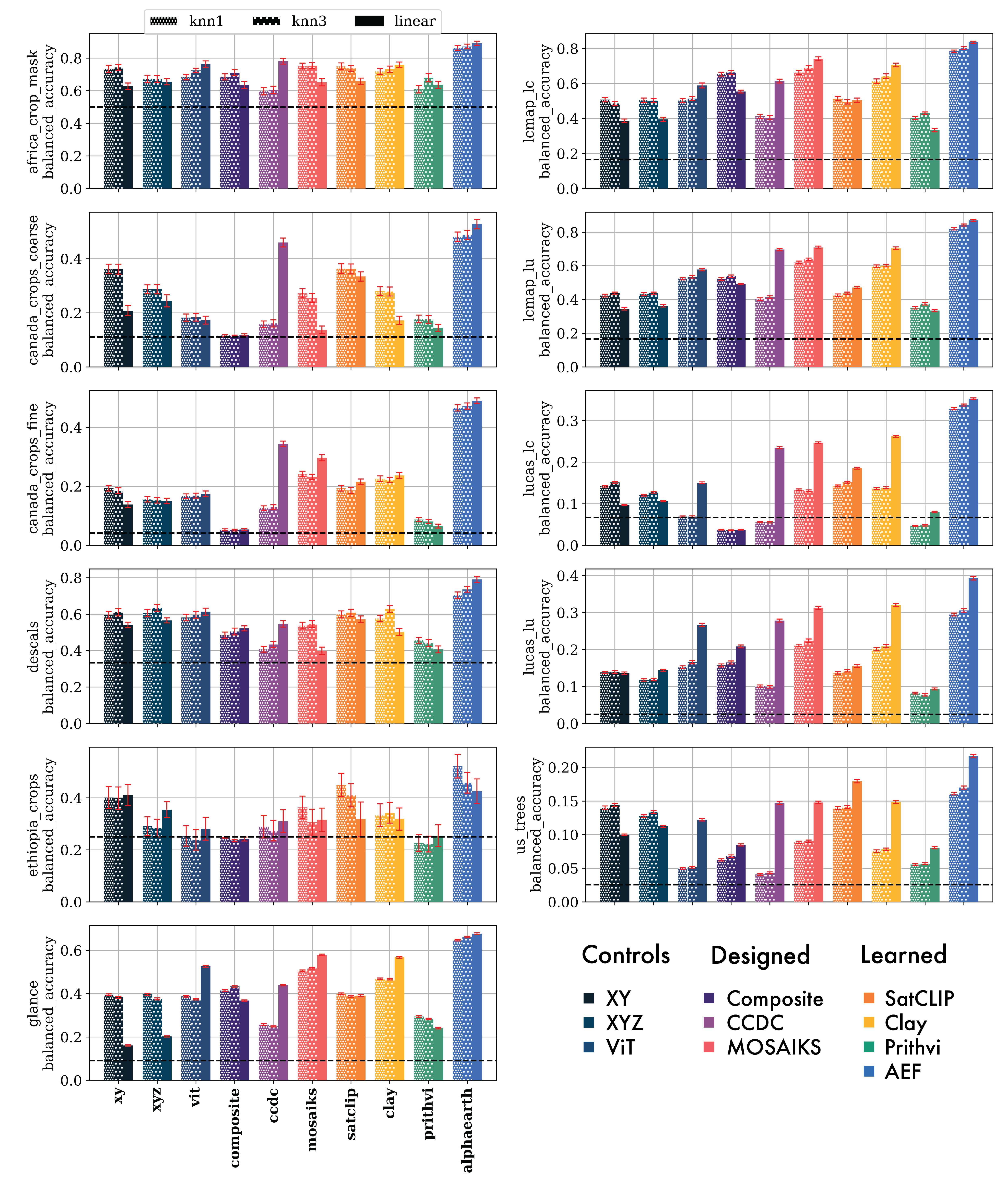}
    \caption{Classification results reported in terms of balanced accuracy. Black dotted line indicates expected accuracy given random chance / number of classes. Error bars indicate $1\sigma$ ($68.27\%$) confidence interval by bootstrapping or bootstrapping and k-folds for small datasets, e.g., ethiopia\_crops, canada\_crops\_*.}
    \label{fig:classification}
\end{figure*}

AEF showed consistently strong performance across all classification evaluations, while the next-best approach varies considerably with some approaches performing no better than the null expectation for some datasets (Figure \ref{fig:classification}). This result indicates the utility of AEF as a general-purpose feature space suitable for a multitude of classification problems ranging from simple, binary legends to detailed land use, land cover, and even genus-level tree species mapping. While further iteration on training labels and/or secondary predictors may improve absolute accuracies, AEF shows previously unachievable performance in low-shot regimes with simple classifiers, unlocking use cases that may have previously been untenable given sparse observational records and/or highly detailed taxonomies.

For the applications considered, AEF often offers a notable improvement over other archetypal examples of other designed and learned feature spaces. CCDC harmonics were the next-best predictors for Canada crops (coarse and fine) and Africa crop mask, suggesting spectral-temporal information characterizing phenology is more important than spatial resolution or multi-sensor inputs for these crop-mapping applications. SatCLIP was the next-best approach for Ethiopia crops and US trees, evaluations we would expect should benefit phenological information. Preference for SatCLIP here suggests encoding localized EO features was beneficial for these tasks, and improvements for SatCLIP relative to the coordinate and ViT controls indicate that these gains can be attributed to inclusion of EO-specific information content. We find that MOSAIKS is the next-best predictor for GLancE and LCMAP land cover, while Clay for LUCAS land use and land cover. Preference for MOSAIKS indicates spatial context provides valuable information for land cover mapping at both national and global scales, as it would for Clay which also does not have access to multitemporal information.

Interestingly, we found that in some cases, controls were selected as the second-best approach, specifically the XYZ control for the Descals oil palm evaluation and the ViT for the LCMAP land use evaluation. This suggests that AEF is more effectively leveraging EO-specific data sources to generate learned representations than other baselines that do not consistently outperform controls. In general, we found Prithvi to exhibit poor performance. This is not particularly surprising given the model is not explicitly designed to provide a feature space, and confirms that Prithvi is not well-suited for this sort of low-shot classification and requires additional fine-tuning. However, it is interesting to note the lack of parity with the ViT control that has not been trained on EO data, does not have access to multitemporal information, nor is intended to function as a feature space.

Of the low-shot classification methods considered, AEF features typically exhibited the greatest balanced accuracies for the linear classifier experiments for the max trial groups, with the exception of Ethiopia crops, where kNN with k=1 is preferred (Figure \ref{fig:classification}). We believe the Ethiopia crops evaluation is particularly challenging given its extreme sparsity and fine scale, and so simple nearest neighbor classification was the best method of transfer for a handful of methods. Given the generally poor performance of all methods on this four class classification problem, there are opportunities to improve performance on this type of dataset.

\subsection{Regression}\label{S6.3}
\begin{figure}
    \includegraphics[width=\columnwidth]{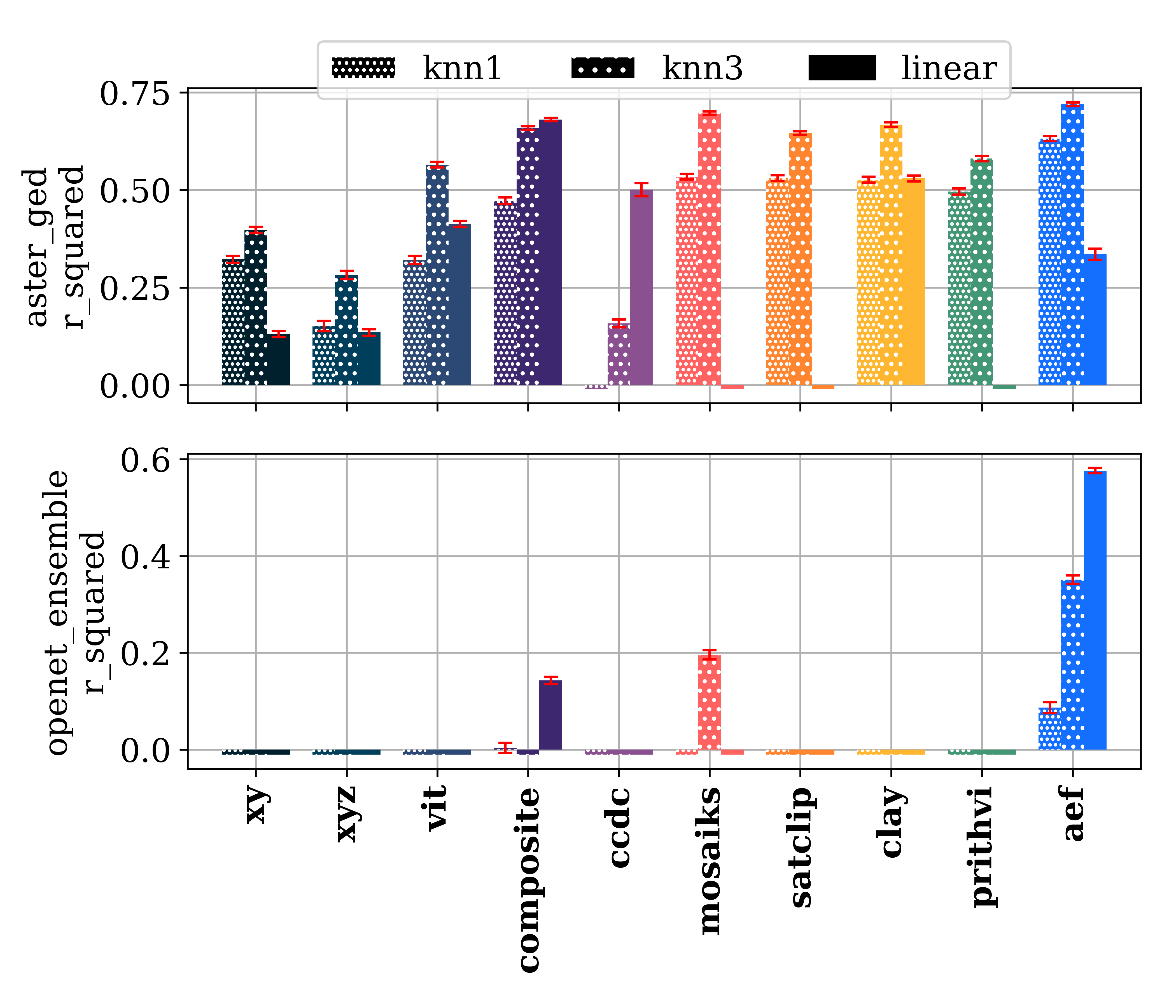}
    \caption{Regression results reported in terms of mean $R^2$ values. Error bars indicate $1\sigma$ (68.27\%) confidence interval by bootstrapping. Negative $R^2$ estimates were clamped to zero for visualization purposes.}
    \label{fig:regression-r2}
\end{figure}

\begin{figure}
    \includegraphics[width=\columnwidth]{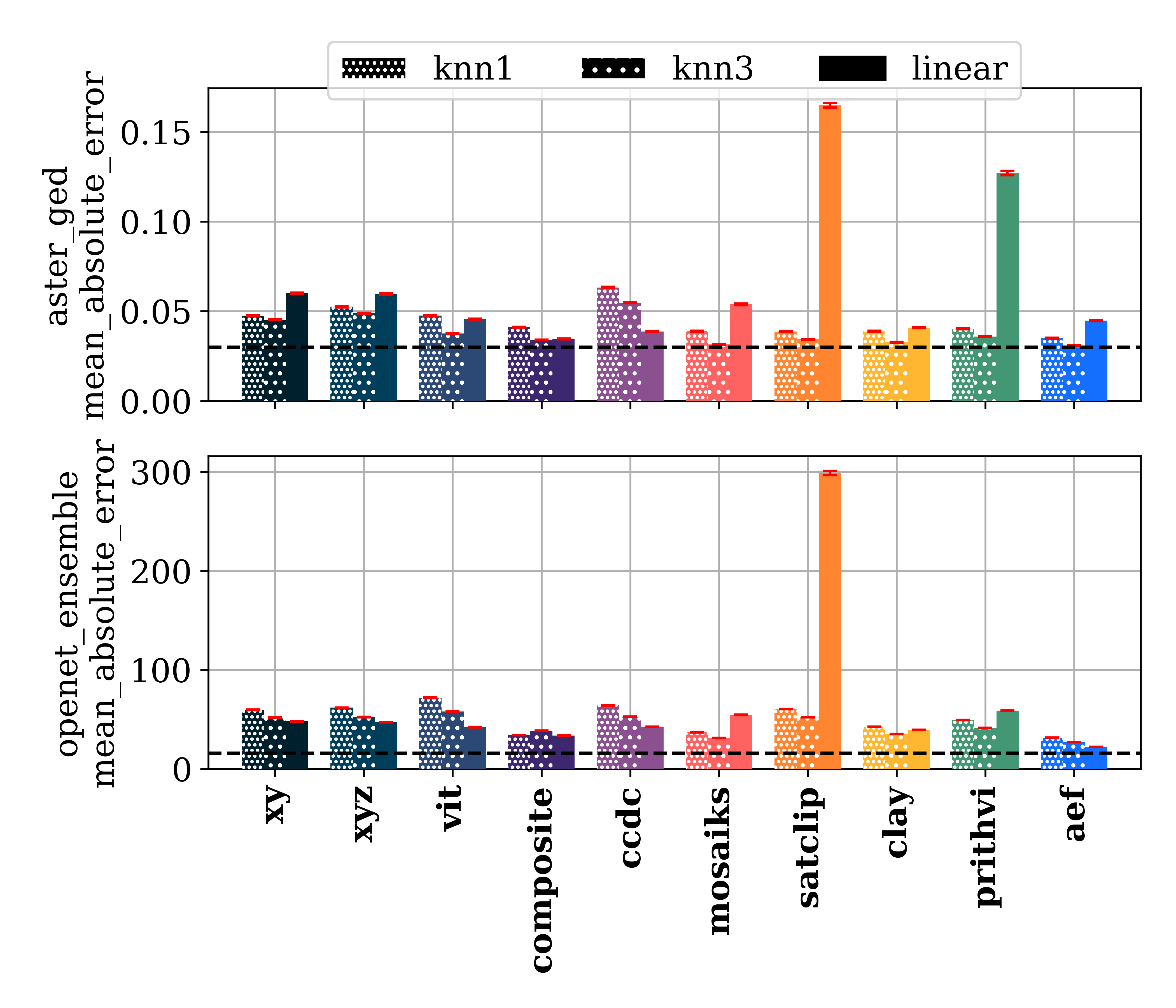}
    \caption{ Regression results reported in terms of Mean Absolute Error (MAE). Error bars indicate $1\sigma$ (68.27\%) confidence interval by bootstrapping. Dotted line represents approximate expectation of error from product publications.}
    \label{fig:regression-mae}
\end{figure}

For regression tasks, AEF again exhibited the best overall performance in terms of both gains in $R^2$ and reductions in MAE (Figures \ref{fig:regression-r2} and \ref{fig:regression-mae}). We find that $R^2$ values for AEF are always within a valid range, noting that several other approaches produced negative $R^2$ values i.e. worse-than-null performance for ASTER GED (emissivity prediction). OpenET ensemble (evapotranspiration prediction) was evidently more challenging where almost all other methods had negative values (clipped to $-0.01$ for plotting purposes; Figure \ref{fig:regression-r2}). 

For predicting ASTER emissivity values, MOSAIKS is the next-best approach and composites generally show strong performance, while spectral-temporal CCDC features and XY and XYZ positional encodings are demonstrably worse. This indicates that local spatial patterns and overall reflectance (as opposed to seasonality in reflectance) are important factors for this use case.

When evaluating on the OpenET Ensemble dataset, we found that AEF was the only approach to produce viable results using all kNN and linear predictors considered. Of the other baselines, composites with linear probes and MOSAIKS with knn with k=3 produced the only other viable results in terms of having $R^2$ values greater than 0 (i.e., greater variance explained than the mean). Looking at results in terms of MAE where lower values indicate better performance, we find AEF has an error rate in line with what would be expected for both ASTER GED and OpenET performance expectations for the data products these evaluation datasets were sampled from (Figure  \ref{fig:regression-mae}). Here we can more clearly see variability in performance, with particularly large errors for SatCLIP and Prithvi linear trials.

Though we note the small sample use cases represented here, we do note that continuous measurements, as opposed to discrete categorical labels, are often associated with field-based observational datasets, and we expect to see external comparisons including AEF on additional regression problems in the future following our data release (see Global embeddings dataset section).

\subsection{Change detection}\label{S6.4}
\begin{figure}
    \includegraphics[width=\columnwidth]{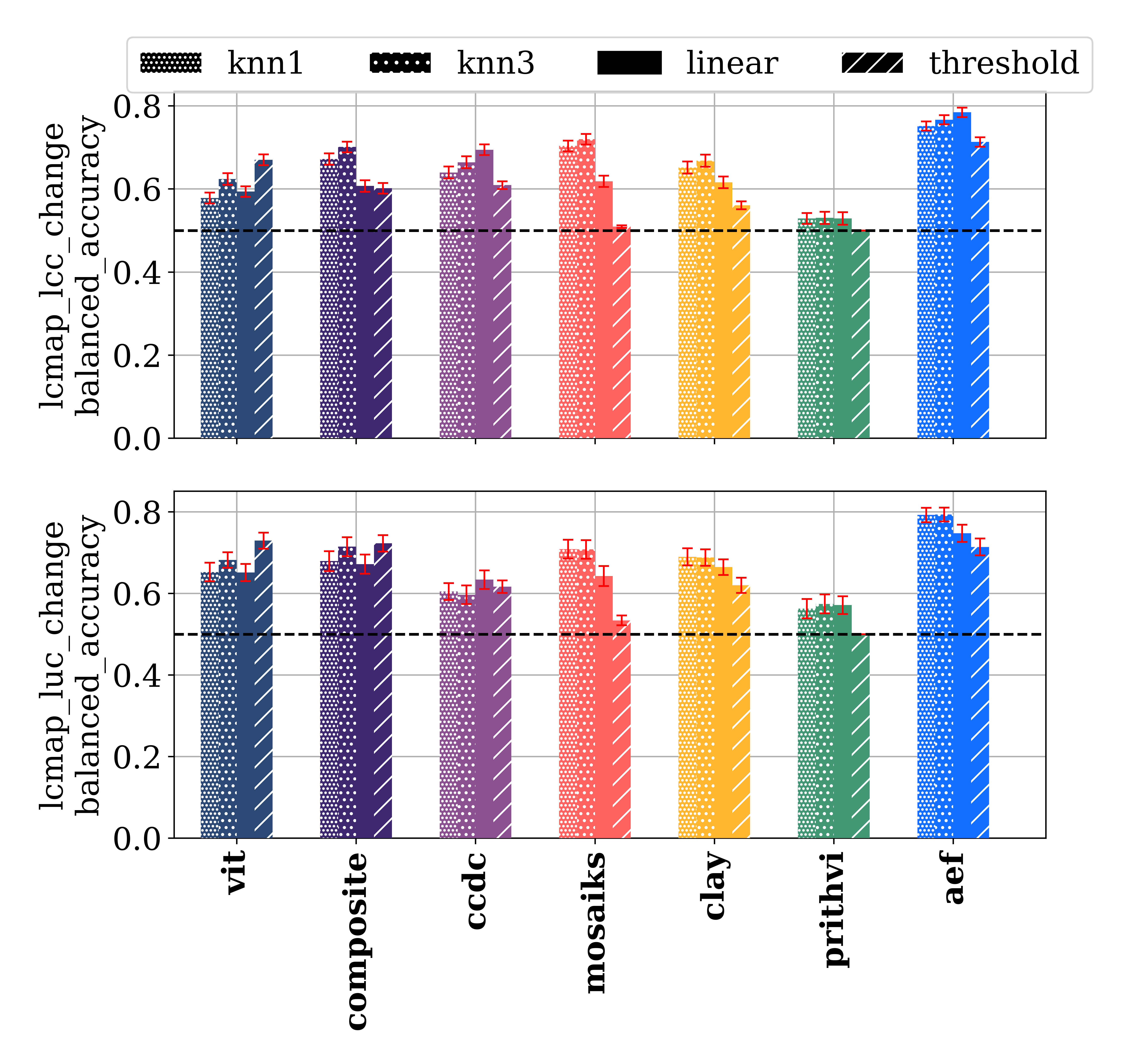}
    \caption{Change detection results reported in terms of balanced accuracy. The black dotted-line indicates random chance for classification evaluations given the number of classes. Error bars indicate $1\sigma$ (68.27\%)confidence interval by bootstrapping.}
    \label{fig:changedetection}
\end{figure}

The SatCLIP, XY, and XYZ baselines were omitted from change detection comparisons given these approaches are location-only, i.e., have no time handling or way to differentiate between observations at different times. We find generally less differentiation in performance across methods, with the exception of Prithvi and linear trials of MOSAIKS, which perform at or near the random baseline (Figure \ref{fig:changedetection}).

\section{Ablations}\label{S7}
\subsection{Training observations}\label{S7.1}
We share a full set of plots detailing the performance scaling of AEF relative to other learned baselines as additional observations are added (Figure \ref{fig:ablate-sources}). Baselines were omitted when $R^2$  values were < 0, or for change detection evaluations when the baseline was not time-varying (SatCLIP). We note no obvious trend among the baselines, indicating methodologies that perform better on some problems more so than others. AEF generally improves monotonically as additional observations are added for 9-of-15 evaluation datasets. There is some unpredictable non-monotonicity for some evals, but we do not identify any obvious grouping or regional bias.

\begin{figure*}
    \includegraphics[width=\textwidth]{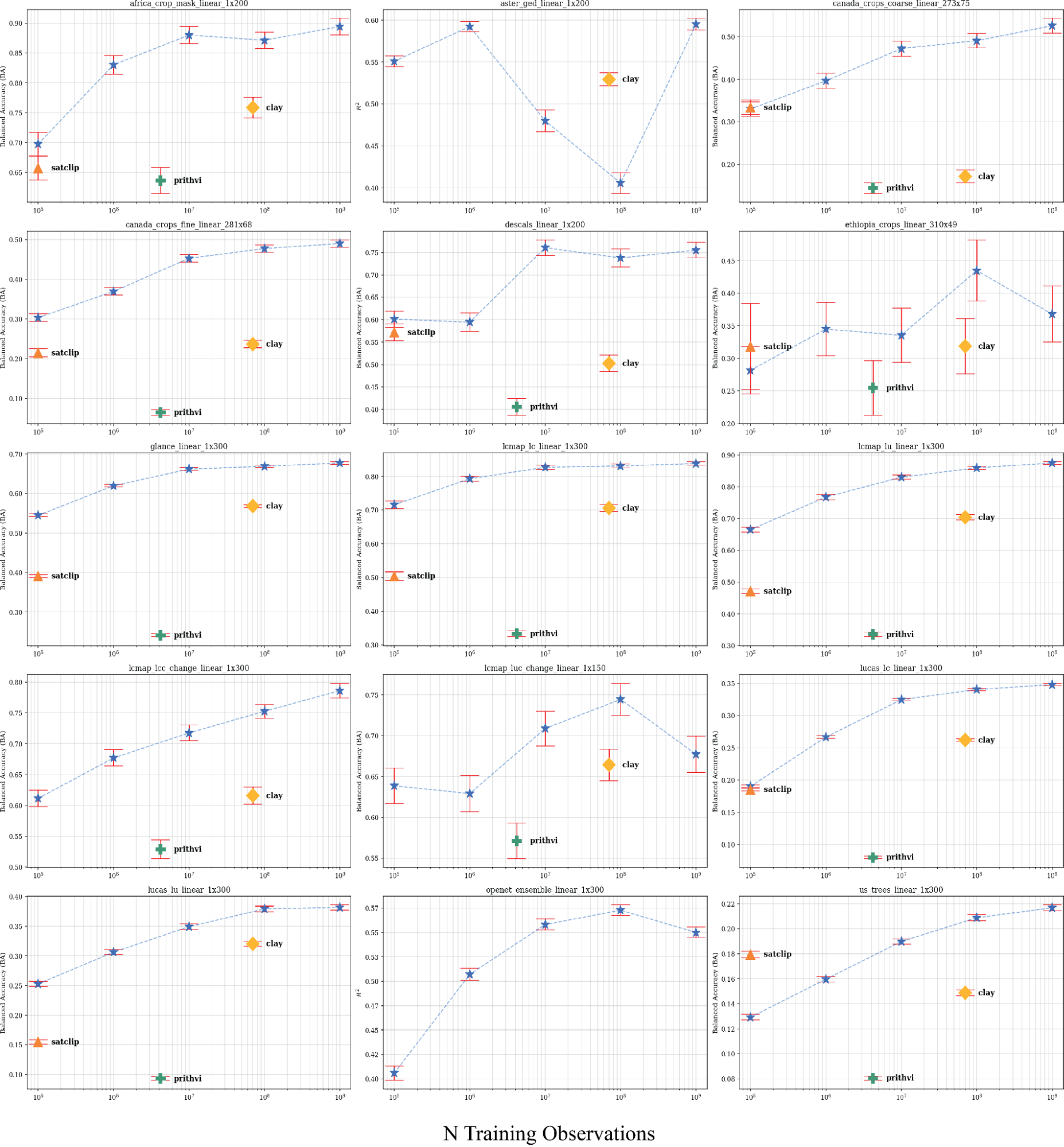}
    \caption{Effects of scaling observations for evals in linear probe max-trial regime. Error bars indicate $1\sigma$ BA / $R^2$ or \textasciitilde 68.27\% confidence interval by bootstrapping and k-folds when possible. AEF is represented by the dotted line and star markers.}
    \label{fig:ablate-obs}
\end{figure*}

\subsection{Sources}\label{S7.2}

\begin{figure*}
    \includegraphics[width=\textwidth]{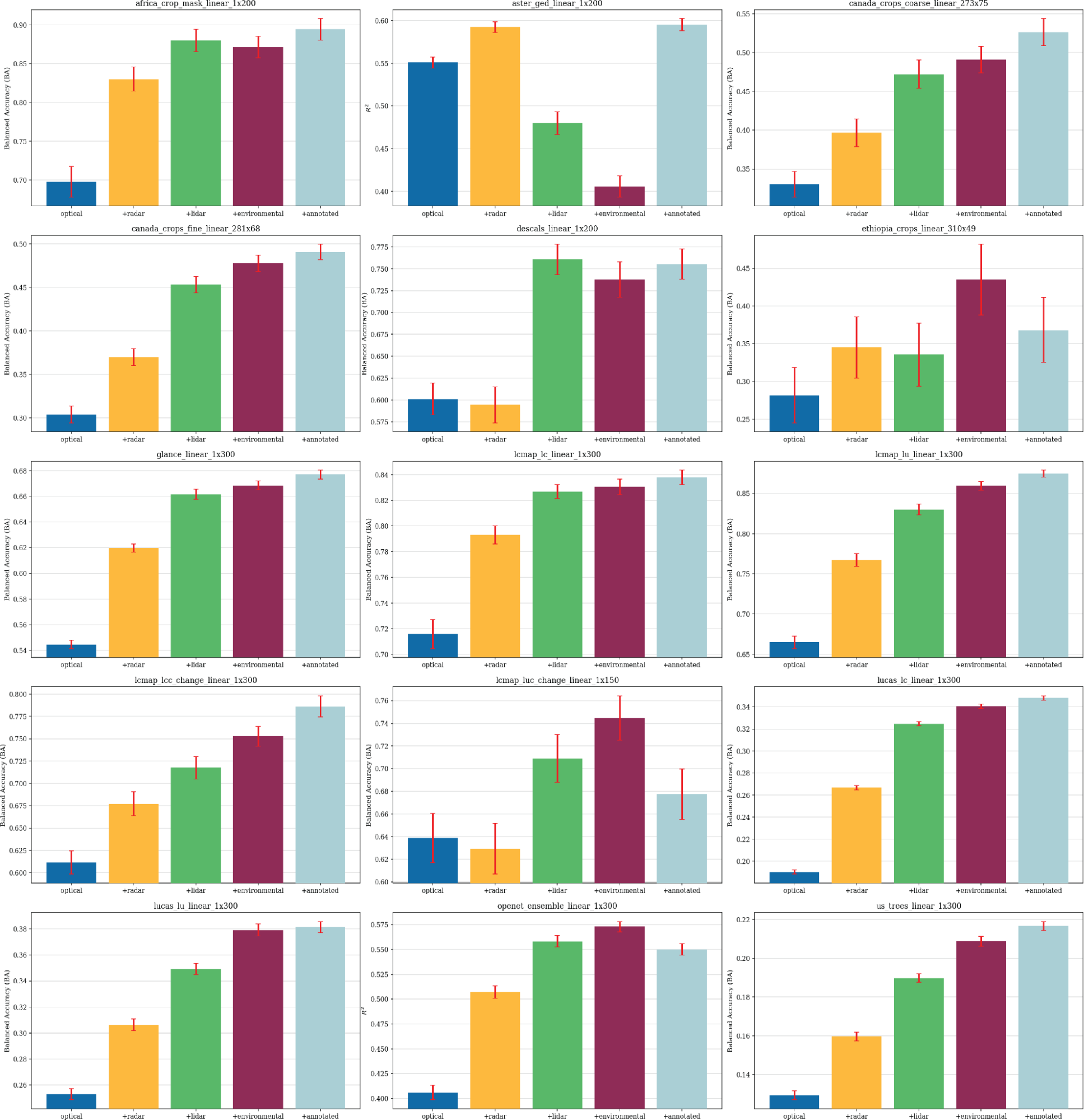}
    \caption{Effects of additional source groups for evals in linear probe max-trial regime. Error bars indicate $1\sigma$ BA / $R^2$ or \textasciitilde 68.27\% confidence interval by bootstrapping and k-folds when possible. The London-fog-blue bar to the far right matches the version of AEF used in other comparisons.}
    \label{fig:ablate-sources}
\end{figure*}

We share a full set of plots detailing the performance of AEF relative to ablated by source groups. The groups are as follows: Optical (Sentinel-2, Landsat-8 / Landsat-9), Radar (Sentinel-1, PALSAR2 ScanSAR), LiDAR (GEDI), Environmental (GLO 30, ERA5 Land, GRACE), and Annotated (NLCD, Wikipedia) (Figure \ref{fig:ablate-sources}). We note a variety of patterns characterized by source groups, though 11-of-15 evaluations were most performant with all groups. Evidently, different evaluations "prefer" different types of measurements; in some cases this is expected e.g. the Descals oil palm evaluation is most performant with Optical + Radar + LiDAR as these groups offer the most information regarding sub-canopy structure, where climatic variables and free-form text annotations / land-cover labels are less informative. Unsurprisingly, all LULC evaluations (excluding change) are most performant with all groups including the Annotated group, though interestingly this does not lead to the biggest performance gain compared to adding radar and lidar data. 

\subsection{Bottleneck characteristics}\label{S7.3}

\begin{figure*}
    % \centering
    \includegraphics[width=\textwidth]{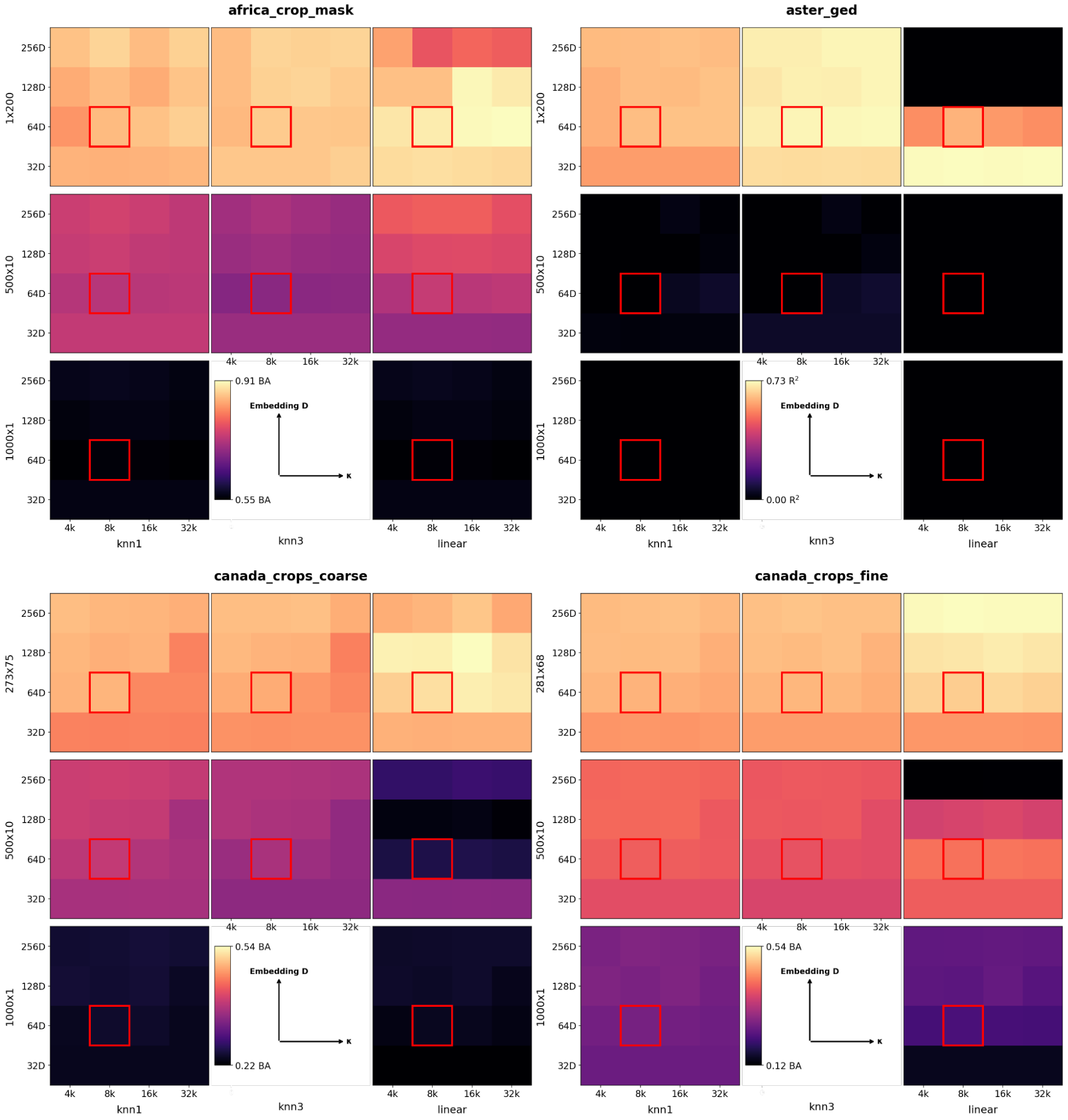}
    \caption{Evaluation performance as a function of embedding dimension (Embedding $D$) and VMF kappa ($\kappa$) for all trial sizes (1 shot, 10 shot, and max shot) and methods of transfer (nearest neighbors for k=1, k=3, and linear probe). The red square indicates the parameter setting (Embedding $D = 64$, $\kappa = 8\kappa$) used for AEF.}
    \label{fig:ablate-bottleneck}
\end{figure*}%
\begin{figure*}\ContinuedFloat
    % \centering
    \includegraphics[width=\textwidth]{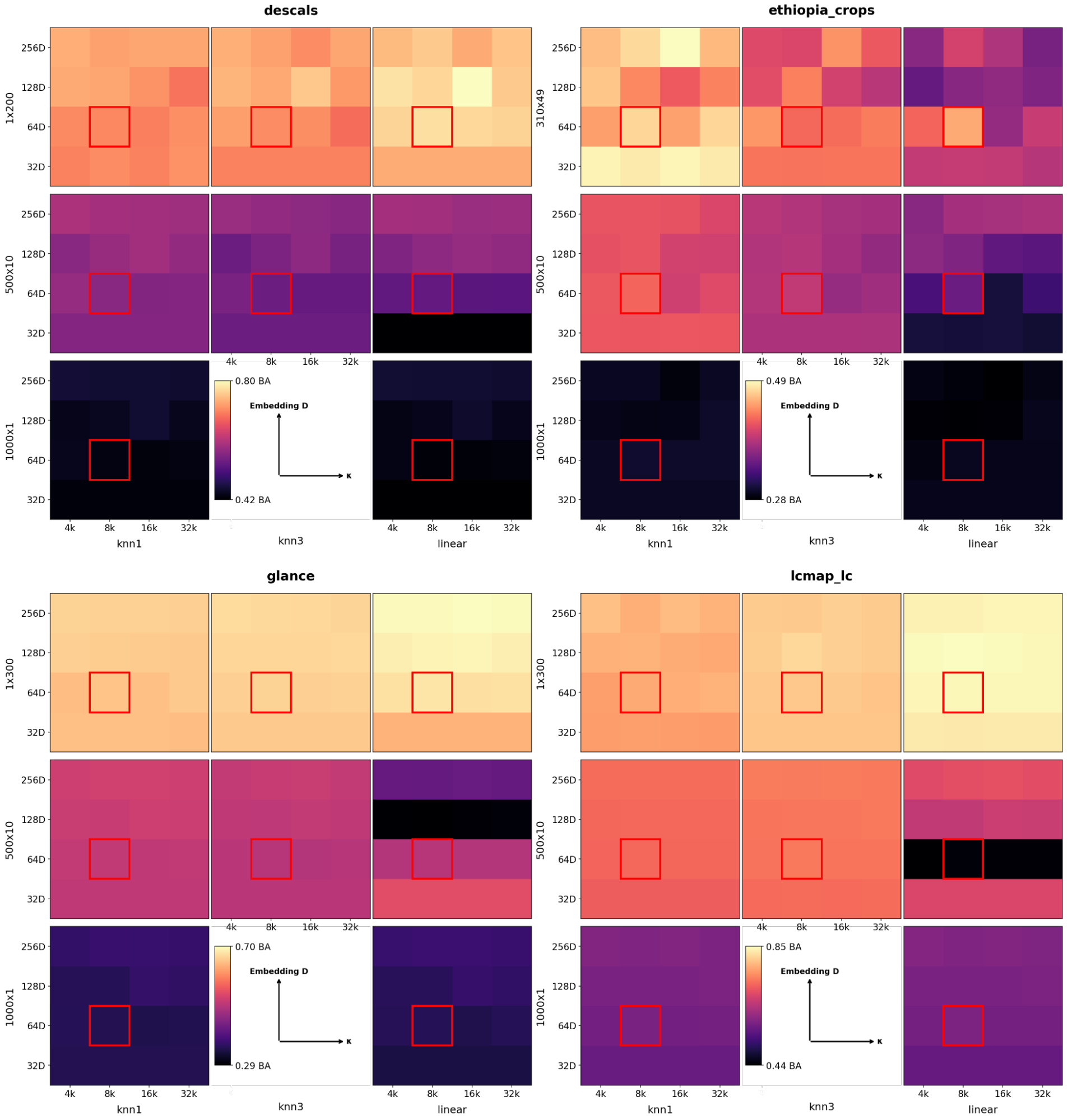}
    \caption{(con't) Evaluation performance as a function of embedding dimension (Embedding $D$) and VMF kappa ($\kappa$) for all trial sizes (1 shot, 10 shot, and max shot) and methods of transfer (nearest neighbors for k=1, k=3, and linear probe). The red square indicates the parameter setting (Embedding $D = 64$, $\kappa = 8\kappa$) used for AEF.}
\end{figure*}%
\begin{figure*}\ContinuedFloat
    % \centering
    \includegraphics[width=\textwidth]{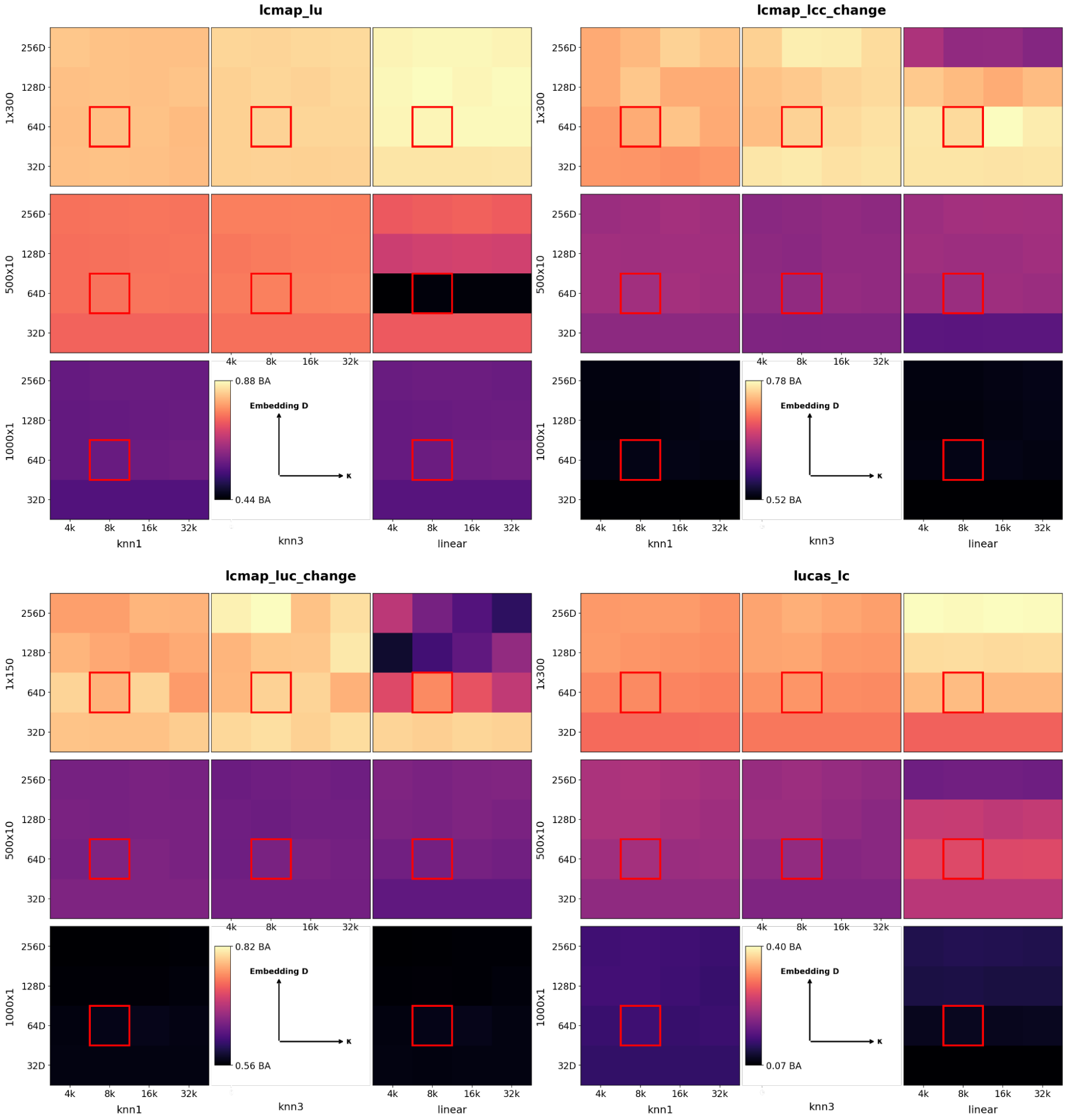}
    \caption{(con't) Evaluation performance as a function of embedding dimension (Embedding $D$) and VMF kappa ($\kappa$) for all trial sizes (1 shot, 10 shot, and max shot) and methods of transfer (nearest neighbors for k=1, k=3, and linear probe). The red square indicates the parameter setting (Embedding $D = 64$, $\kappa = 8\kappa$) used for AEF.}
\end{figure*}%
\begin{figure*}\ContinuedFloat
    \centering
    \includegraphics[width=\textwidth]{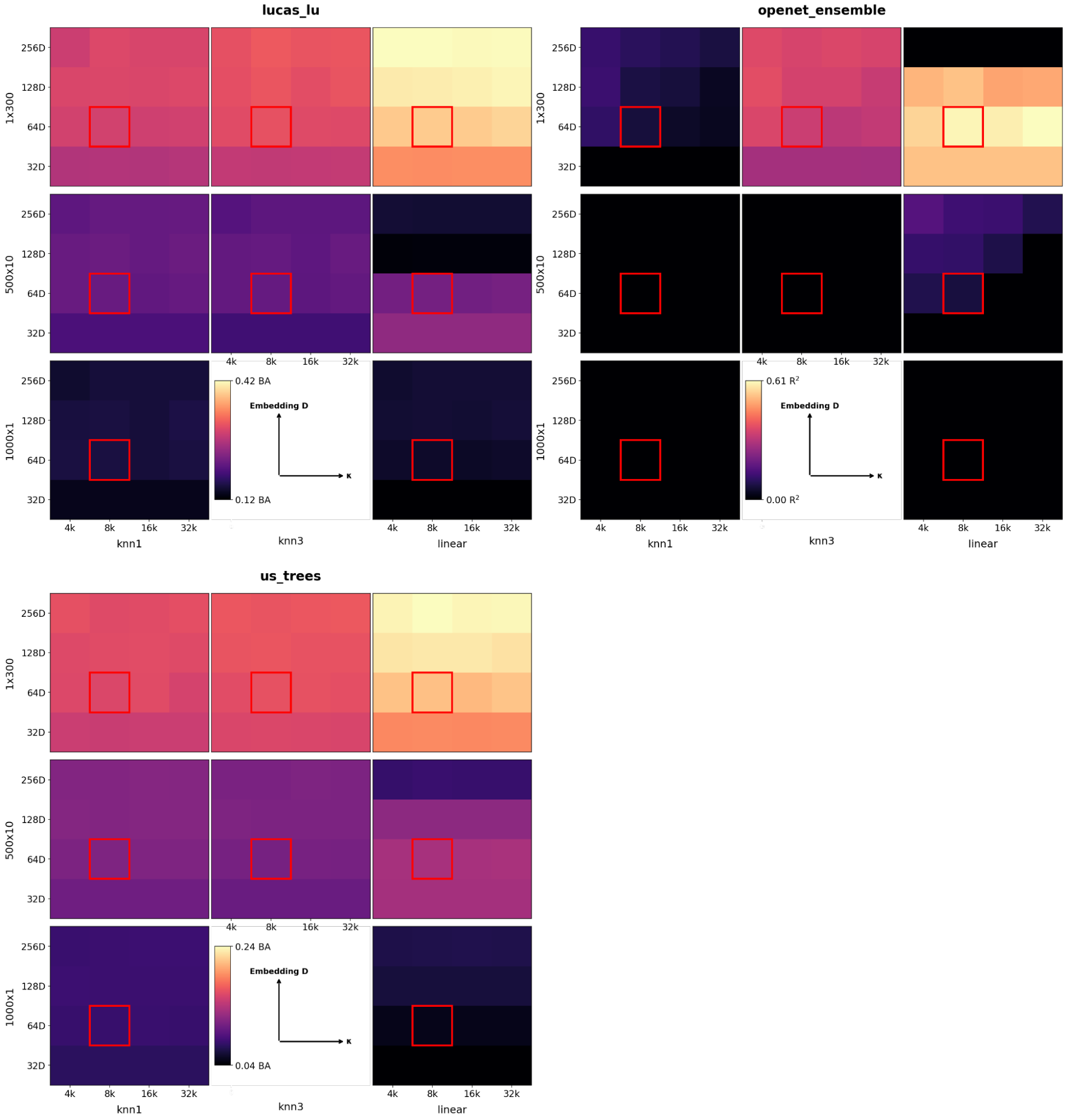}
    \caption{(con't) Evaluation performance as a function of embedding dimension (Embedding $D$) and VMF kappa ($\kappa$) for all trial sizes (1 shot, 10 shot, and max shot) and methods of transfer (nearest neighbors for k=1, k=3, and linear probe). The red square indicates the parameter setting (Embedding $D = 64$, $\kappa = 8\kappa$) used for AEF.}
\end{figure*}

AEF's reconstruction task relies on a noisy bottleneck to compress and extrapolate information from the sparse input sequence. Two model hyperparameters, the embedding dimension, and the channel noise parameterized by VMF $\kappa$, directly affect the capacity of this bottleneck. Lower settings of $\kappa$ in particular will also affect the smoothness of the latent (embedding) manifold due to the regularizing effect of the noise \citep{kingma2013auto}. This last property is desirable when using embeddings for nearest neighbor retrieval as distances measured along a smooth (lower dimensional) manifold are more meaningful than those in a higher dimensional space (assuming the manifold hypothesis \citep{fefferman2016testing}. Therefore contention exists between a smoother embedding space (lower $\kappa$), and the information capacity of the channel and therefore embedding (higher $\kappa$). We explore this in the context of embedding dimension across all methods of transfer and trial group sizes in Figure \ref{fig:ablate-bottleneck}. The setting used for AEF was Embedding $D = 64$, $\kappa = 8\text{e}{3}$.

We find that performance as a function of bottleneck characteristics varies considerably depending on the evaluation dataset. One expected trend emerges for datasets with larger legends compared to e.g. two or three class classification problems in the max-trial setting: Canada crops (fine), LUCAS derived datasets, and iNaturalist trees all tend to perform better with more concentrated noise and higher embedding dimensions. We also, expectedly, find that a noisier bottleneck seems to improve, or not impact, the performance in the smaller trial groups (500x10, 1000x1). This is most evident in Canada crops derived datasets and OpenET ensemble.

\section{Inference}\label{S8}

\subsection{Quantization}\label{S8.1}
To reduce the storage and compute requirements for working with our released embedding field data, we opted to test a number of post-quantization schemes. We tried quantizing 32-bit float values to signed 8-bit and 16-bit integers using a method identical to the following pseudo-JAX method:

\definecolor{kwstyle}{rgb}{0.2,0.0,0.7}
\definecolor{backcolor}{rgb}{0.97,0.97,0.97}

\lstset{ 
    backgroundcolor=\color{backcolor},   
    keywordstyle=\color{kwstyle},
    basicstyle=\ttfamily\tiny,
    breakatwhitespace=false,         
    breaklines=true,                 
    captionpos=b,                    
    keepspaces=true,                 
    showspaces=false,                
    showstringspaces=false,
    showtabs=false,                  
    tabsize=2
}
\lstset{language=Python}

\begin{lstlisting}[frame=single]
def quantize(
    x: chex.Array,
    power: float, 
    scale: float, 
    min_value: float, 
    max_value: float, 
    quantization_type: jnp.dtype) -> chex.Array:
  
  sat = jnp.abs(x) ** (1 / power) * jnp.sign(x)
  snapped = jnp.round(sat * scale)
 
  return jnp.clip(
      snapped, 
      min_value, 
      max_value
  ).astype(quantization_type)
\end{lstlisting}

Values were dequantized using a method identical to the following pseudo-JAX method:

\begin{lstlisting}[frame=single]
def dequantize(
    y: chex.Array,
    power: float, 
    scale: float, 
    quantization_type: jnp.dtype) -> chex.Array:

  rescaled = y.astype(jnp.float32) / scale
  return (
      jnp.abs(rescaled) ** power * jnp.sign(rescaled)
  )
\end{lstlisting}

The exponentiation was introduced to preserve information in the least significant digits of dequantized values. We used the following scale, minimum, and maximum values according to Table \ref{table:quantization}.

\begin{table}
\footnotesize
\centering
\begin{tabular}{p{0.25\linewidth} p{0.18\linewidth} p{0.15\linewidth} p{0.15\linewidth}}
\toprule
{\bf Integer type} & {\bf Scale} & {\bf Min value} & {\bf  Max value} \\
\midrule
int8 (s8) & 127.5 & \-127 & 127 \\
\midrule
int16 (s16) & 32767.5 & \-32767 & 32767 \\
\bottomrule
\end{tabular}
\caption{Quantization parameters.}
\label{table:quantization}
\end{table}

We did not initially assume that 8-bits ought to be enough for any evaluation given that quantization was not part of the learning process, we were nonetheless pleased to note little performance variability compared to the non-quantized embeddings as shown in Figure \ref{fig:quantization}A-C. As quantization was not part of our comparisons, we chose to quantize to 8-bits with power = 2 as this gave the best storage / performance tradeoff based on evaluation performance. We believe our comparison results with this quantization strategy would be largely the same as without.

\begin{figure}
    \begin{subfigure}{\columnwidth}
        \includegraphics[width=\columnwidth]{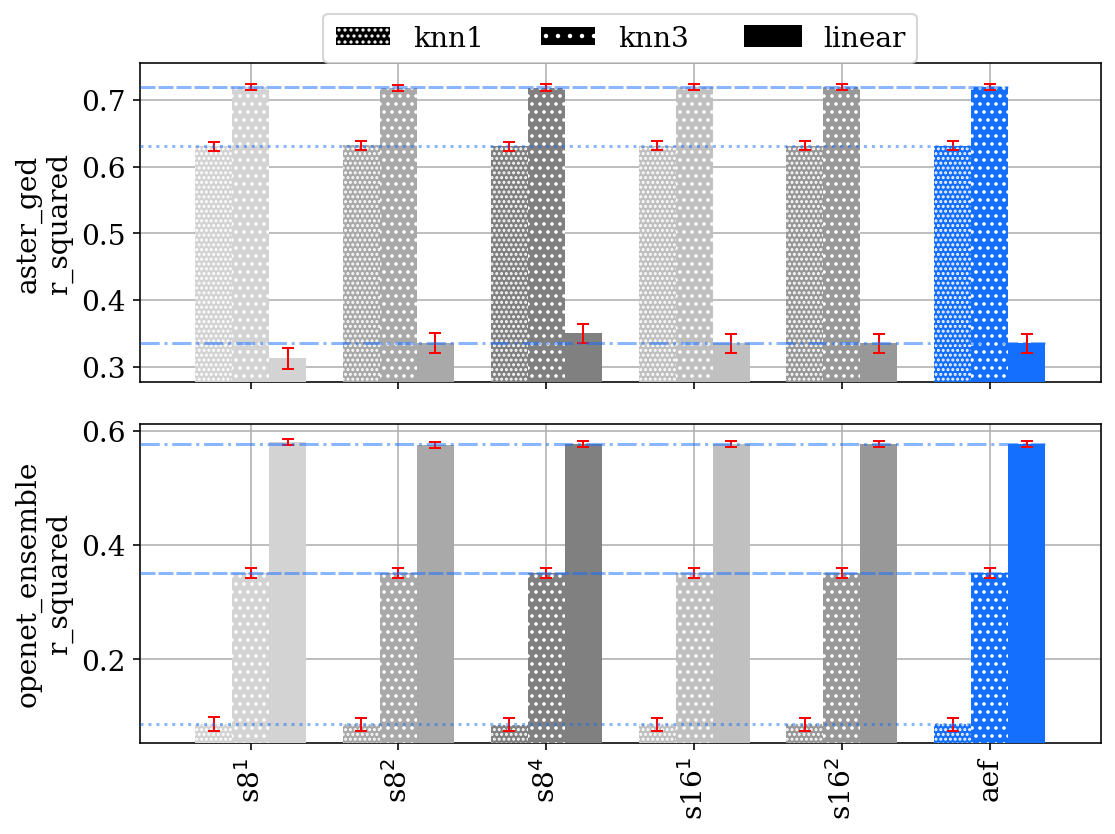}
        \subcaption{Quantization results for regression evals.}
        \label{fig:quantization-regression}
    \end{subfigure}
\end{figure}

\begin{figure}\ContinuedFloat
    \begin{subfigure}{\columnwidth}
        \includegraphics[width=\columnwidth]{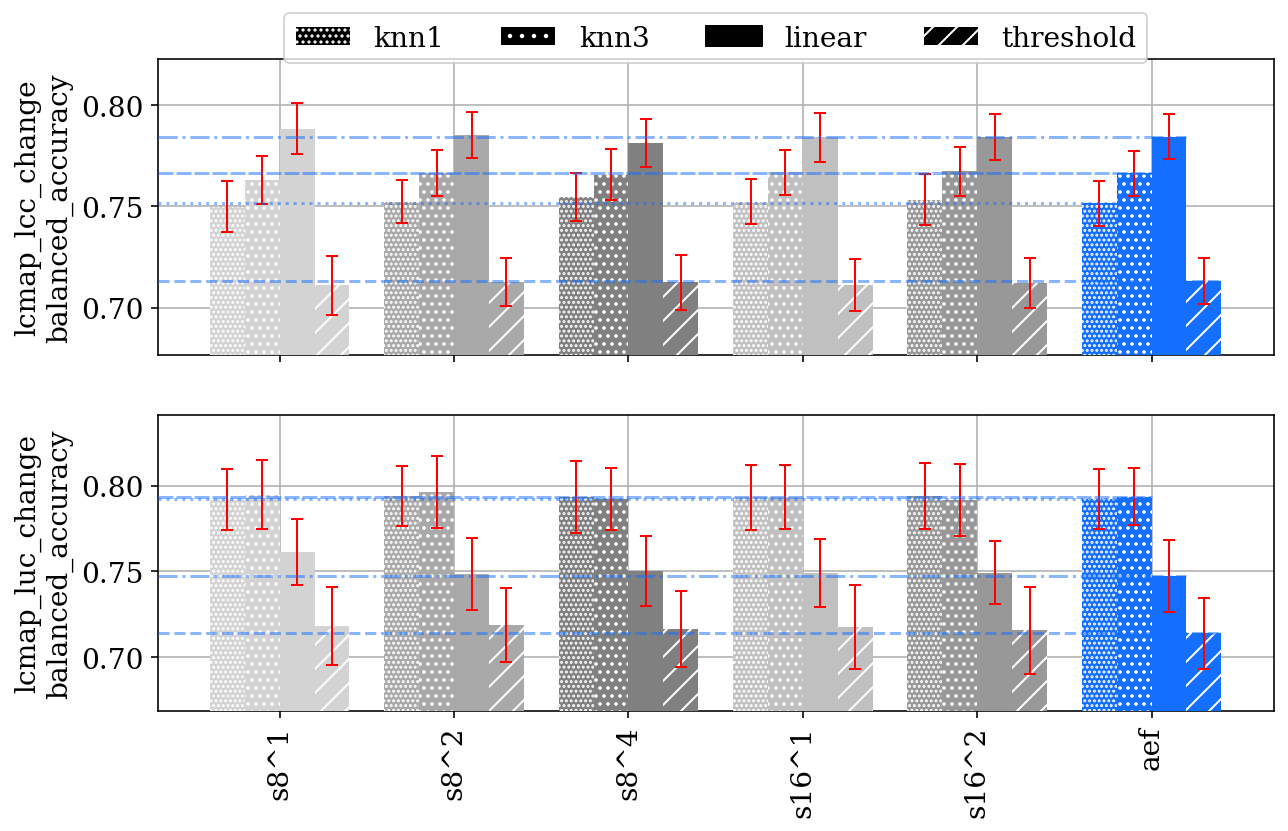}
        \subcaption{Quantization results for change detection evals.}
        \label{fig:quantization-changedetection}
    \end{subfigure}
\end{figure}

\begin{figure*}\ContinuedFloat
    \begin{subfigure}{\textwidth}
        \includegraphics[width=\textwidth]{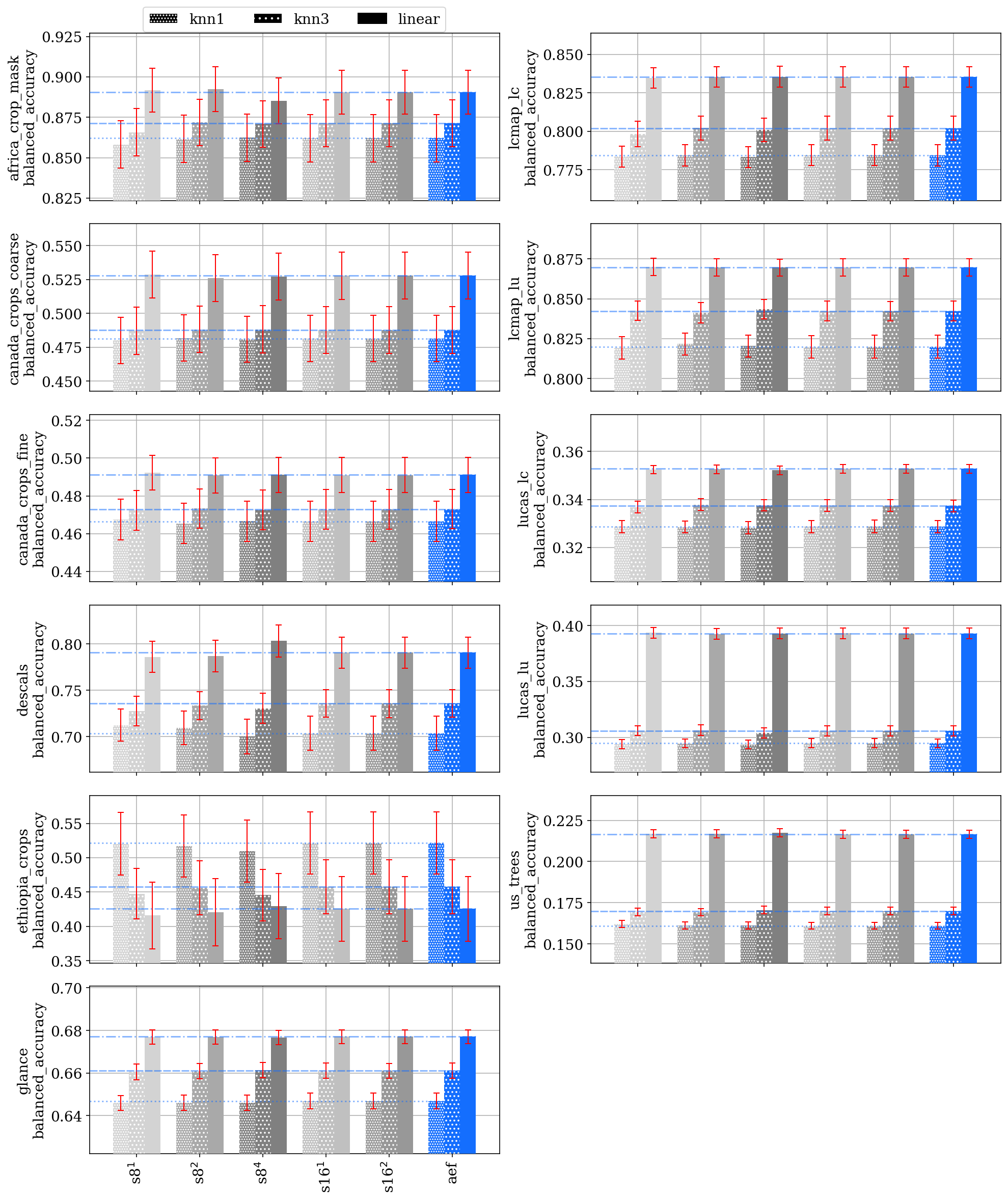}
        \subcaption{Quantization results for classification evals.}
        \label{fig:quantization-classification}
    \end{subfigure}
    \caption{Quantization results. $\text{s}8^2$ is the quantization strategy used for our released embedding fields data. The exponent on the x-axis labels indicates the quantization power.}
    \label{fig:quantization}
\end{figure*}

\subsection{Creating embedding fields}\label{S8.2}
To produce global, annual, embedding field layers, we divide the world into UTM zones, and run inference in a tiling of each zone by 960m x 960m. Before input sources are collected, each tile is individually buffered by 160m on each side (overtiling) to provide a 1.28km x 1.28km tile for inference. Sources are collected using the same protocol as for our training data, a model forward pass is run, and the outer 80m is trimmed from each tile before rendering back onto the UTM zone. We include the tiles that slightly extend beyond UTM zone degree boundaries to avoid seams when using the data in different projections.

Our inference system is the same as our training data collection system, and is backed entirely by Earth Engine. A great degree of care was taken to ensure our inference system respects the shared capacity of Earth Engine while scaling to hundreds of billions of observations. As we've demonstrated strong performance in our annual-period evaluations, we hope our annual embedding fields enable more practitioners to achieve similar results without the need or expense of pushing field campaigns to meet the needs of custom deep learning workflows.

\subsection{Production model (AEF v2.1)}\label{S8.3}
The results presented in this study reflect the performance of v2.0 of the AEF model. Based on feedback we received on annual embedding fields generated with the v2.0 model, we made a number of fixes and improvements. Specifically:

\begin{enumerate}
  \item We removed requirements that specific targets or input sensors be present in our training sample and re-generated our training dataset. This lead to the addition of a large number of samples from Antarctica that had previously been dropped due to limited coverage, and increased the count of our training video sequences from 8,412,511 to 10,182,450 sequences.
  \item Independent tests revealed a performance regression for crop classification in the conterminous United States. We determined that this was related to the inclusion of NLCD in the training mixture. We addressed this by adding additional data from the USDA Cropland Data Layers (CDL) \citep{usda2024cdl} through 2023 (omitting the data from 2024) as a target, and the loss weight for NLCD and CDL was lowered from $0.50$ to $0.25$.
  \item We identified and fixed a bug in our processing software that where incorrect handling of time codes for Sentinel-2 images acquired on January 1 resulted in divergent embedding values and visible swath artifacts in affected years.
  \item We identified and fixed a bug in our processing software where frame sub-sampling, like that during training, was applied at inference time. Embeddings generated with AEF v2.1 now use a full year of imagery for inference.
  \item We took further steps to mitigate tiling artifacts by applying frame-dropout to the teacher model mirroring that applied to the student.
  \item We also achieved a reduction in subtle artifacts from multi-resolution pixel targets by modifying re-gridding to include random shifts within the grid size prior to downsampling.
\end{enumerate}

\end{document}